\documentclass[journal]{IEEEtran}
\usepackage{amsmath}
\usepackage{cite}
\usepackage{graphicx}
\usepackage{subfigure}
\usepackage{float}
\usepackage{stfloats}
\usepackage{subfloat}
\usepackage{multirow}
\usepackage{array}
\usepackage{lineno}
\usepackage{color}  %
\usepackage{amsmath}
\usepackage{amsfonts}
\begin{document}

\title{Mixed X-Ray Image Separation for\\Artworks with Concealed Designs}

\author{Wei Pu$^*$,
Jun-Jie Huang$^*$,
Barak Sober,
Nathan Daly,
Catherine Higgitt,
Ingrid Daubechies,
Pier Luigi Dragotti, and
Miguel R.D. Rodrigues.

\thanks{
W. Pu and M. Rodrigues are with the Department of Electronic and Electrical Engineering, University College London, UK.
J.-J. Huang and P. L. Dragotti are with Electrical and Electronic Engineering Department, Imperial College London, UK.
B. Sober is with the Department of Statistics and Data Science, The Hebrew University of Jerusalem.
N. Daly and C. Higgitt are with the Scientific Department, National Gallery, London, UK.
I. Daubechies is with the Department of Electrical and Computer Engineering, Department of Mathematics, and Rhodes Information Initiative, Duke University, US.\\
\indent Corresponding author: Jun-Jie Huang. \\
\indent $^*$These authors contributed equally to this work.\\
\indent This work is sponsored by Engineering and Physical Sciences Research Council (Ref. EP/R032785/1) and the Royal Society (Ref. NIF/R1/180735).
}}

\maketitle
\maketitle
\begin{abstract}
In this paper, we focus on X-ray images of paintings with concealed sub-surface designs (\textit{e.g.}, deriving from reuse of the painting support or revision of a composition by the artist), which include contributions from both the surface painting and the concealed features. In particular, we propose a self-supervised deep learning-based image separation approach that can be applied to the X-ray images from such paintings to separate them into two hypothetical X-ray images. One of these reconstructed images is related to the X-ray image of the concealed painting, while the second one contains only information related to the X-ray of the visible painting. 
The proposed separation network consists of two components: the analysis and the synthesis sub-networks.
The analysis sub-network is based on learned coupled iterative shrinkage thresholding algorithms (LCISTA) designed using algorithm unrolling techniques, and the synthesis sub-network consists of several linear mappings. The learning algorithm operates in a totally self-supervised fashion without requiring a sample set that contains both the mixed X-ray images and the separated ones.
The proposed method is demonstrated on a real painting with concealed content, \textsl{Doña Isabel de Porcel} by Francisco de Goya, to show its effectiveness.
\end{abstract}

\begin{IEEEkeywords}
Art Investigation, Image Separation, Deep Neural Networks, Convolutional Neural Networks, Unrolling technique
\end{IEEEkeywords}

\section{Introduction} \label{section I}

Old Master paintings are often the subject of detailed technical examination, whether to investigate an artist’s materials and technique or in support of conservation or restoration treatments. 
{\color{black}In recent years, the technical examination of paintings has undergone a major digital revolution, with the widespread adoption of cutting-edge analytical and imaging technologies, generating large and typically multi-dimensional datasets\cite{ArtI1,ArtI2,ArtI3}.}

While they have a long history of use, traditional X-radiographs (X-ray images) still play a vital role in informing the technical study, conservation, and preservation of artworks in cultural heritage institutions due to the ability of X-rays to penetrate deep into a painting's stratigraphy\cite{Xray1,Xray2}. They can help to establish the condition of a painting (\textit{e.g.}, losses and damages not apparent at the surface), the status of different paint passages (\textit{e.g.}, to identify retouching, fills or other conservation interventions) or provide information about the painting support (\textit{e.g.}, type of canvas or the construction of a canvas or panel). X-ray images also provide insight into how the artist built up the different paint layers, thus revealing \textsl{pentimenti} -- changes made by the artist during painting -- which may include concealed earlier designs that were painted over when the artist revised the design or if the painting support was reused by the artist. There are many such artworks with concealed paintings with research from the Van Gogh Museum in Amsterdam showing that 20 of 130 paintings by Van Gogh, \textit{i.e.}, nearly 15$\%$, contained concealed paintings\cite{VanGogh}.

Therefore, in order to improve the understanding of the artworks and artists' working practice, there is a lot of interest in the ability to derive clearer visualisations of such hidden designs. Some works have proposed approaches leveraging various imaging modalities to enhance visualisation of concealed images in paintings\cite{Hidden4}, improve imaging of underdrawings\cite{Hidden5,Leon,Hidden6, Su2} or help reveal overwritten texts such as those found in palimpsests. X-ray image separation approaches have been proposed in a series of works such as \cite{IS1,IS2,IS3,IS4,IS5}. However, such approaches apply to double-sided painting panels where -- in addition to the mixed X-ray image -- one also has access to two RGB (visible) images associated with the front and back sides of the artwork. 

\begin{figure}[h]
    \centering
    \subfigure[]{\includegraphics[height=0.33\textwidth]{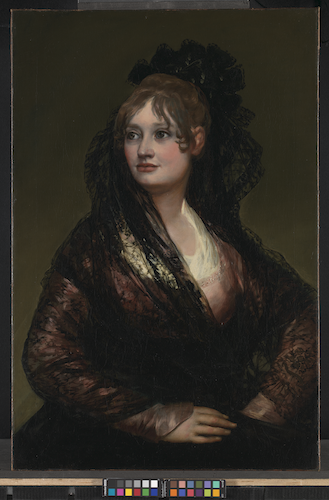}}
    \hfil
    \subfigure[]{\includegraphics[height=0.33\textwidth]{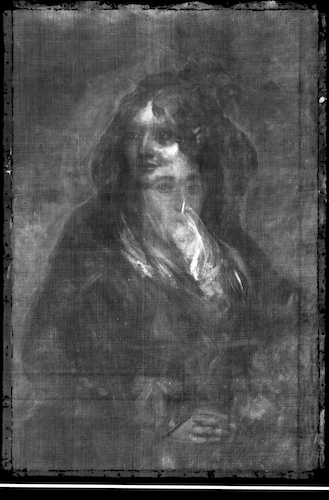}}
    \caption{ Francisco de Goya, \textsl{Doña Isabel de Porcel}(NG1473), before 1805. Oil on canvas. © The National Gallery, London. (a). RGB image. (b). X-ray image.}\label{Goya}
\end{figure}

The approach described in the current paper applies instead to a much more challenging scenario where one has access only to the mixed X-ray image (Fig. \ref{Goya} b) plus a single RGB image associated with the visible portion of the painting (Fig. \ref{Goya} (a)).
Extracting images of the concealed paintings from mixed X-ray images is challenging since X-ray images are two-dimensional representations of three-dimensional works of art, which means that the individual X-ray image includes features from both the concealed design and that seen at the surface (see for example Fig. \ref{Goya}), as well as features associated with the painting support (\textit{e.g.} canvas weave or wood grain). This makes it difficult for experts to distinguish the elements belonging to the concealed painting.

In this paper, inspired by our previous work in \cite{IS6}, we propose a new deep learning-based approach to separate a mixed X-ray image associated with paintings with concealed designs into their hypothetical constituent images, corresponding to the X-ray images of the visible painting and of the concealed image below.\footnote{In this work, features in the mixed X-ray images associated with the painting support are not directly treated as our primary aim is to improve visualisation of concealed design.}
Our proposed approach -- which is entirely self-supervised -- leverages only the mixed X-ray image along with the RGB image of the painting. Our method has been applied to real paintings with mixed X-ray images, including Francisco de Goya's portrait of \textsl{Doña Isabel de Porcel}, illustrated in Fig. \ref{Goya}, that has been painted directly on top of another portrait of a male figure\cite{Goya}\footnote{An improved visualisation and a greater understanding of the underlying portrait has been possible through the use of macro X-ray fluorescence scanning, but it is also of great interest to attempt to obtain a clearer image of just the lower figure\cite{R1}.}.
In contrast to \cite{IS6}, by leveraging sparsity-driven models and algorithm unrolling techniques, the architecture presented in this paper was designed to be more readily interpretable, with the added benefit of leading up to better X-ray image separation results.

\subsection{Paper Contributions}

In this paper, we propose a separation network for the task of X-ray image separation associated with paintings with concealed designs given the visible image of the surface painting. Our contributions are as follows:
\begin{itemize}

\item Firstly, we formulate the X-ray image separation problem of paintings with concealed designs by leveraging a coupled sparse coding model, propose a coupled iterative shrinkage-thresholding algorithm (CISTA), and turn this CISTA  solver into a layered network architecture -- denoted by learned coupled iterative shrinkage-thresholding algorithm (LCISTA) -- using algorithm unrolling techniques.

\item Secondly, we propose a separation network, which consists of an analysis component and a synthesis component, for the mixed X-ray image separation task that utilizes the RGB image of the surface painting. 
The analysis component aims to extract the sparse representations from the mixed X-ray image and RGB image of the surface painting while the synthesis component aims to reconstruct the mixed X-ray image and the RGB image of the surface painting from such sparse representations.
The analysis component is designed based upon LCISTA, and the synthesis component is designed using a linear convolutional mapping.

\item Thirdly, we propose a composite loss function involving reconstruction and exclusion losses to learn the weights of the separation network.

\item Finally, we apply our proposed approach to synthetic and real datasets, showcasing the effectiveness of the proposed algorithm. 

\end{itemize}


\subsection{Paper Organization}
The paper is organized as follows: Section II introduces the related work on source separation and on the X-ray image separation problem. In Section III, the proposed deep learning-based X-ray separation method for the paintings with concealed designs is presented. Section IV shows the X-ray separation results of the proposed method on both synthetic and real data, and Section V concludes the paper.

\section{Related work}


From the perspective of signal processing, the X-ray image separation problem mentioned above is relevant to the source separation problem, where the aim is to reconstruct the individual source signals from some mixture of these signals. In general, source separation problems can be categorized into two broad categories: 1) blind source separation (BSS) and 2) informed source separation (ISS). 

In BSS, the individual source signals are recovered merely from the mixed signal, which is highly ill-posed. The corresponding approaches include independent component analysis (ICA)\cite{ICA1,ICA2}, robust-principal component analysis (RPCA)\cite{RPCA1,RPCA2,RPCA3}, or morphological component analysis (MCA)\cite{MCA1,MCA2,MCA3}. These methods however always impose strong prior assumptions on the source signal such as independence\cite{ICA1,ICA2}, sparsity\cite{RPCA1,RPCA2,RPCA3}, low-rankness\cite{RPCA1,RPCA2,RPCA3}, and non-Gaussianity\cite{MCA1,MCA2}. 
In principle, BSS-type algorithms can be applied to our task, however, the corresponding separation performances cannot meet the real application requirement because 1) such typical prior assumptions adopted do not always hold and 2) the availability of side information (\textit{e.g.} the RGB image of the surface painting) is not exploited.

In ISS, we have some additional information to promote a better separation. Depending on whether or not the additional information belongs to the direct observation of source signals, the ISS problem can be divided into two types: 1) ISS with source information and 2) ISS with side information.
The ISS problem with source information, which appears in many scenarios in audio, speech and language processing, can involve two different degrees of supervision: 1) fully-supervised and 2) semi-supervised. The fully-supervised case typically assumes one has access to a training dataset containing a number of examples of mixed and associated component signals that can be used to train a deep neural network carrying out the separation task \cite{SeparateSupervised1,SeparateSupervised2}. In the semi-supervised case one may have access to samples of one individual source but not other sources. For example, a neural egg separation (NES) method\cite{SeparateSemiSupervised} integrated with generative adversarial networks (GANs) \cite{GAN} has been recently proposed to tackle the semi-supervised source separation challenge.

With the ISS problem with side information, although one does not have direct observation of the distribution or examples of the source signal, some other information (\textit{i.e.}, supplementary information associated with the source signal from some other data modality) is available to aid the separation. 
In the particular application of technical examination of art, some approaches have been proposed to separate mixed X-ray images with double-sided paintings, where RGB images pertaining to both front and back sides of the art work are available. These include sparsity-based methods\cite{IS1}, Gaussian mixture model-based approaches{\cite{IS2}} and deep neural networks\cite{IS3,IS4,IS5}. 

\begin{figure}
\centering
\includegraphics[width=0.45\textwidth]{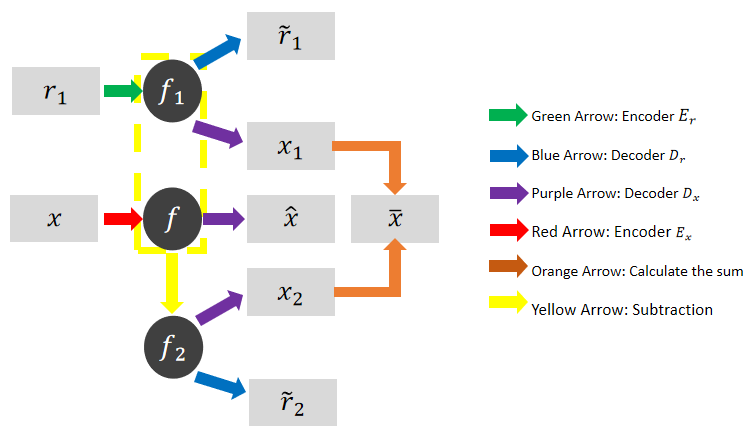}
\caption{Connected auto-encoder in \cite{IS6}.}\label{fig1}
\end{figure}

However, compared with the X-ray image separation problem of double-sided paintings\cite{IS1,IS2,IS3,IS4,IS5}, the mixed X-ray image separation of painting with concealed designs is a much more challenging scenario. There have been attempts to address this challenge.
In \cite{IS6}, the authors firstly assume that the mixed X-ray image is the sum of the individual X-ray images as follows:\footnote{This linear mixing assumption is motivated by the fact that paintings (and panels) can be quite thin so that higher-order attenuation effects can be neglected.} 
\begin{align}
\label{1.1}
{\boldsymbol x}  \approx {\boldsymbol x}_1  + {\boldsymbol x}_2,
\end{align}
where ${\boldsymbol x} $ denotes a mixed X-ray image patch and ${\boldsymbol x}_1 $ and ${\boldsymbol x}_2 $ denote the individual X-ray image patches corresponding to the surface and concealed paintings, respectively. 
Then, a connected auto-encoder as shown in Fig. \ref{fig1} is proposed to separate the mixed X-ray image.
The connected auto-encoder involves:
\begin{itemize}
	\item Extracting feature map ${\boldsymbol  f}_1$ from the RGB image patch ${\boldsymbol  r}_1$ using an encoder $E_r$ (represented by the green arrow), where ${\boldsymbol  f}_1$ denotes the latent feature map associated with the surface painting, and ${\boldsymbol  r}_1$ denotes RGB image patch of surface painting;
	\item Extracting feature map $\boldsymbol f$ from the image patch ${\boldsymbol x}$ using the encoder $E_x$ (represented by the red arrows), where $\boldsymbol f$ denotes the latent feature map of the mixed X-ray image;
	\item Obtaining latent feature map ${\boldsymbol  f}_2$ corresponding to the concealed design using a ``subtraction" layer (represented by the yellow arrows) as follows ${\boldsymbol  f}_2 = {\boldsymbol f}-{\boldsymbol  f}_1$;
	\item Reconstructing X-ray image patches $\hat {\boldsymbol x}_1$, $\hat {\boldsymbol x}_2$ and $\hat {\boldsymbol x}$ from ${\boldsymbol  f}_1$, ${\boldsymbol  f}_2$ and $\boldsymbol f$ using decoder $D_x$ (represented by the purple arrow);
    \item Regenerating RGB image patch $\hat {\boldsymbol  r}_1$ from ${\boldsymbol  f}_1$ using decoder $D_r$ (represented by the blue arrow);
	\item Recovering mixed X-ray image patch $\bar {\boldsymbol x}$ using an ``addition" layer (represented by the orange arrows) as ${\bar {\boldsymbol x}} = {\hat {\boldsymbol x}}_1+{\hat {\boldsymbol x}}_2$.
\end{itemize}

The encoders/decoders $E_r$, $E_x$, $D_r$ and $D_x$ were also taken to be three-layer convolutional neural networks and are learned by minimizing the following composite loss:
\begin{align}
\label{1.2}
\left \| {\boldsymbol x} -\hat {\boldsymbol x} \right \|_F + \left \| {\boldsymbol x} -\bar {\boldsymbol x} \right \|_F + \lambda_1 \left \| {\boldsymbol  r}_1-\hat {\boldsymbol  r}_1 \right \|_F + \lambda_2 E({\boldsymbol  g}_1, \hat {\boldsymbol x}_2),
\end{align}
where $\lambda_1$ and $\lambda_2$ are regularization parameters, ${\boldsymbol  g}_1$ denotes the greyscale version of the RGB image patch of the surface painting and $E({\boldsymbol x}_1,{\boldsymbol x}_2)$ -- the exclusion loss adopted from \cite{EL} -- measures the correlation between two edge maps at multiple spatial resolutions as follows:
\begin{align}
\label{1.3}
E({\boldsymbol x}_1,{\boldsymbol x}_2) = \sum_{n=1}^N || {\cal C} ( h^{\downarrow n}(\sigma_1 |\nabla {\boldsymbol x}_1|), h^{\downarrow n}(\sigma_2 |\nabla {\boldsymbol x}_2|) ) ||_F.
\end{align}
In (\ref{1.3}), ${\cal C} ( {\boldsymbol x}_1, {\boldsymbol x}_2 ) =\tanh({\boldsymbol x}_1) \odot \tanh({\boldsymbol x}_2)$, $\odot$ is element-wise multiplication, $h^{\downarrow n} (\cdot)$ denotes the down-sampling operation by a factor of $2^{n-1}$ with bilinear interpolation, $\nabla {\boldsymbol x}_1$ and $\nabla {\boldsymbol x}_2$ denote the gradients of $ {\boldsymbol x}_1$ and $ {\boldsymbol x}_2$, respectively, and in \cite{IS6}, the authors set $N=3$, $\sigma_1 = \sqrt{{\| {\boldsymbol x}_2\|_F}/{\|{\boldsymbol  x}_1\|_F}}$, and $\sigma_2 = \sqrt{{\|{\boldsymbol  x}_1\|_F}/{\|{\boldsymbol x}_2\|_F}}$.

The method in \cite{IS6} leads to some partially successful separation results. 
{\color{black}However, the method in \cite{IS6} appears to incorporate too much information from the RGB image when extracting the X-ray image of the surface painting, so that the resulting image is very similar to a grayscale version of the RGB image and appears to have lost some of the detailed information present in the mixed X-ray image.}
Our new proposed method described in the sequel addressed these drawbacks and it is also more interpretable because we design the separation apparatus from first principles.

\section{Proposed Approach}
In order to address the limitations of the method in \cite{IS6}, a new approach to separate mixed X-ray images taken from paintings with concealed designs given access to the visible image of the surface painting is proposed.

Our proposed approach develops state-of-the-art sparsity-driven image processing techniques coupled with algorithm unrolling techniques\cite{Unroll}. We next elaborate about the underlying models, the proposed separation network, and the separation network learning strategy.

\subsection{Model}
We propose to use convolutional sparse coding techniques to model each individual image patch. The convolutional sparse coding paradigm is an extension of the sparse coding model in which a redundant dictionary is modeled as a concatenation of circulant matrices. In the convolutional sparse coding paradigm, the global sparsity constraint of the target signal, which describes the target signal as a linear combination of a few atoms in the redundant dictionary, is exploited to promote accurate reconstruction.
The rationale behind using convolutional sparse coding is that on the one hand it yields state-of-the-art performance\cite{CNN}, whereas on the other hand it can be unrolled into a convolutional neural network architecture\cite{UnfoldingCNN}.
Additionally, the algorithm based on convolutional neural networks is capable of dealing with the RGB version of the visual image patches instead of the grayscale version of the same image patches, so it can also capture colour information that might be relevant to improve the mixed X-ray image separation performance. 

The convolutional sparse coding model can be characterized as follows: 
\begin{align}
\label{1.4}
&{\boldsymbol x}_1 = \sum_{k=1}^K {\boldsymbol  \Xi}_k * {\boldsymbol  z}_{1,k}, \quad {\boldsymbol x}_2 = \sum_{k=1}^K {\boldsymbol  \Xi}_k * {\boldsymbol  z}_{2,k},\nonumber\\
&{\boldsymbol  r}_{1,s} =\sum_{k=1}^K {\boldsymbol  \Omega}_{k,s} * {\boldsymbol  z}_{1,k},\quad {\boldsymbol x} = \sum_{k=1}^K {\boldsymbol  \Xi}_k * ({\boldsymbol  z}_{1,k}+{\boldsymbol  z}_{2,k}),
\end{align}
where ${\boldsymbol x}_1 $ and ${\boldsymbol x}_2 $ denote the individual X-ray image patches corresponding to the surface and concealed paintings, respectively, ${\boldsymbol  r}_{1,s} $ for $s=1,2,3$ denotes the red, green and blue channel patches of RGB image of the surface painting, ${\boldsymbol  z}_{1,k}$ and ${\boldsymbol  z}_{2,k}$ denote the sparse representations underlying the X-ray image patches of the surface painting and concealed design, respectively, and $k = 1,2,\cdots,K$ indexes the channel number.
$ {\boldsymbol  \Omega}_{k,s}  $ denotes the $k$-th convolutional dictionary filter for the RGB image patches of the $s$-th channel, ${\boldsymbol  \Xi}_k  $ denotes the $k$-th convolutional dictionary filter for the X-ray image patches, and $*$ denotes the convolution processing. 
The convolution operation ${\boldsymbol  w} = {\boldsymbol  a} * {\boldsymbol  b}$ between two image patches ${\boldsymbol  a}$ and ${\boldsymbol  b}$ is given by:
\begin{align}
\label{1.5}
{\boldsymbol  w}(i,j) = \sum_p \sum_q {\boldsymbol  a}(p,q) {\boldsymbol  b}(i-p+1,j-q+1) .
\end{align}

Note that the model in (\ref{1.4}) immediately links the various images by imposing that the X-ray and RGB image patches associated with the same layer of the painting share the same sparse representation. Moreover, the X-ray image patches of the surface painting and concealed design share the same dictionaries. This model also imposes that the mixed X-ray is equal to the sum of the individual X-rays (as in other works as mentioned earlier\cite{IS5,IS6}).

\subsection{Separation Network}
\begin{figure}[t]
\centering
\includegraphics[width=0.45\textwidth]{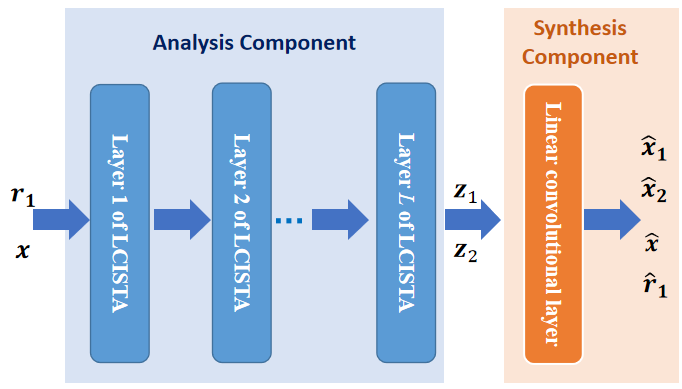}
\caption{General structure of the proposed separation network.}\label{f1}
\end{figure}
We then propose to separate the mixed X-ray image patch into its individual constituent image patches using a deep neural network structure, which consists of two components: the analysis and synthesis components shown in Fig. \ref{f1}. The analysis component produces the sparse representations of the mixed X-ray image and RGB image of the surface painting and the synthesis component produces the reconstruction of the mixed X-ray image and RGB image of the surface painting. The analysis component is designed based on algorithm unrolling techniques and the synthesis component is designed based on a linear convolutional model.
Specifically, we use the following four steps to design the analysis and synthesis components:
\begin{enumerate}
\item Firstly, we formulate the X-ray image separation problem into a coupled sparse coding problem under the hypothesis that dictionaries are known.

\item Secondly, we develop a solver to this problem using coupled iterative shrinkage-thresholding algorithm (CISTA).

\item Thirdly, we design the analysis component by turning the CISTA solver into a layered network architecture -- denoted by learned coupled iterative shrinkage-thresholding algorithm (LCISTA) -- based on algorithm unrolling techniques\cite{Unroll}.

\item Finally, we design the synthesis component based on the linear model presented in (\ref{1.4}).
\end{enumerate}
In what follows, we describe these steps in more detail.

\subsubsection{X-Ray Image Separation Problem Formulation}
As discussed previously, the analysis component in the separation network extracts sparse codes ${\boldsymbol  z}_{1}$ and ${\boldsymbol  z}_{2}$ from the inputs ${\boldsymbol  r}_{1}$ and ${\boldsymbol  x}$. 
In order to design the analysis component using an algorithm unrolling technique, we need to formulate a coupled sparse coding problem to estimate the sparse codes ${\boldsymbol  z}_{1}$ and ${\boldsymbol  z}_{2}$ from ${\boldsymbol  r}_{1}$ and ${\boldsymbol  x}$ assuming to begin with that the dictionaries in (\ref{1.4}) are known. Then, we can design the analysis component by unrolling the corresponding solver of the coupled sparse coding problem.

{\color{black}
Prior to problem formulation, we introduce two auxiliary parameters ${\boldsymbol  y}_1 $ and ${\boldsymbol  y}_2 $, representing information image patches of the surface painting and concealed design, respectively. The main purpose for introducing ${\boldsymbol  y}_1 $ and ${\boldsymbol  y}_2 $ is to facilitate the subsequent addition of exclusion loss in the information image domain. 
Correspondingly, the convolutional sparse coding model in (\ref{1.4}) is changed into 
\begin{align}
\label{1.4.1}
&{\boldsymbol x}_1 = {\boldsymbol  \Psi}*  {\boldsymbol  y}_1, \quad {\boldsymbol x}_2 =  {\boldsymbol  \Psi}*  {\boldsymbol  y}_2,\nonumber\\
&{\boldsymbol  r}_{1,s} ={\boldsymbol  \Phi}_{s} * {\boldsymbol  y}_{1},\quad {\boldsymbol x} = {\boldsymbol  \Psi}*  ({\boldsymbol  y}_1+{\boldsymbol  y}_2),\nonumber\\
&{\boldsymbol  y}_{1} = \sum_{k=1}^K {\boldsymbol  \Theta}_k * {\boldsymbol  z}_{1,k} \quad {\boldsymbol  y}_{2} = \sum_{k=1}^K {\boldsymbol  \Theta}_k * {\boldsymbol  z}_{2,k},
\end{align}
where ${\boldsymbol  \Psi}$, ${\boldsymbol  \Theta}_k$ and ${\boldsymbol  \Phi}_{s}$ denote the dictionaries with respect to the X-ray image patches, information image patches and RGB image patches, respectively.}
The X-ray image separation problem associated with the model in (\ref{1.4}) can be formulated as follows:
\begin{align}
\label{2.3}
\min_{{\boldsymbol  y}_1, {\boldsymbol  y}_2,{\boldsymbol  z}_{1,k},{\boldsymbol  z}_{2,k}}  \quad &  \|{\boldsymbol x} - {\boldsymbol  \Psi}*  {\boldsymbol  y}_1-  {\boldsymbol  \Psi}*  {\boldsymbol  y}_2\|_F^2  \nonumber\\
+ &\tau_1\|{\boldsymbol  y}_{1} - \sum_{k=1}^K {\boldsymbol  \Theta}_k * {\boldsymbol  z}_{1,k}\|_F^2 \nonumber\\
+ &\tau_2\|{\boldsymbol  y}_{2} - \sum_{k=1}^K {\boldsymbol  \Theta}_k * {\boldsymbol  z}_{2,k}\|_F^2 \nonumber\\
+ &\gamma \sum_{s=1}^3 \|{\boldsymbol  r}_{1,s} - {\boldsymbol  \Phi}_{s} * {\boldsymbol  y}_{1}\|_F^2  \nonumber\\
+ &\lambda_1  \sum_{k=1}^K \| {\boldsymbol  z}_{1,k} \|_1 + \lambda_2  \sum_{k=1}^K \| {\boldsymbol  z}_{2,k} \|_1 \nonumber \\
+ & \sum_{i=1}^I \mu_i \| ({\boldsymbol  W}_i * {\boldsymbol  y}_1) \odot ({\boldsymbol  W}_i * {\boldsymbol  y}_2)\|_1,
\end{align}
where $\gamma$, $\tau_1$, $\tau_2$, $\lambda_1$, $\lambda_2$, and $\mu_k$ are the regularization parameters.
In the last term, ${\boldsymbol  W} = [{\boldsymbol  W}_1,{\boldsymbol  W}_2,\cdots,{\boldsymbol  W}_I]$ denotes a redundant wavelet transform which is a union of $I$ orthogonal transforms and is used to sparsify ${\boldsymbol  y}_1$ and ${\boldsymbol  y}_2 $ for exclusion loss evaluation.
At this stage, we assume that dictionaries ${\boldsymbol  \Psi}$, ${\boldsymbol  \Theta}_k$ and ${\boldsymbol  \Phi}_{s}$ are known.

In (\ref{2.3}), the first to fourth terms correspond to the data consistency terms with respect to the mixed X-ray image patch, information image patch of the surface painting and concealed design, and the RGB image patch of the surface painting, respectively.
The fifth and sixth terms correspond to a $l_1$ regularization term to guarantee the sparsity of the representations.
The last term corresponds to a simplified version of exclusion loss\cite{EL}, in order to simplify the subsequent optimization algorithm and network design. 
By using the simplified version of exclusion loss, we expect to obtain the edge maps of the information image patches ${\boldsymbol  y}_1 $ and ${\boldsymbol  y}_2 $ using wavelet transforms, in order to promote their disentanglement ((since the images from the visible and concealed design are typically different)). 

The problem in (\ref{2.3}) without the exclusion loss term is ill-posed. 
That is, there are some undesired minimizers of (\ref{2.3}).
For example, the separated X-ray image of the concealed design based on one possible solution of ${\boldsymbol  z}_2 $ may contain much content from the surface painting. The reason to introduce the simplified version of exclusion loss is to give constraints on the information image patches ${\boldsymbol  y}_1 $ and ${\boldsymbol  y}_2 $ to make them as different as possible from one another.
This ensures that information associated with the surface painting does not incorrectly appear in the separated X-ray image of the concealed design.

\begin{figure*}[t]
\centering
\includegraphics[width=0.8\textwidth]{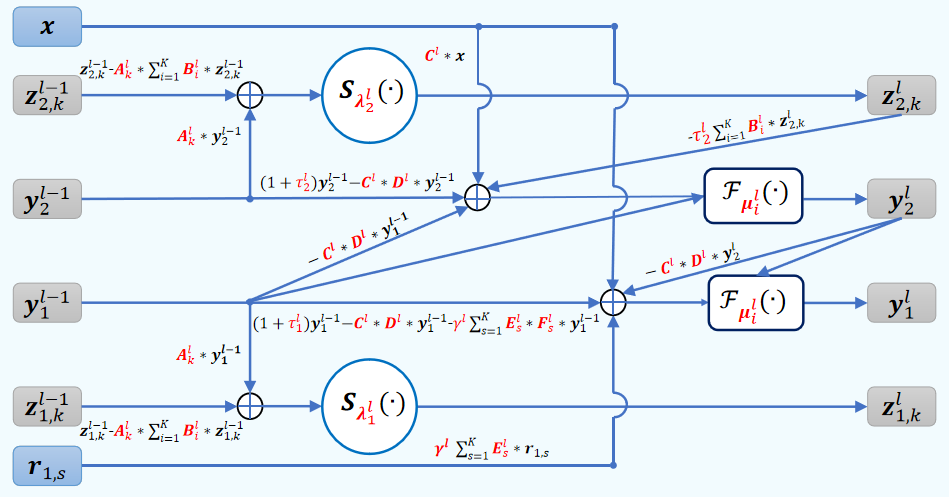}
\caption{Structure of the $l$-th layer in LCISTA.}\label{f2}
\end{figure*}

\subsubsection{Solver}
Next, we use an iterative CISTA algorithm in order to solve the coupled sparse coding problem in (\ref{2.3}).
We split the problem in (\ref{2.3}) into several sub-problems and solve each of them iteratively. In particular, (\ref{2.3}) is changed into (\ref{2.4}),
\begin{figure*}[h]
\begin{align}
\label{2.4}
{\boldsymbol  z}^{l}_{2,k} = \arg \min_{{\boldsymbol  z}_{2,k}}   &  \tau_2\|{\boldsymbol  y}^{l-1}_{2} - \sum_{k=1}^K {\boldsymbol  \Theta}_k * {\boldsymbol  z}_{2,k}\|_F^2 +\lambda_2  \sum_{k=1}^K \| {\boldsymbol  z}_{2,k} \|_1, \quad
{\boldsymbol  z}^{l}_{1,k} = \arg \min_{{\boldsymbol  z}_{1,k}}     \tau_1\|{\boldsymbol  y}^{l-1}_{1} - \sum_{k=1}^K {\boldsymbol  \Theta}_k * {\boldsymbol  z}_{1,k}\|_F^2 +\lambda_1  \sum_{k=1}^K \| {\boldsymbol  z}_{1,k} \|_1, \nonumber \\
{\boldsymbol  y}^{l}_{2} = \arg \min_{{\boldsymbol  y}_{2}}   &  \|{\boldsymbol x} - {\boldsymbol  \Psi}*  {\boldsymbol  y}^{l-1}_1-  {\boldsymbol  \Psi}*  {\boldsymbol  y}_2\|_F^2  
+ \tau_2\|{\boldsymbol  y}_{2} - \sum_{k=1}^K {\boldsymbol  \Theta}_k * {\boldsymbol  z}^{l}_{2,k}\|_F^2 + \sum_{i=1}^I \mu_i \| {\boldsymbol  W}_i {\boldsymbol  y}^{l-1}_1 \odot {\boldsymbol  W}_i {\boldsymbol  y}_2\|_1, \nonumber \\
{\boldsymbol  y}^{l}_{1} = \arg \min_{{\boldsymbol  y}_{1}}   &  \|{\boldsymbol x} - {\boldsymbol  \Psi}*  {\boldsymbol  y}_1-  {\boldsymbol  \Psi}*  {\boldsymbol  y}^{l}_2\|_F^2  
+ \tau_1\|{\boldsymbol  y}_{1} - \sum_{k=1}^K {\boldsymbol  \Theta}_k * {\boldsymbol  z}^{l}_{1,k}\|_F^2 + \gamma \sum_{s=1}^3 \|{\boldsymbol  r}_{1,s} - {\boldsymbol  \Phi}_{s} * {\boldsymbol  y}_{1}\|_F^2  
+ \sum_{i=1}^I \mu_i \| {\boldsymbol  W}_i {\boldsymbol  y}_1 \odot {\boldsymbol  W}_i {\boldsymbol  y}^{l}_2\|_1,
\end{align}
\end{figure*}
where $l$ denotes the iteration number.
By taking the gradients on the data consistency terms in (\ref{2.4}) and executing a proximal step on the last term of each sub-problem, we obtain a series of iterations\cite{TV} shown in (\ref{2.5}),
\begin{figure*}[h]
\begin{align}
\label{2.5}
{\boldsymbol  z}_{2,k}^{l}
=&  {\cal S}_{\frac{\lambda_2}{\xi} }\left( {\boldsymbol z}_{2,k}^{l-1} + \frac{ {  {\boldsymbol  \Theta}}_k^T}{\xi} *({\boldsymbol  y}^{l-1}_2 -\sum_{i=1}^K {\boldsymbol  \Theta}_i * {\boldsymbol  z}_{2,i}^{l-1} ) \right), \quad
{\boldsymbol  z}_{1,k}^{l}
=  {\cal S}_{\frac{\lambda_1}{\xi} }\left( {\boldsymbol z}_{1,k}^{l-1} + \frac{ {  {\boldsymbol  \Theta}}_k^T}{\xi} *({\boldsymbol  y}^{l-1}_1 -\sum_{i=1}^K {\boldsymbol  \Theta}_i * {\boldsymbol  z}_{1,i}^{l-1} ) \right), \nonumber\\
{\boldsymbol  y}_2^{l} =& {\cal F}_{\frac{\mu_i}{\xi} }\left ({\boldsymbol y}_1^{l-1},{\boldsymbol  y}_2^{l-1} + \frac{{\boldsymbol  \Psi}^T}{\xi}*( {\boldsymbol x} - {\boldsymbol  \Psi} * ({\boldsymbol  y}_1^{l-1} +  {\boldsymbol  y}_2^{l-1}))  + \frac{\tau_2}{\xi} ({\boldsymbol  y}_2^{l-1} -\sum_{k=1}^K{\boldsymbol  \Theta}_i * {\boldsymbol  z}_{2,i}^{l}) \right),\nonumber\\
{\boldsymbol  y}_1^{l} =& {\cal F}_{\frac{\mu_i}{\xi} }\left ({\boldsymbol y}_2^{l},{\boldsymbol  y}_1^{l-1} + \frac{{\boldsymbol  \Psi}^T}{\xi}*( {\boldsymbol x} - {\boldsymbol  \Psi} ({\boldsymbol  y}_1^{l-1} + {\boldsymbol  y}_2^{l}))  + \frac{\tau_1}{\xi} ({\boldsymbol  y}_1^{l-1} - \sum_{k=1}^K  {\boldsymbol  \Theta}_i * {\boldsymbol  z}_{1,i}^{l}) +  \frac{\gamma}{\xi} \sum_{s=1}^3 {\boldsymbol  \Phi}_s^T * ({\boldsymbol  r}_{1,s} - {\boldsymbol  \Phi}_{s} * {\boldsymbol  y}_{1}^{l-1} ) \right ),
\end{align} 
\end{figure*}
where $\frac{1}{\xi} >0$ is the step size and operator $S_{\sigma} (\cdot)$ is the soft thresholding operator applied element-wise on its input as 
\begin{align}
\label{2.6}
{\cal S}_{\sigma}( {\boldsymbol x} ) = {\rm sign}({\boldsymbol x})\cdot \max (|{\boldsymbol x}| -\sigma , 0).
\end{align}
Here, we define ${\cal F}_a({\boldsymbol b},{\boldsymbol c}) = \sum_{i=1}^I {\boldsymbol  W}_i^T * {\cal S}_{a\|{\boldsymbol  W}_i * {\boldsymbol  b}\|_1}\left( {\boldsymbol  W}_i * {\boldsymbol c} \right)$ for simplicity. ${\cal F}_a({\boldsymbol b},{\boldsymbol c})$ is a parallel proximal operator which computes several independent proximals\cite{TV}. It has been theoretically proven in \cite{TV} that the algorithm converges when using a parallel proximal operator to solve the least-squares cost function with the simplified exclusion loss as a regularizer.

\subsubsection{Analysis component}
Our third step is to design the analysis component of the separation network using unfolding techniques\cite{Unroll}. The iterative solver is converted into a feedforward layered neural network architecture, \textit{i.e.}, LCISTA. We can then map each solver iteration operation in (\ref{2.5}) onto a feedforward neural network operation, and likewise we can also map $L$ solver iterations onto a $L$ layer feedforward neural network. We change (\ref{2.5}) into (\ref{2.7}), where convolutional filters $[{\boldsymbol  A}^l_k, {\boldsymbol  B}^l_k, {\boldsymbol  C}^l, {\boldsymbol  D}^l, {\boldsymbol  E}^l_{s}, {\boldsymbol  F}^l_{s}]$ and scalar parameters $[ \tau_1^l, \tau_2^l, \lambda_1^l, \lambda_2^l, \mu_i^l, \gamma^l ]$ are set to be learnable parameters.
Each network layer of LCISTA is represented in Fig. \ref{f2}, and the learnable parameters are emphasized in red.

The rationale for adopting new parameters to describe the neural network layer instead of the original ones derives from the fact that we can further learn this using entirely self-supervised mechanisms.
\begin{figure*}[h]
\begin{align}
\label{2.7}
{\boldsymbol  z}_{2,k}^{l}
=&  {\cal S}_{\lambda_2^{l} }\left( {\boldsymbol z}_{2,k}^{l-1} + {\boldsymbol  A}_k^{l}*({\boldsymbol  y}^{l-1}_2 -\sum_{i=1}^K {\boldsymbol  B}^{l}_k * {\boldsymbol  z}_{2,i}^{l-1} ) \right), \quad
{\boldsymbol  z}_{1,k}^{l}
=  {\cal S}_{\lambda_1^{l}}\left( {\boldsymbol z}_{1,k}^{l-1} + {\boldsymbol  A}_k^{l}*({\boldsymbol  y}^{l-1}_1 -\sum_{i=1}^K {\boldsymbol  B}^{l}_k * {\boldsymbol  z}_{1,i}^{l-1} ) \right), \nonumber\\
{\boldsymbol  y}_2^{l} =& {\cal F}_{\mu_i^{l}}\left ({\boldsymbol y}_1^{l-1},{\boldsymbol  y}_2^{l-1} + {\boldsymbol  C}^{l}*( {\boldsymbol x} - {\boldsymbol  D}^{l} * ({\boldsymbol  y}_1^{l-1} +  {\boldsymbol  y}_2^{l-1}))  + {\tau^{l}_2} ({\boldsymbol  y}_2^{l-1} -\sum_{i=1}^K{\boldsymbol  B}^{l}_i * {\boldsymbol  z}_{2,i}^{l}) \right),\nonumber\\
{\boldsymbol  y}_1^{l} =& {\cal F}_{\mu_i^{l}}\left ({\boldsymbol y}_2^{l},{\boldsymbol  y}_1^{l-1} + {\boldsymbol  C}^{l}*( {\boldsymbol x} - {\boldsymbol  D}^{l}* ({\boldsymbol  y}_1^{l-1} + {\boldsymbol  y}_2^{l}))  + {\tau_1^{l}} ({\boldsymbol  y}_1^{l-1} - \sum_{i=1}^K  {\boldsymbol  B}^{l}_i * {\boldsymbol  z}_{1,i}^{l}) +  {\gamma^{l}} \sum_{s=1}^3 {\boldsymbol  E}^{l}_s * ({\boldsymbol  r}_{1,s} - {\boldsymbol  F}^{l}_{s} * {\boldsymbol  y}_{1}^{l-1} )\right),
\end{align} 
\end{figure*}
Note that the learnable parameters in the analysis component are set to be the same in each layer to give the separation network more restrictions and to promote a better separation performance.

\subsubsection{Synthesis component}
Suppose we have $L$ layers in total in the analysis component of the separation network, and assume the outputs of the analysis component are ${\boldsymbol  z}_{1,k}^L$ and ${\boldsymbol  z}_{2,k}^L$. Then, the synthesis component is designed to convert the sparse feature ${\boldsymbol  z}_{1,k}^L$ into an estimate of the visual image patches and of the mixed X-ray image patches. Specifically, in line with our model in (4) and (6), we have that
\begin{align}
\label{2.8}
\hat {\boldsymbol  r}_{1,s} = \sum_{k=1}^K {\boldsymbol  W}_{\Omega;k,s} * {\boldsymbol  z}_{1,k}^L  ,
\end{align}
and 
\begin{align}
\label{2.9}
\hat {\boldsymbol x} = \sum_{k=1}^K  {\boldsymbol  W}_{\Xi;k} * ({\boldsymbol  z}_{1,k}^L + {\boldsymbol  z}_{2,k}^L).
\end{align}
Here, ${\boldsymbol  W}_{\Omega;k,s} $ and ${\boldsymbol  W}_{\Xi;k}$ are also set to be learnable parameters too.

It is important to introduce the synthesis component because it allows the proposed separation approach to work in a totally self-supervised manner (note that we do not have access to true sparse representation ${\boldsymbol  z}_{1,k}$ and${\boldsymbol  z}_{2,k}$ to train the analysis network but we do have access to mixed X-ray image and RGB image patches to train the concatenation of the analysis and synthesis networks). We design the synthesis component to regenerate the RGB image patch ${\boldsymbol  r}_{1,s}$ and mixed X-ray image patch ${\boldsymbol  x}$ from ${\boldsymbol  z}_{1,k}$ and ${\boldsymbol  z}_{2,k}$ so that standard reconstruction losses can be utilized to guide the training phase.

\subsection{Learning Strategy}

During the training of the proposed separation network, we randomly initialize the learnable parameters of the network, \textit{i.e.}, initialized convolutional filters $[{\boldsymbol  A}^l_k, {\boldsymbol  B}^l_k, {\boldsymbol  C}^l, {\boldsymbol  D}^l, {\boldsymbol  E}^l_{s}, {\boldsymbol  F}^l_{s}, {\boldsymbol  W}_{\Omega;k,s}, {\boldsymbol  W}_{\Omega;k,s}]$ satisfy multivariate Gaussian distributions and initialized scalars $[ \tau_1^l, \tau_2^l, \lambda_1^l, \lambda_2^l, \mu_i^l, \gamma^l ]$ are uniformly distributed in the interval $(0,1]$. The inputs of the separation network are set as
\begin{align}
\label{2.13}
&{\boldsymbol  z}_{1,k}^0  ={\boldsymbol  0}, \quad\quad {\boldsymbol  z}_{1,k}^0  = {\boldsymbol  0}, \nonumber\\
&{\boldsymbol  y}_{1}^0  ={\boldsymbol  g}_{1}, \quad\quad {\boldsymbol  y}_{2}^0  ={\boldsymbol  x} - {\boldsymbol  g}_{1} .
\end{align}  

Then, the learnable parameter of the whole networks ${\boldsymbol  w} = [{\boldsymbol  A}^l_k, {\boldsymbol  B}^l_k, {\boldsymbol  C}^l, {\boldsymbol  D}^l, {\boldsymbol  E}^l_{s}, {\boldsymbol  F}^l_{s}, {\boldsymbol  W}_{\Omega;k,s}, {\boldsymbol  W}_{\Omega;k,s}, \tau_1^l, \tau_2^l, \lambda_1^l, \lambda_2^l, \mu_i^l, \gamma^l ]$ are learnt as follow:
\begin{align}
\label{2.14}
\min_{\boldsymbol  w} \quad & \| {\boldsymbol x} - \hat {\boldsymbol x} \|^2_F +\eta_1 \sum_{s=1}^3\| {\boldsymbol r}_{1,s} - \hat {\boldsymbol r}_{1,s}  \|^2_F+ \eta_2  E({\boldsymbol y}_1^L,{\boldsymbol y}_2^L),
\end{align}  
where $\eta_1$ and $\eta_2$ are the hyper-parameters pertaining to the reconstruction loss of the surface painting image patch and exclusion loss, respectively. We then optimize the separation network learnable parameters by using stochastic gradient descent (SGD) with learning rate $l_r = 10^{-3-e_p/40}$, where $e_p$ denotes the epoch number.
Additionally, we use 120 epochs in total to train the separation network, and set the network architecture parameters as $K=64$ and $I=4$. All the convolutional filters are of size $5 \times 5$.

Note once again that the proposed network is learnt in a totally self-supervised fashion based on a training set containing only patches of the mixed X-ray image and RGB image of the surface painting.

\section{Experimental Results}
We conducted a number of experiments to assess the effectiveness of our proposed X-ray separation approach. These involved:
\begin{itemize}

\item an analysis of the effect of the various regularization parameters associated with our approach on X-ray separation performance;

\item an analysis of the effectiveness of our approach in relation to the state-of-the-art method in \cite{IS6}, both on synthetically mixed X-ray images and real mixed X-ray images.

\end{itemize}

\subsection{Datasets}

\begin{figure}[t]
    \centering
    \subfigure[]{\includegraphics[width=0.23\textwidth]{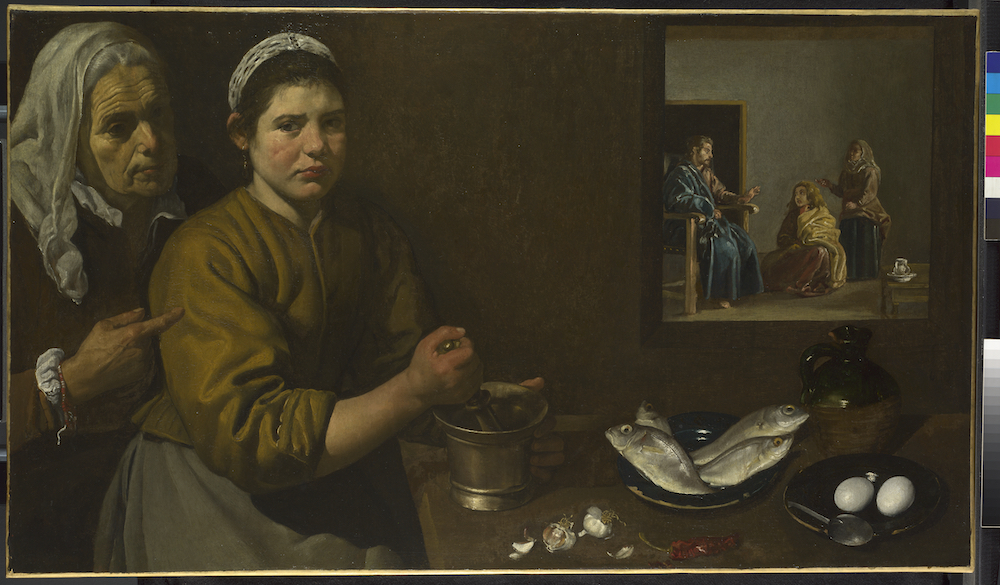}}
    \hfil
    \subfigure[]{\includegraphics[width=0.23\textwidth]{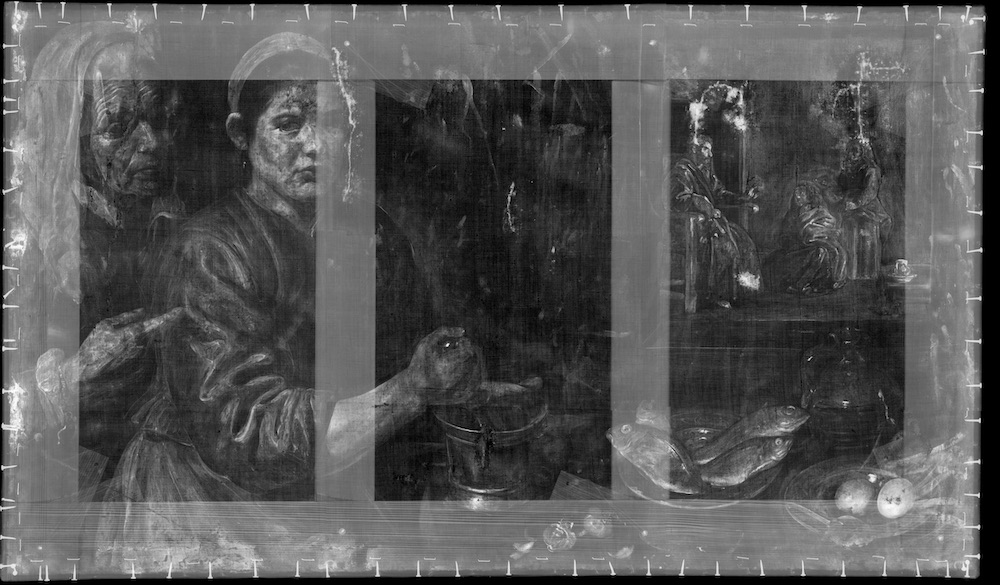}}
    \caption{ Diego Velázquez, \textsl{Kitchen Scene with Christ in the House of Martha and Mary} (NG1375), probably 1618. Oil on canvas © The National Gallery, London. (a). RGB image. (b). X-ray image.}\label{F-4-1-1}
\end{figure}

\begin{figure}[t]
\centering
\includegraphics[width=0.48\textwidth]{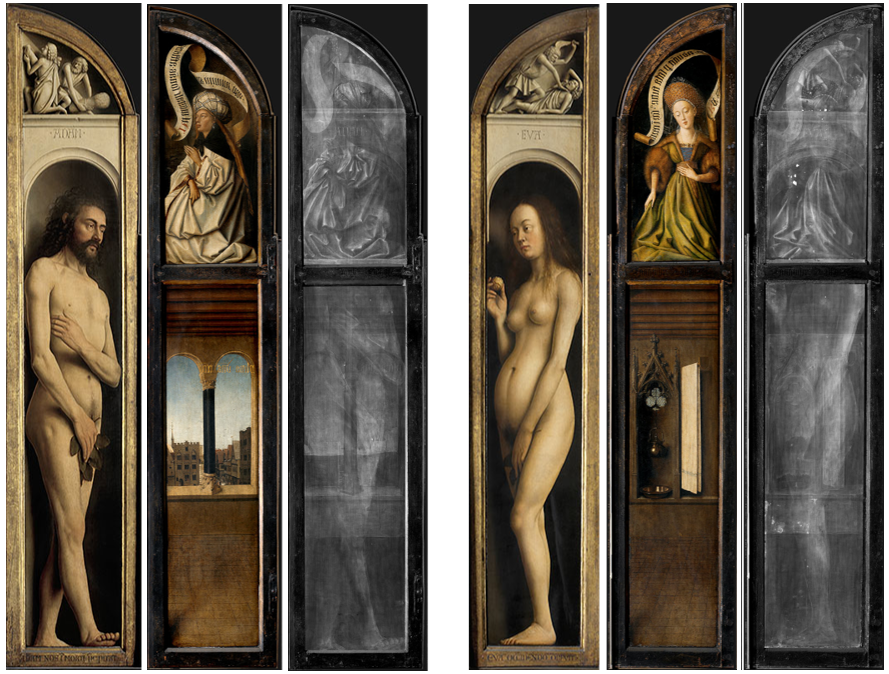}
\caption{Two double-sided panels from the \textsl{Ghent Altarpiece}\cite{Ghent}: (left) Adam panel, (right) Eve panel. Columns 1-3 correspond to the visible image of front side, the visible image of rear side and the mixed X-ray image of Adam panel. Columns 4-6 correspond to the visible image of front side, the visible image of rear side and the mixed X-ray image of Eve panel.}\label{Ghent}
\end{figure}

Our experiments rely on a number of datasets associated with real paintings, including:

\begin{itemize}

\item \textsl{Kitchen Scene with Christ in the House of Martha and Mary} by Diego Velázquez (Fig. \ref{F-4-1-1}). There is a single composition on this canvas and it was used to showcase the performance of our algorithm on synthetically mixed X-ray data. As this dataset is the only one for which the ground truth is available, it was also used to select the values for the hyper-parameters.

\item The \textsl{Ghent Altarpiece} by Hubert and Jan van Eyck. This large, complex 15th-century polyptych altarpiece comprises a series of panels – including panels with a composition on both sides (see Fig. \ref{Ghent}) – that we use to showcase the performance of our algorithm on real mixed X-ray data.

\item \textsl{Doña Isabel de Porcel} by Francisco de Goya, illustrated in Fig. \ref{Goya}. The portrait of Isabel has been painted directly on top of another portrait of a male figure – that we use to showcase the performance of our algorithm on real mixed X-ray data.

\end{itemize}

\subsection{Hyper-parameter Selection}

\subsubsection{Set-up}
\begin{figure}[t]
    \centering
    \subfigure[]{\includegraphics[width=0.0925\textwidth]{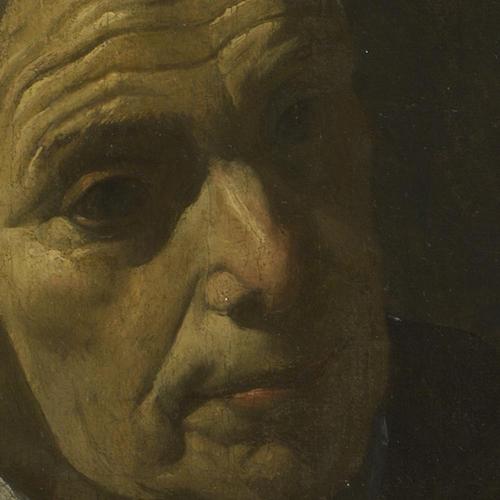}}
    \subfigure[]{\includegraphics[width=0.0925\textwidth]{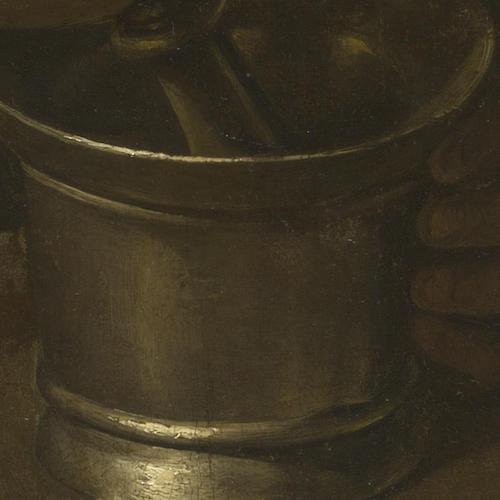}}
    \subfigure[]{\includegraphics[width=0.0925\textwidth]{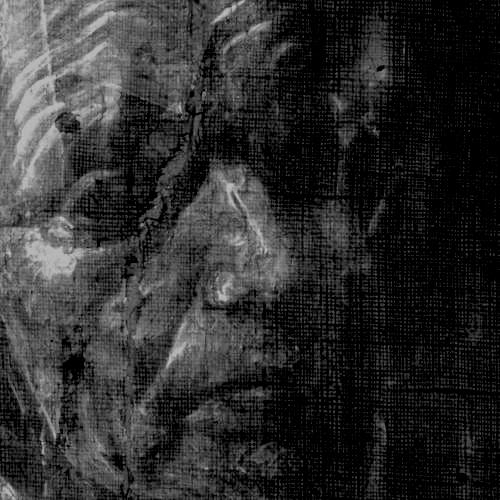}}
    \subfigure[]{\includegraphics[width=0.0925\textwidth]{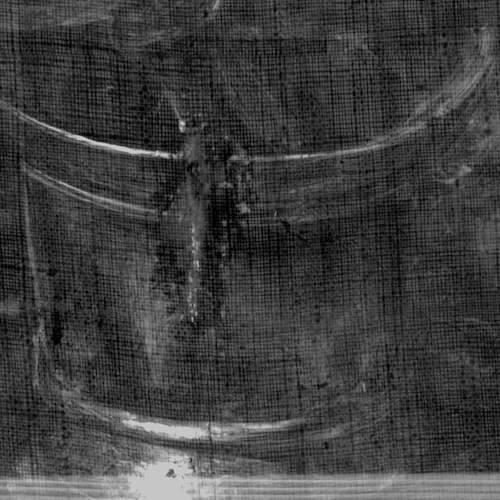}}
    \subfigure[]{\includegraphics[width=0.0925\textwidth]{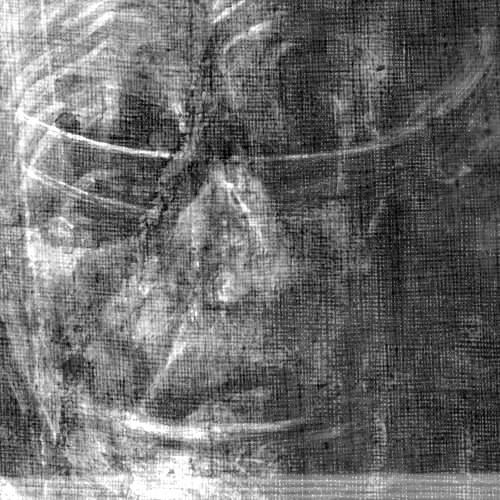}}
    \caption{Images used for hyper-parameter selection. (a) First RGB image (the `surface image'); (b) second RGB image (the `concealed design'); (c) X-ray image corresponding to the `surface image'; (d) X-ray image corresponding to `concealed design'; (e) Synthetically mixed X-ray image.}\label{F-4-P-1-1}
\end{figure}

In these experiments, we used two small areas with the same size (showing the face of the old lady and the bucket on the table) from the oil painting \textsl{Kitchen Scene with Christ in the House of Martha and Mary} by Diego Velázquez to create a synthetically mixed X-ray image (see Fig. \ref{F-4-P-1-1}). Here, we set the first area, the face of the old lady (Fig. \ref{F-4-P-1-1} (a) and (c)), as the surface painting, while the second area, the mortar on the table (Fig. \ref{F-4-P-1-1} (b) and (d)), is set as the concealed design.

The images – which are of size $600 \times 600$ pixels – were divided into patches of size $50 \times 50$ pixels with 45 pixels overlap (both in the horizontal and vertical direction), resulting in 3,136 patches. The patches associated with the synthetically mixed X-ray were then separated independently. The various patches associated with the individual separated X-rays were finally put together by placing the patches in their original order and averaging the overlapping portions. All patches were utilized in the training of the network by randomly shuffling their order. 

In order to determine the optimal values for $\eta_1$ and $\eta_2$, given that the ground truth images are available for the synthetic dataset, we chose to assess the separation performance by reporting the
average Mean Squared Error (MSE) given by:
\begin{align}
\label{e17}
{\rm MSE} = \frac{1}{2MN}  || {\boldsymbol X} -\hat {\boldsymbol X}||^2_F ,
\end{align}
where ${\boldsymbol X}$ denotes the ground truth image, $\hat {\boldsymbol X}$ is the corresponding recovered image, and $M$ and $N$ denote the size of the image.
This experimental procedure was carried out for different combinations of $\eta_1$ and $\eta_2$. We restricted these hyper-parameters to lie in the interval $\eta_1 \in [0,1], \eta_2 \in [0,0.4]$. We also selected instances of the hyper-parameters $\eta_1$ and $\eta_2$ from these intervals in steps of 0.025 and 0.01, respectively.

\subsubsection{Hyper-parameter selection}

\begin{figure}[t]
\centering
\includegraphics[width=0.4\textwidth]{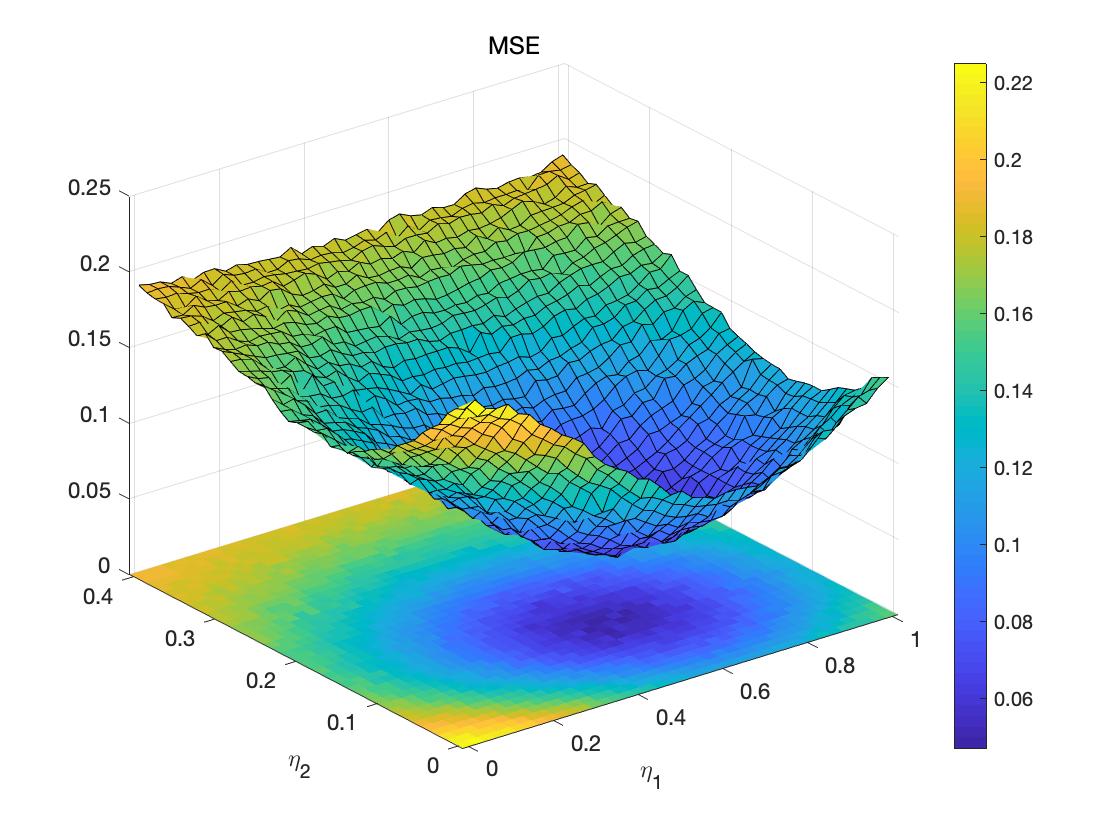}
\caption{Average MSE versus $\eta_1$ and $\eta_2$.}\label{F-4-P-2-1}
\end{figure}

Fig. \ref{F-4-P-2-1} depicts the average MSE errors of reconstructed individual X-ray images as a function of the hyper-parameters $\eta_1$ and $\eta_2$. It is clear that different hyper-parameter values result in different separation performances. According to Fig. \ref{F-4-P-2-1}, we find the optimized hyper-parameter values with the smallest reconstruction errors are $\eta_1 = 0.5$ and $\eta_2 = 0.1$. We note the optimal regularization parameter values may depend slightly on the exact datasets, but we have found that these reported values tend to lead to very good performance across a wide range of datasets.

\begin{figure*}[h]
\centering
    \subfigure{\includegraphics[width=0.115\textwidth]{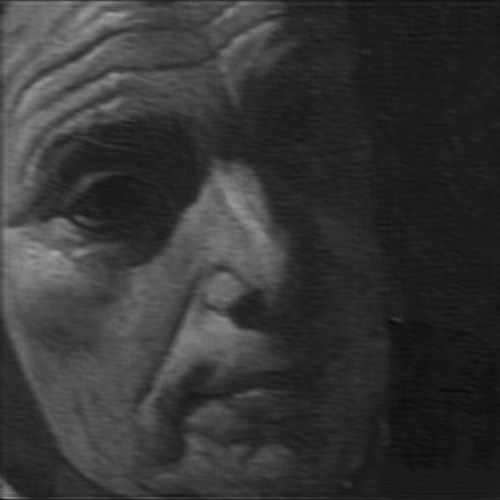}}
    \subfigure{\includegraphics[width=0.115\textwidth]{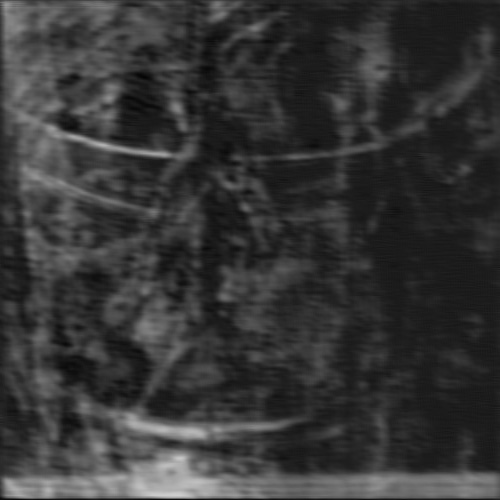}}
    \subfigure{\includegraphics[width=0.115\textwidth]{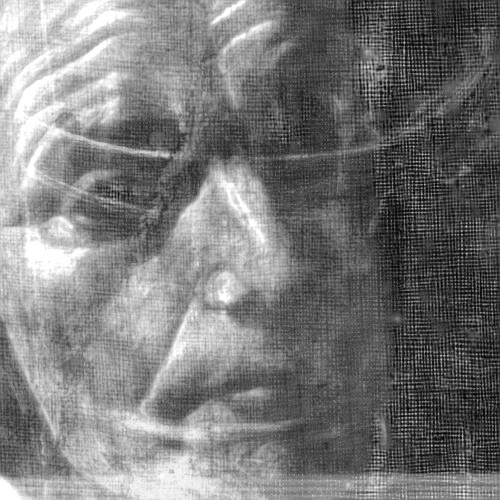}}
    \subfigure{\includegraphics[width=0.115\textwidth]{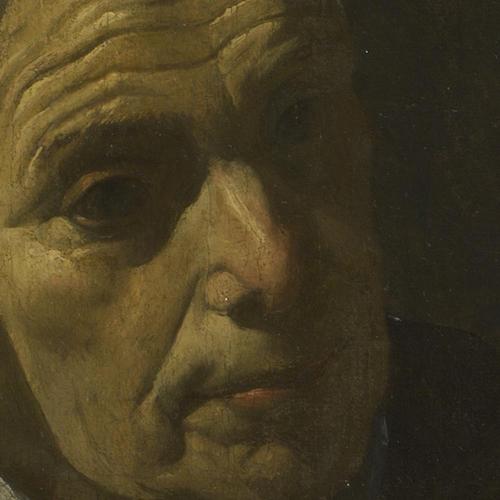}}
    \subfigure{\includegraphics[width=0.115\textwidth]{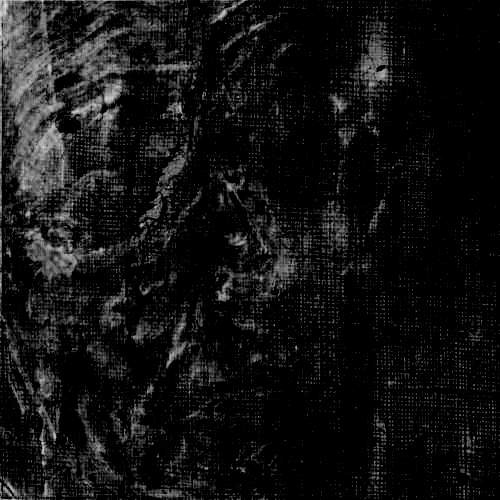}}
    \subfigure{\includegraphics[width=0.115\textwidth]{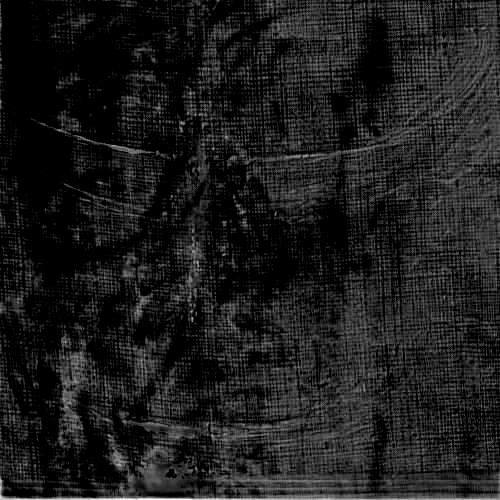}}
    \subfigure{\includegraphics[width=0.115\textwidth]{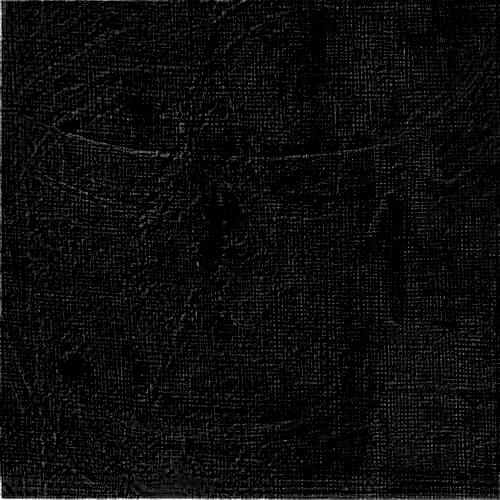}}
    \subfigure{\includegraphics[width=0.115\textwidth]{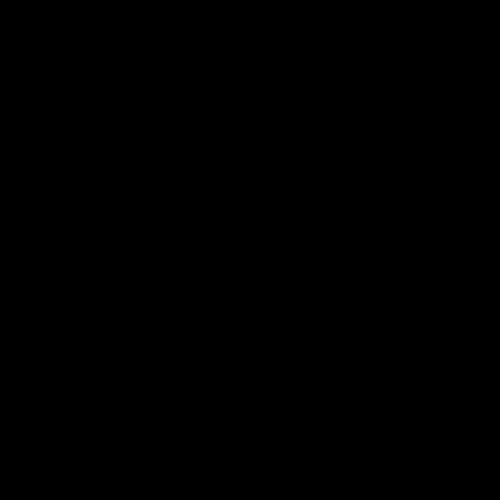}}
    
    \subfigure{\includegraphics[width=0.115\textwidth]{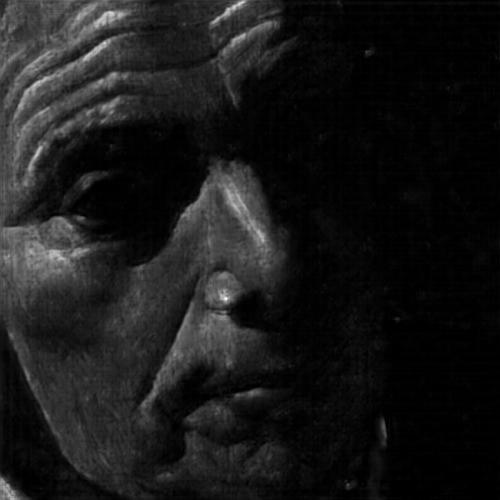}}
    \subfigure{\includegraphics[width=0.115\textwidth]{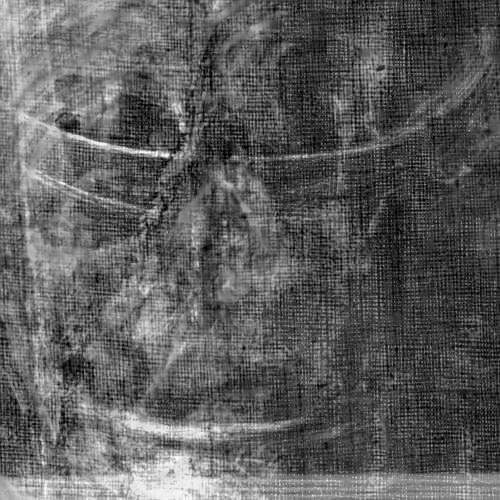}}
    \subfigure{\includegraphics[width=0.115\textwidth]{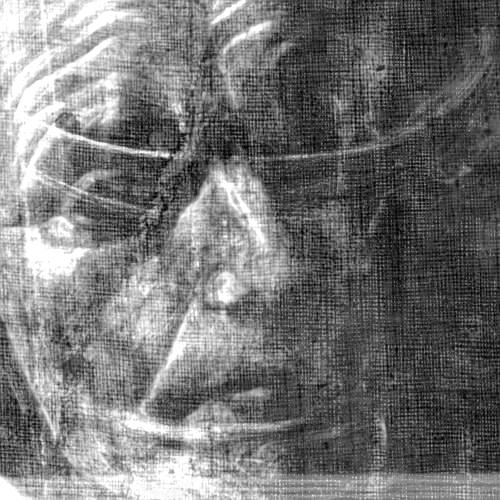}}
    \subfigure{\includegraphics[width=0.115\textwidth]{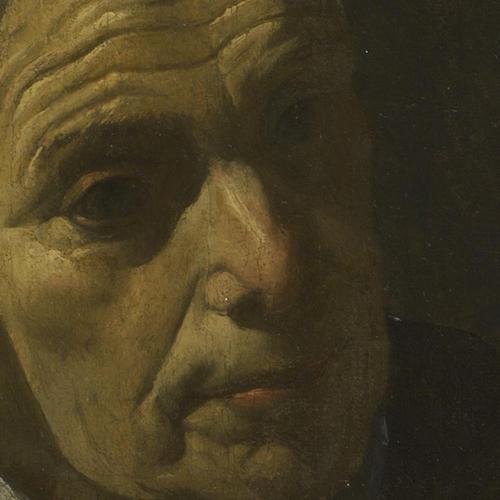}}
    \subfigure{\includegraphics[width=0.115\textwidth]{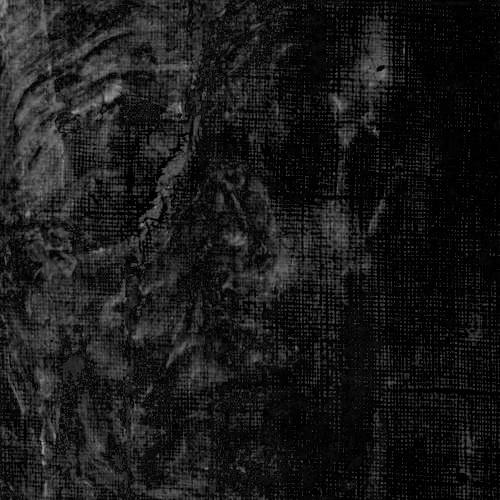}}
    \subfigure{\includegraphics[width=0.115\textwidth]{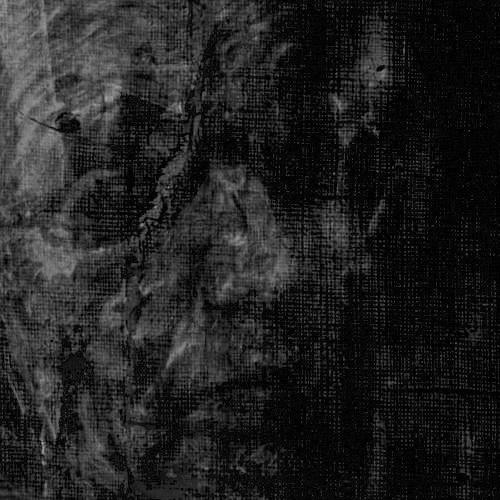}}
    \subfigure{\includegraphics[width=0.115\textwidth]{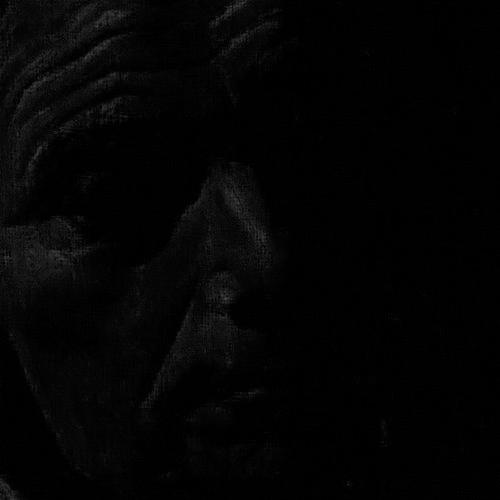}}
    \subfigure{\includegraphics[width=0.115\textwidth]{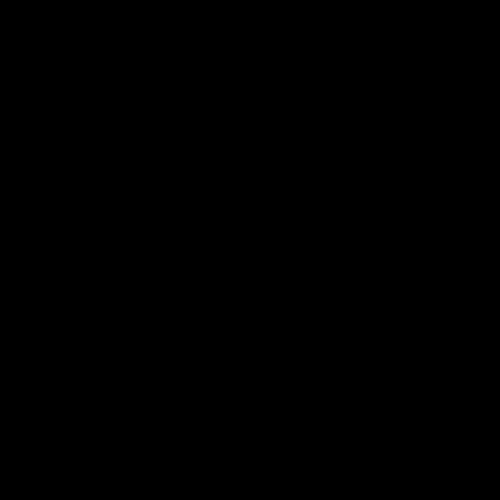}}
    
    \subfigure{\includegraphics[width=0.115\textwidth]{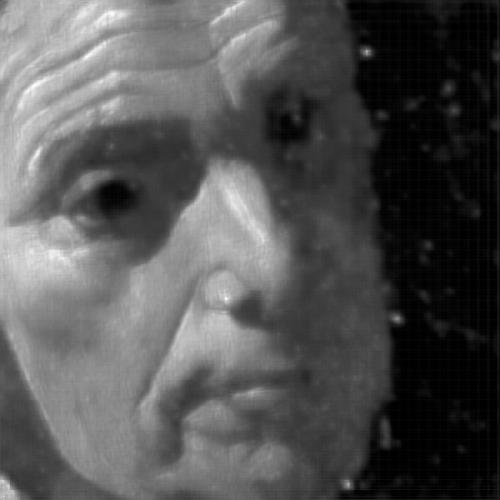}}
    \subfigure{\includegraphics[width=0.115\textwidth]{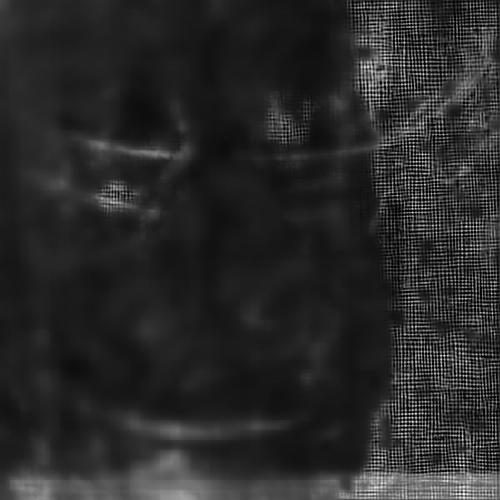}}
    \subfigure{\includegraphics[width=0.115\textwidth]{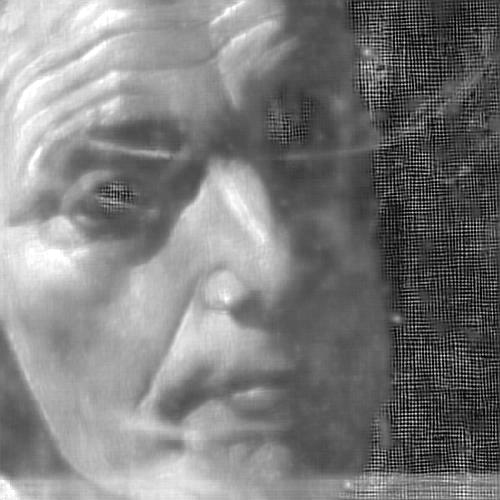}}
    \subfigure{\includegraphics[width=0.115\textwidth]{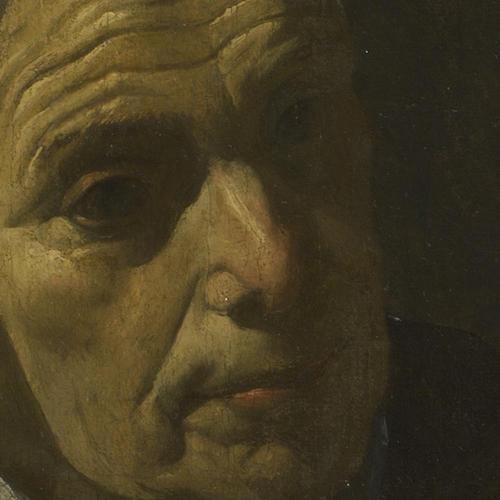}}
    \subfigure{\includegraphics[width=0.115\textwidth]{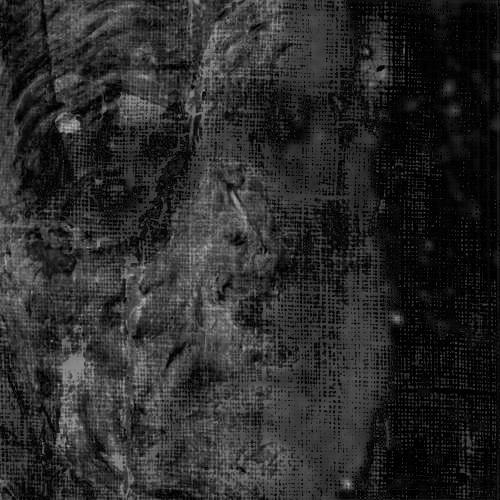}}
    \subfigure{\includegraphics[width=0.115\textwidth]{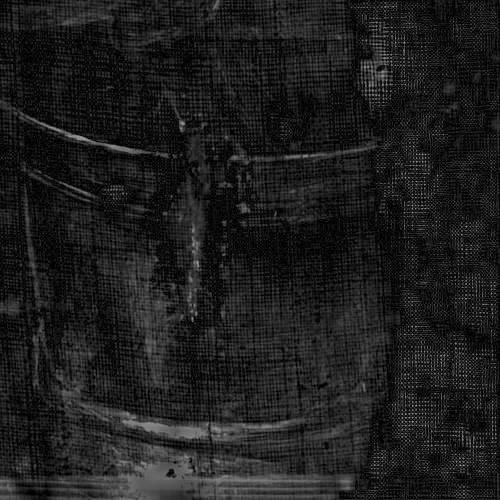}}
    \subfigure{\includegraphics[width=0.115\textwidth]{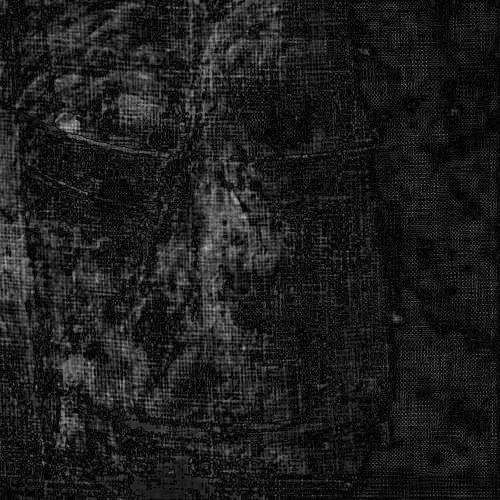}}
    \subfigure{\includegraphics[width=0.115\textwidth]{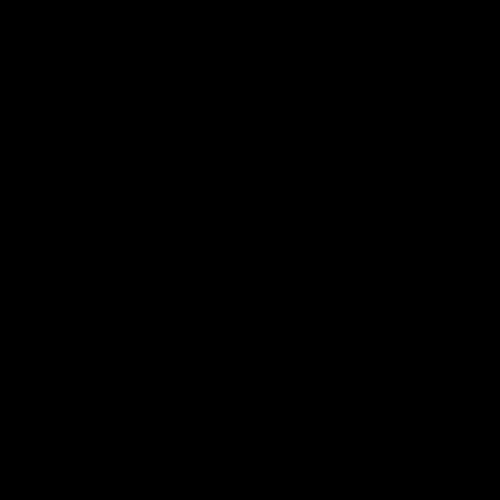}}
    
    \subfigure{\includegraphics[width=0.115\textwidth]{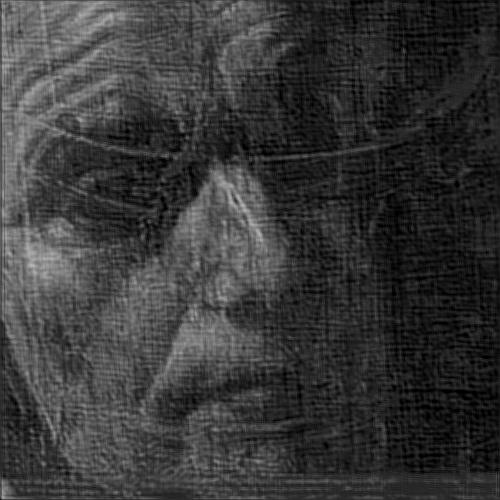}}
    \subfigure{\includegraphics[width=0.115\textwidth]{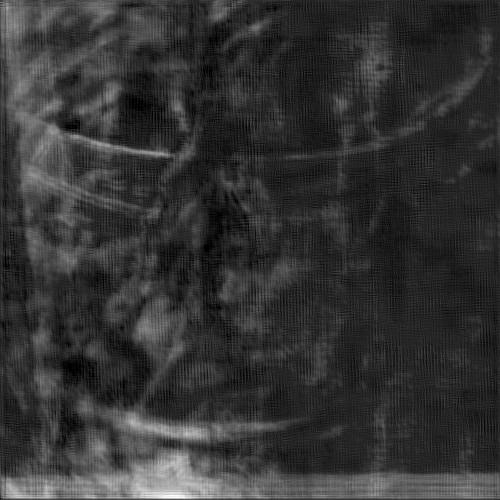}}
    \subfigure{\includegraphics[width=0.115\textwidth]{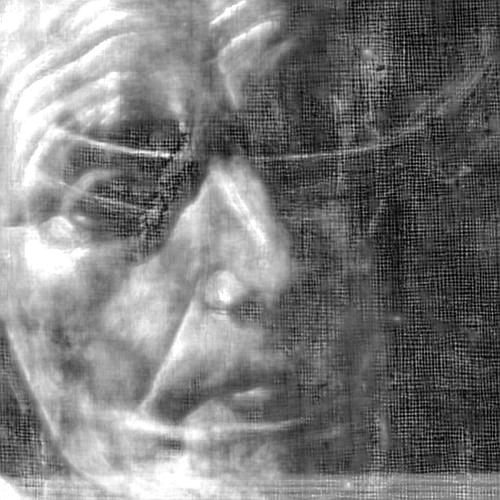}}
    \subfigure{\includegraphics[width=0.115\textwidth]{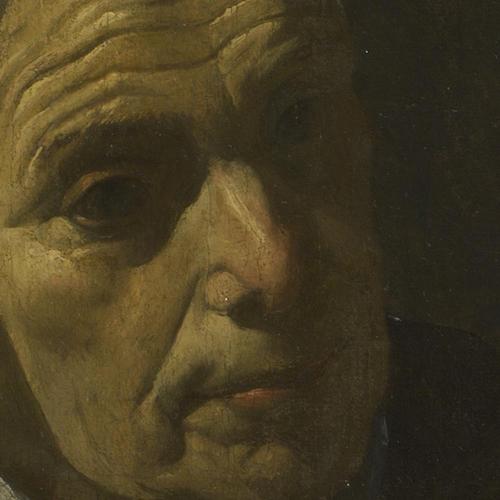}}
    \subfigure{\includegraphics[width=0.115\textwidth]{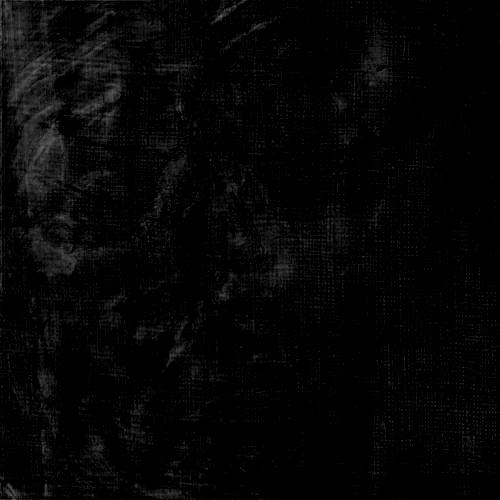}}
    \subfigure{\includegraphics[width=0.115\textwidth]{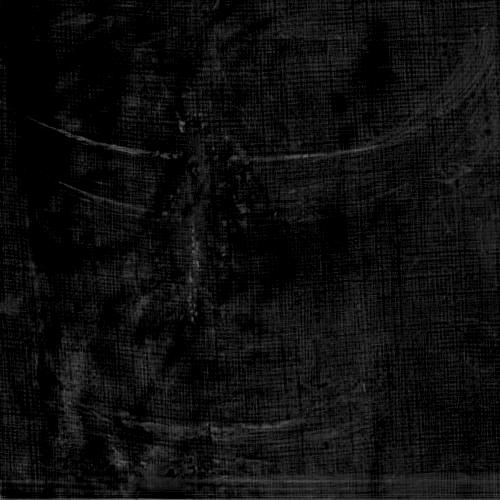}}
    \subfigure{\includegraphics[width=0.115\textwidth]{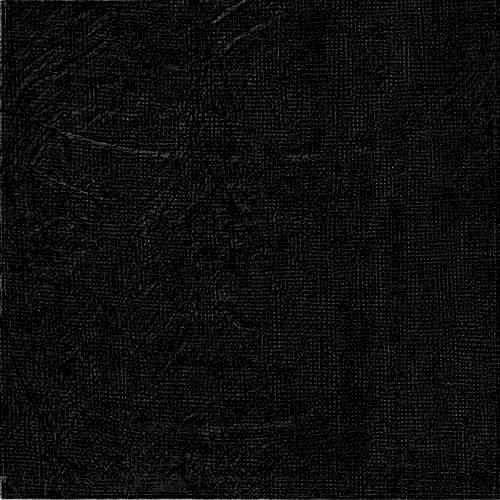}}
    \subfigure{\includegraphics[width=0.115\textwidth]{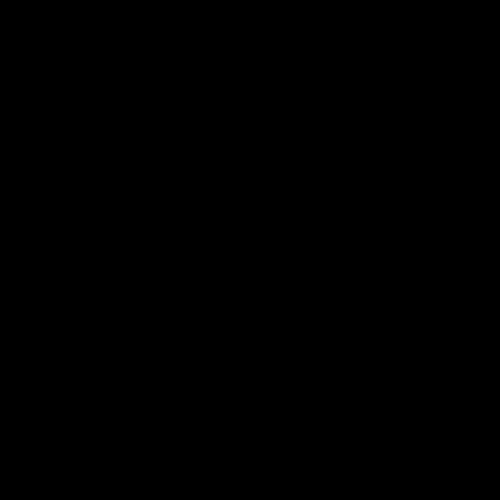}}
\caption{Reconstructed images under different cases. Rows 1 to 4 correspond to cases I to IV, respectively. Columns 1 to 2 correspond to the X-ray image of the `surface painting' and `concealed design', respectively. Column 3 corresponds to the synthetically mixed X-ray images from the separation results in columns 1 and 2. Column 4 corresponds to the reconstructed RGB image of `surface painting'. Columns 5 to 8 correspond to the error maps of the X-ray images associated with the `surface painting', `concealed design', synthetically mixed X-ray image and RGB image of the surface painting.}\label{F-4-P-2-2}
\end{figure*}

In Fig. \ref{F-4-P-2-2}, using the synthetic data shown in Fig. \ref{F-4-P-1-1}, we illustrate the resulting reconstructed images for various values of $\eta_1$ and $\eta_2$, with cases I to IV corresponding to the situations:
\begin{itemize}
\item Case I: Hyper-parameter associated with reconstruction loss of RGB image of the surface painting being dominant, \textit{i.e.}, $\eta_1 = 1$ and $\eta_2 =0.1$;

\item Case II: Hyper-parameter associated with reconstruction loss of the mixed X-ray image being dominant, \textit{i.e.}, $\eta_1 = 0$ and $\eta_2 =0$;

\item Case III: Hyper-parameter associated with exclusion loss being dominant, \textit{i.e.}, $\eta_1 = 0.5$ and $\eta_2 =0.4$;

\item Case IV: The optimized hyper-parameters are chosen.
\end{itemize}

From Fig. \ref{F-4-P-2-1} and \ref{F-4-P-2-2}, we can make the following observations:
\begin{itemize}
\item In case I, the loss function component of the reconstruction error of the RGB image of the `surface painting' dominates over other components implying one tends to promote fidelity of the reconstruction of the individual RGB images. Fig. \ref{F-4-P-2-1} suggests that this may result in a relatively high separation MSE of the separated X-ray images of both the `surface painting' and `concealed design', and Fig. \ref{F-4-P-2-2} also confirms that the error of the separated X-ray images is large. 

\item {\color{black}In case II, the loss function component of the reconstruction error of the synthetically mixed X-ray image dominates over other components implying one tends to promote fidelity of the reconstruction of the mixed X-ray image. Fig. \ref{F-4-P-2-1} also suggests that this may result in a relatively high separation MSE and Fig. \ref{F-4-P-2-2} again confirms that the error of the separated X-ray images is large. The separated X-ray image of the `concealed design' is very similar to the original mixed X-ray image and that of the `surface painting' is like a very high contrast version of the corresponding grayscale version of the visible RGB image. It has thus lost information present in the actual X-ray image associated with features that are not apparent in the visible RGB image of the surface painting such as structural features associated with the support (\textit{e.g.} the canvas weave or stretcher bars and keys) or features corresponding to areas of damage. }

\item In case III, the exclusion loss dominates over other components implying that one tends to promote the dis-similarity between the separated X-ray images for the `surface painting' and the `concealed design'. Fig. \ref{F-4-P-2-1} also suggests that this may result in a relatively high separation MSE and Fig. \ref{F-4-P-2-2} also confirms that the separated `concealed design' loses detail information in the left side.

\item In case IV, the optimized parameters determined from Fig. \ref{F-4-P-2-1} are chosen implying one tends to obtain the optimal separated individual X-ray images with the smallest MSE errors across all of the reconstructed images when compared with the ground truth.
\end{itemize}

\subsection{Experiments with Synthetically Mixed X-ray Data}
\subsubsection{Set-up}

\begin{figure}[t]
    \centering
    \subfigure[]{\includegraphics[width=0.0925\textwidth]{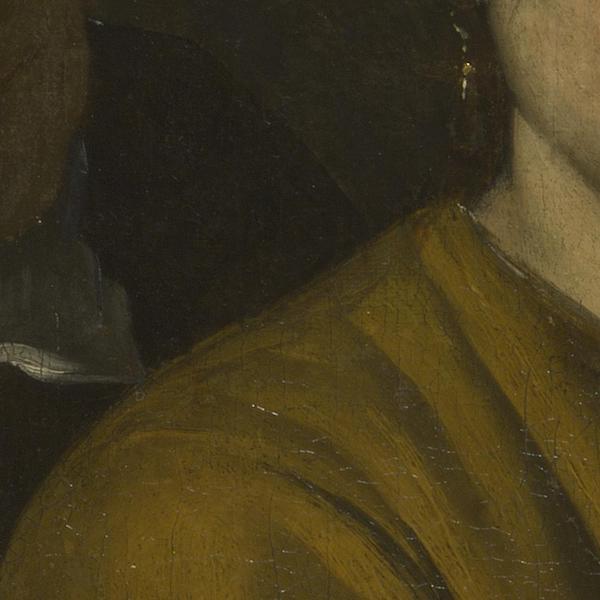}}
    \subfigure[]{\includegraphics[width=0.0925\textwidth]{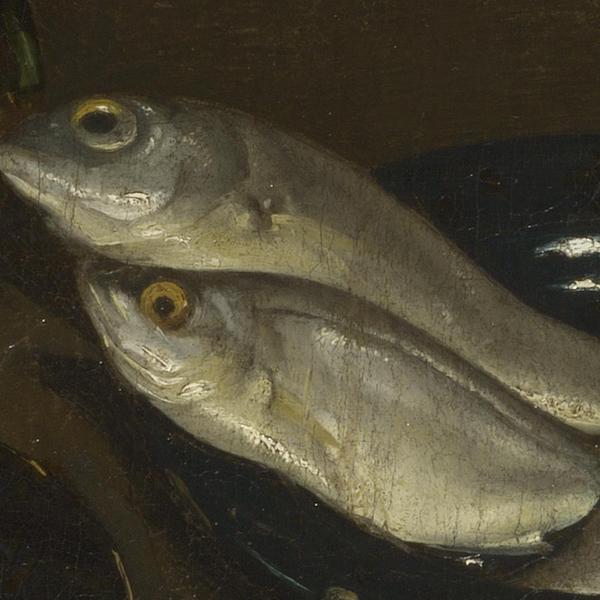}}
    \subfigure[]{\includegraphics[width=0.0925\textwidth]{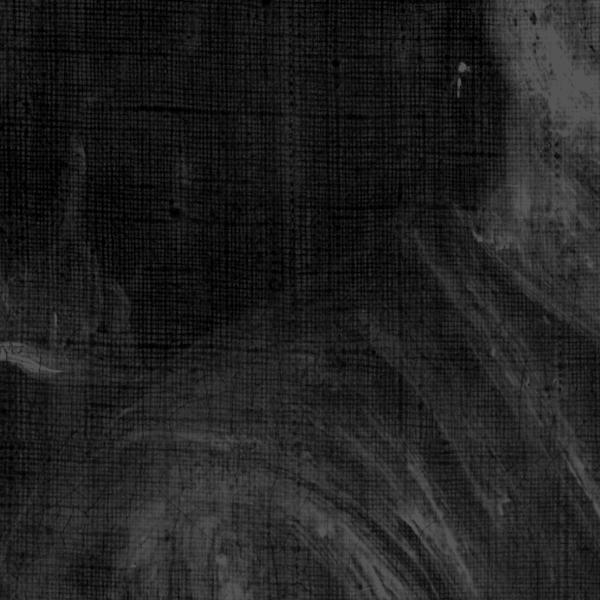}}
    \subfigure[]{\includegraphics[width=0.0925\textwidth]{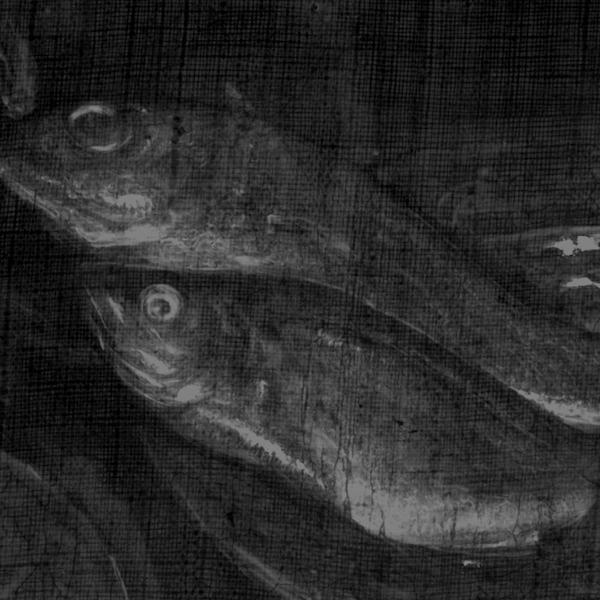}}
    \subfigure[]{\includegraphics[width=0.0925\textwidth]{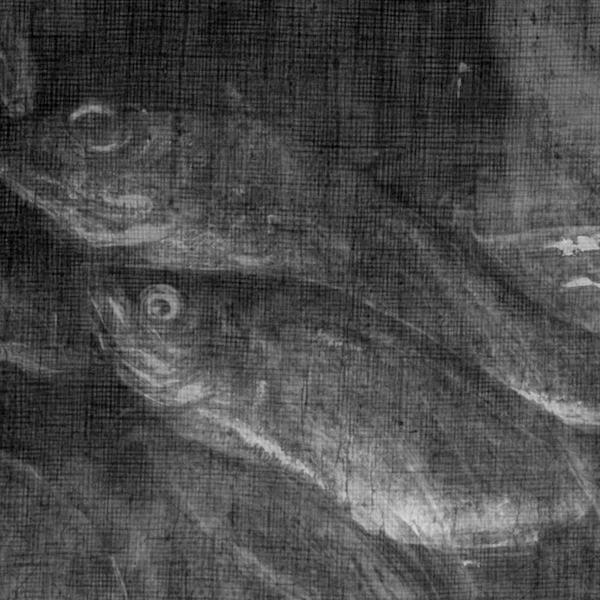}}
    \caption{Images used for experiments with synthetic mixed X-ray data. (a) First RGB image (the `surface image'); (b) second RGB image (the `concealed design'); (c) X-ray image corresponding to the `surface image'; (d) X-ray image corresponding to `concealed design'; (e) Synthetically mixed X-ray image. All images are details from the painting shown in Fig. \ref{F-4-1-1}.}\label{F-4-S-1-1}
\end{figure}

In these experiments, we used another two small areas with the same size from the same painting, corresponding to the young lady's shoulder and the fish on the table, to create a synthetically mixed X-ray image (see Fig. \ref{F-4-S-1-1}). The images – which are of size $500 \times 500$ pixels – were divided into patches of size $50 \times 50$ pixels with 45 pixels overlap (both in the horizontal and vertical direction), resulting in 3,136 patches. The patches associated with the synthetically mixed X-ray were then separated independently. The various patches associated with the individual separated X-rays were finally put together by placing the patches in their original order and averaging the overlapping portions. All patches were utilized in the training of the separation network by randomly shuffling their order. 

As  mentioned  previously, we adopted the optimized hyper-parameter values $\eta_1 = 0.5 $ and $\eta_2 =0.1$. The first area, the shoulder of the young woman (Fig. \ref{F-4-S-1-1} (a) and (c)), is set to be the `surface painting', while the second area, the fish on the table (Fig. \ref{F-4-S-1-1} (b) and (d)), is set to be the `concealed design'.

\subsubsection{Results}

\begin{figure*}[t]
\centering
    \subfigure[]{\includegraphics[width=0.225\textwidth]{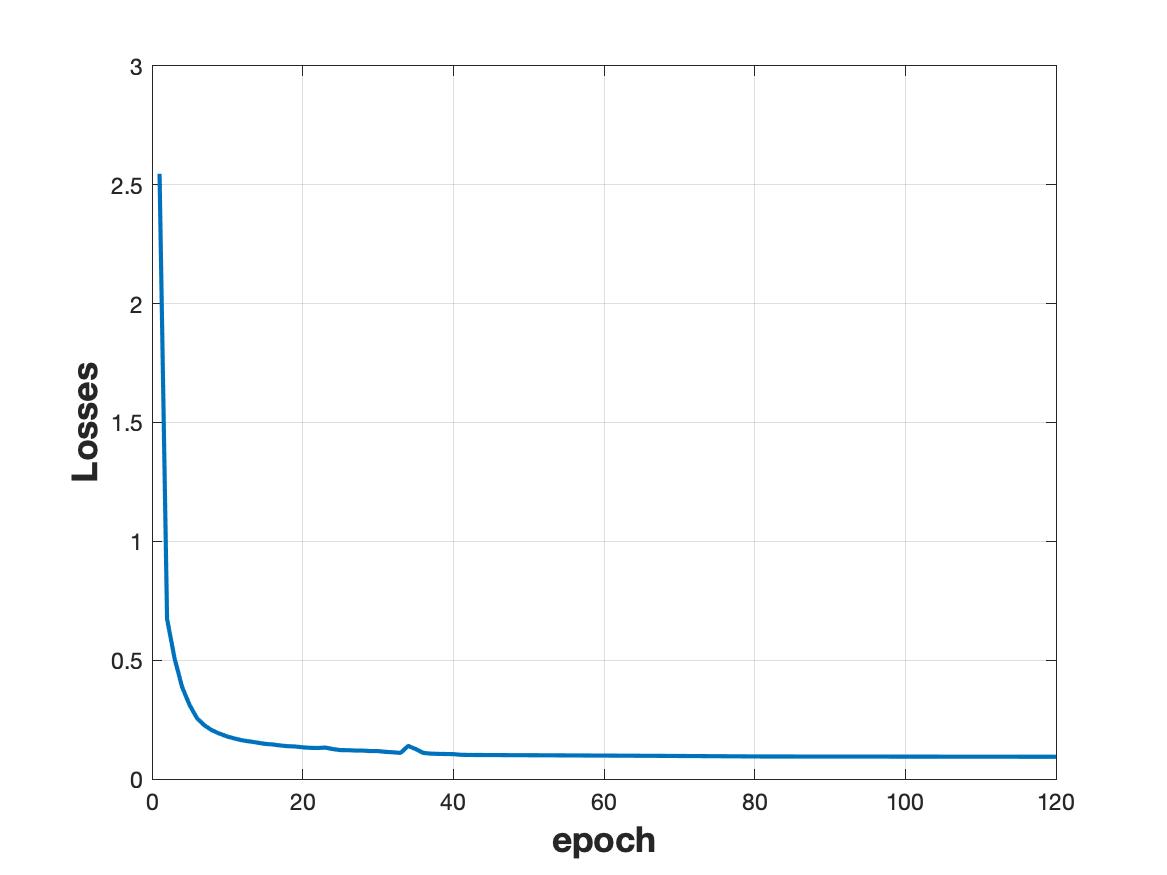}}
    \subfigure[]{\includegraphics[width=0.225\textwidth]{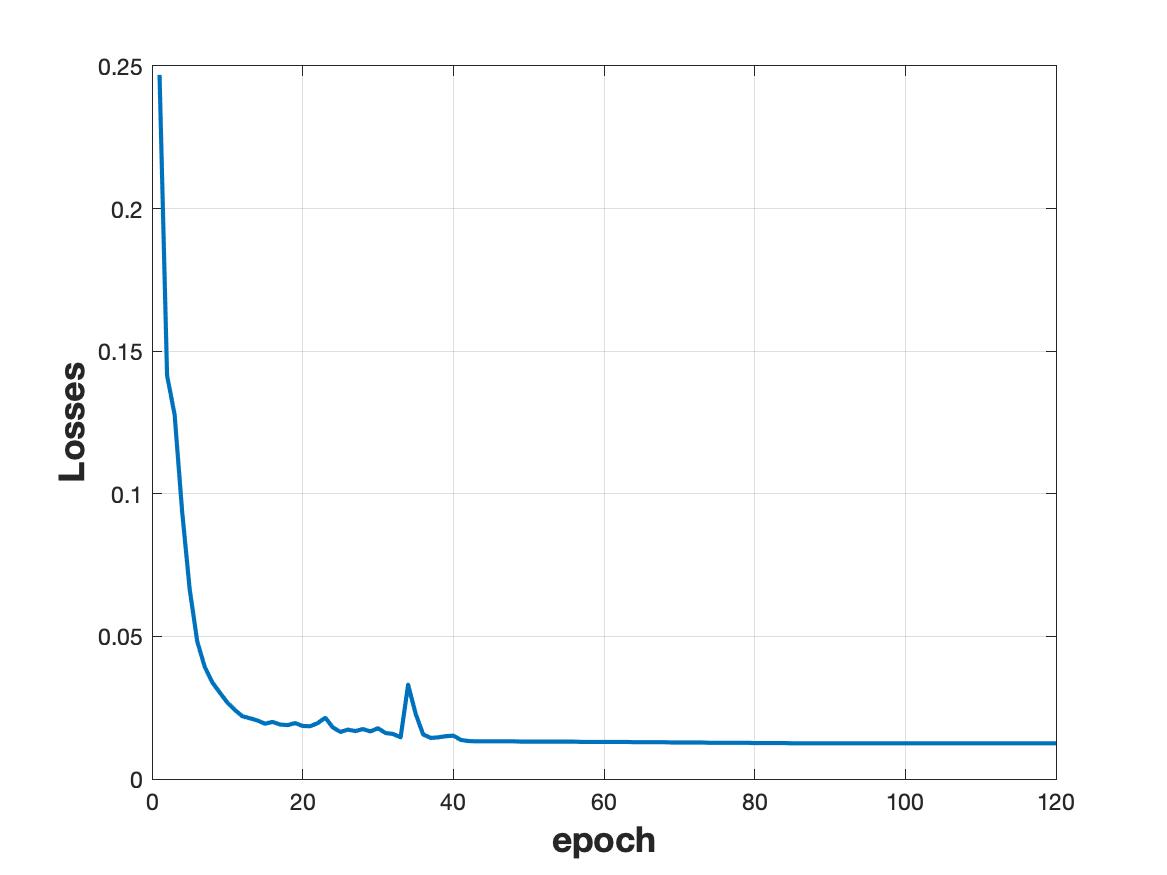}}
    \subfigure[]{\includegraphics[width=0.225\textwidth]{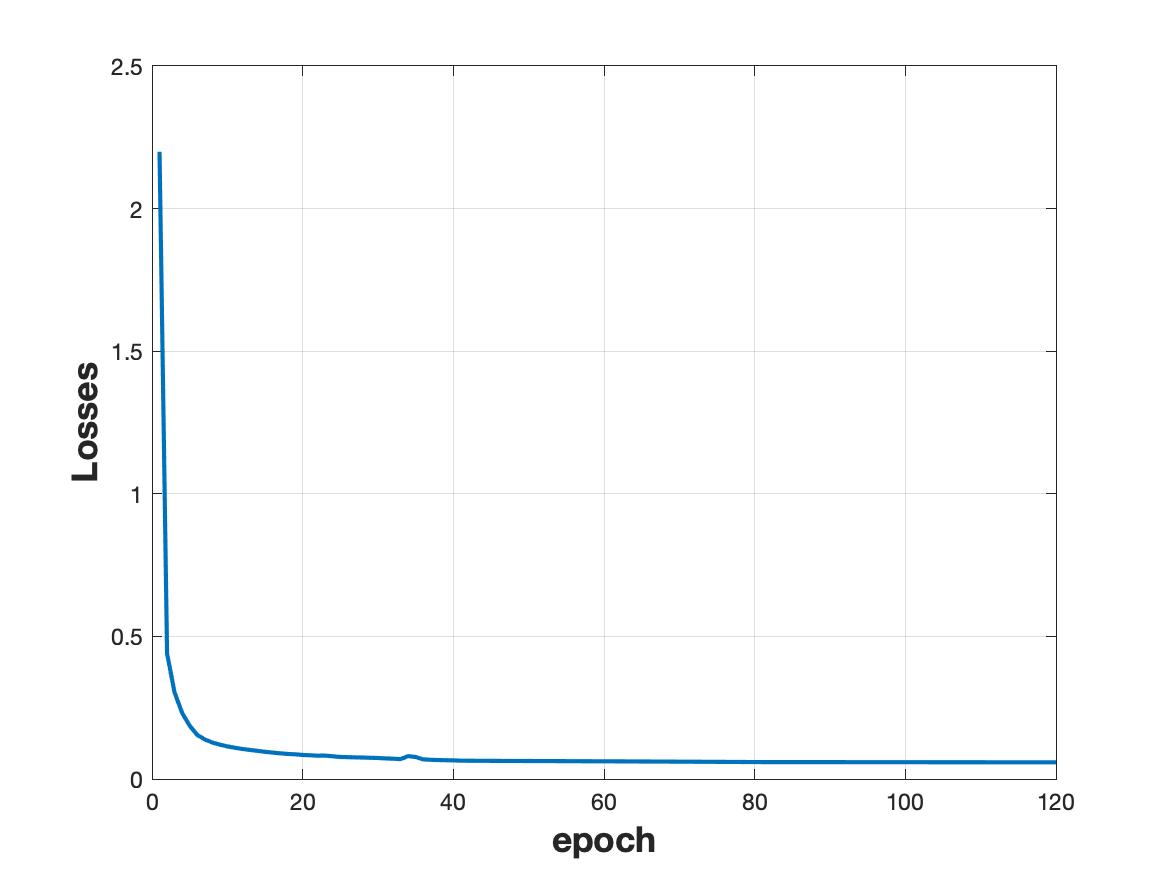}}
    \subfigure[]{\includegraphics[width=0.225\textwidth]{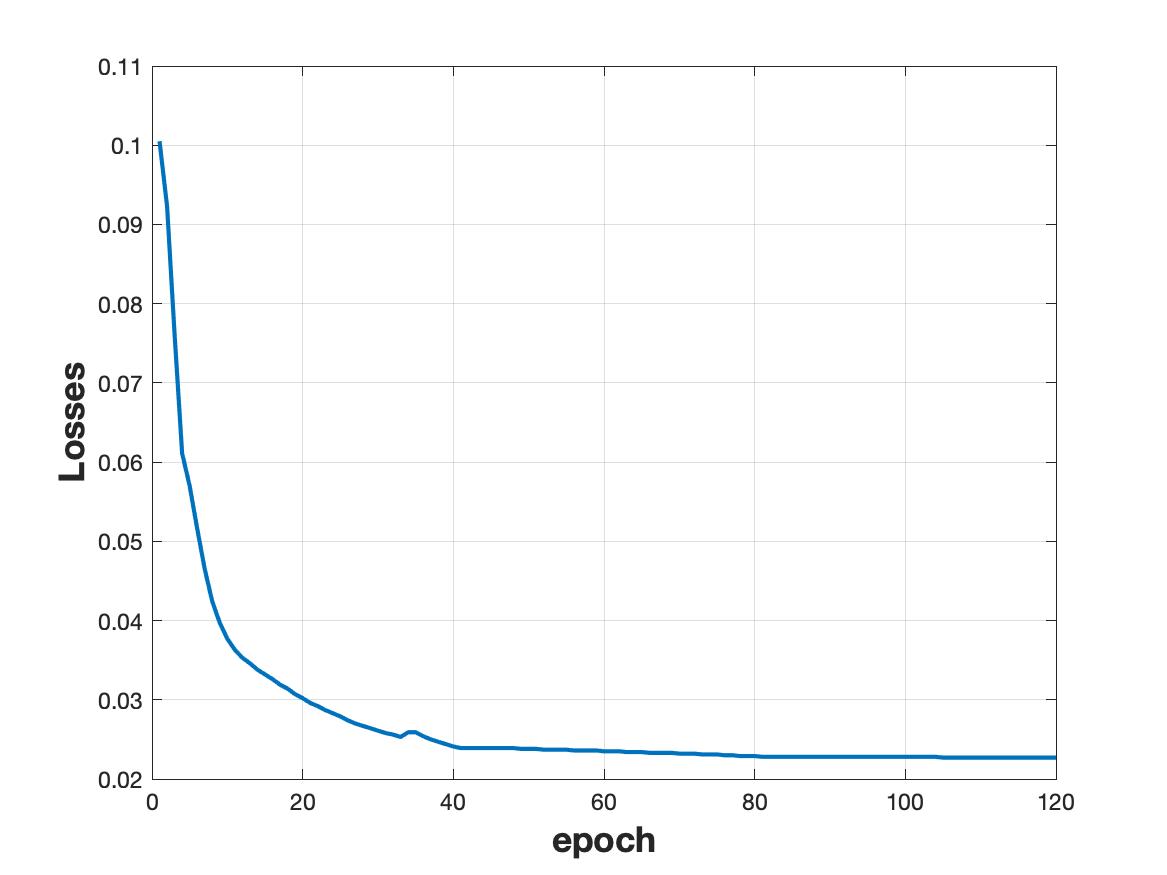}}
\caption{Losses vs. number of epochs on synthetic data. (a). Total loss. (b). Reconstruction loss of the RGB image. (c). Reconstruction loss of the mixed X-ray. (d). Exclusion loss.}\label{F-4-2-2-1}
\end{figure*}

\begin{figure}[t]
\centering
    \subfigure{\includegraphics[width=0.0925\textwidth]{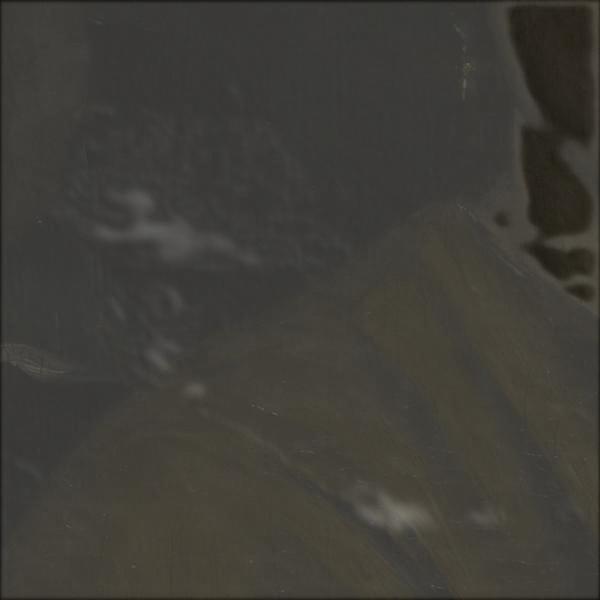}}
    \subfigure{\includegraphics[width=0.0925\textwidth]{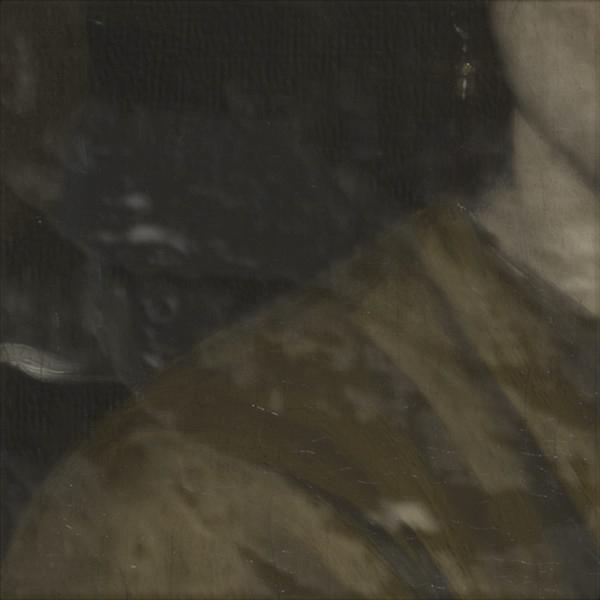}}
    \subfigure{\includegraphics[width=0.0925\textwidth]{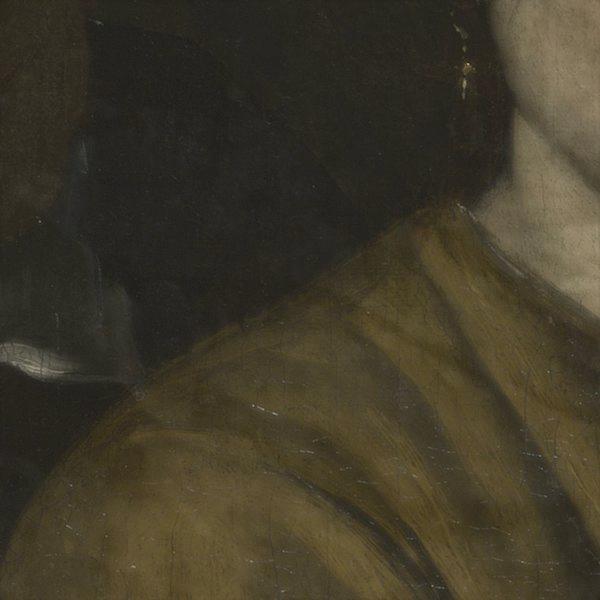}}
    \subfigure{\includegraphics[width=0.0925\textwidth]{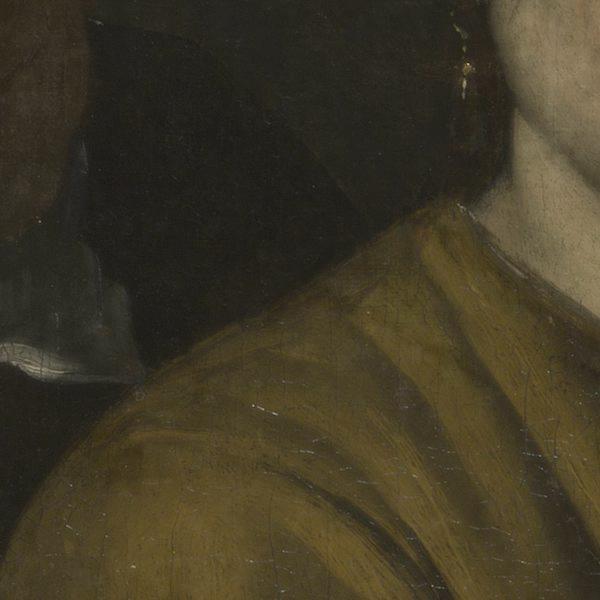}}
    \subfigure{\includegraphics[width=0.0925\textwidth]{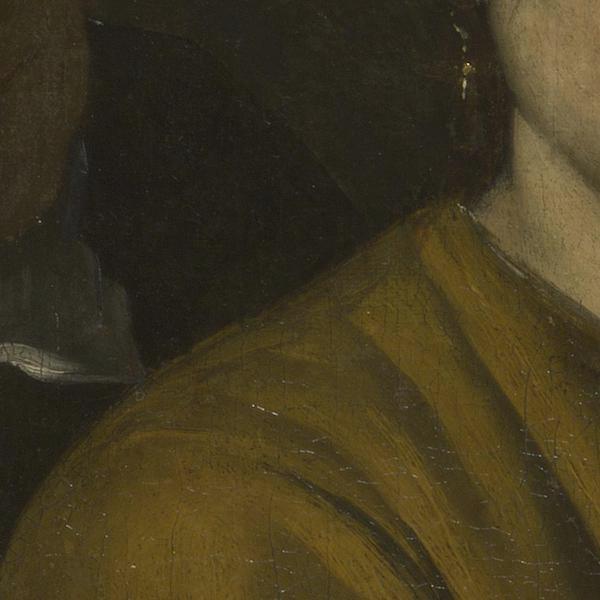}}

    \subfigure{\includegraphics[width=0.0925\textwidth]{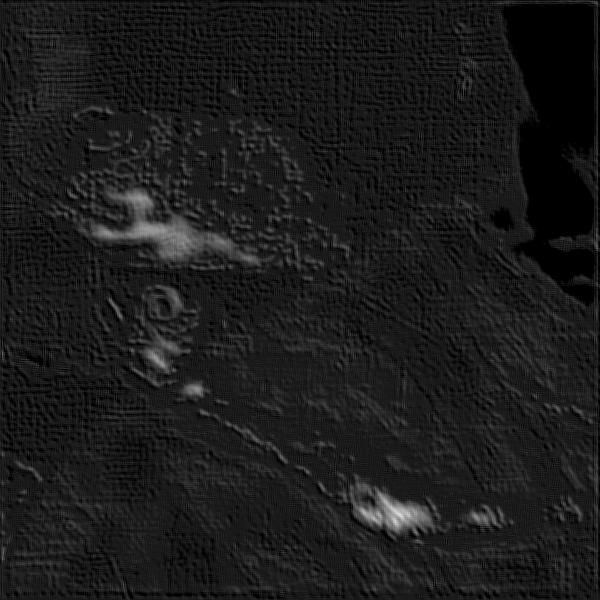}}
    \subfigure{\includegraphics[width=0.0925\textwidth]{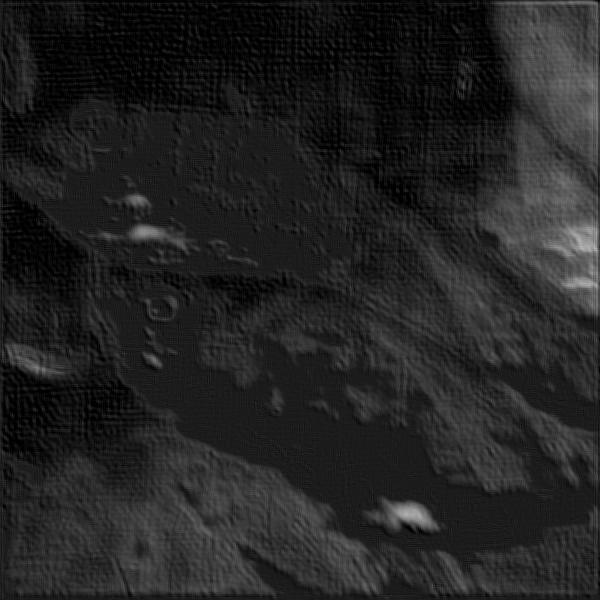}}
    \subfigure{\includegraphics[width=0.0925\textwidth]{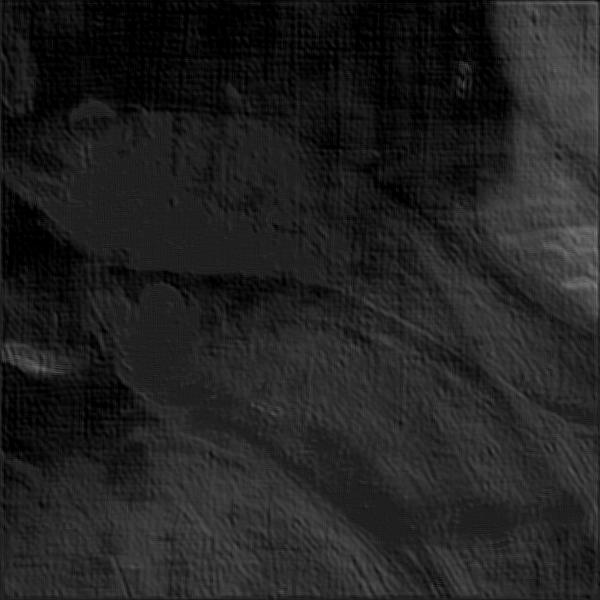}}
    \subfigure{\includegraphics[width=0.0925\textwidth]{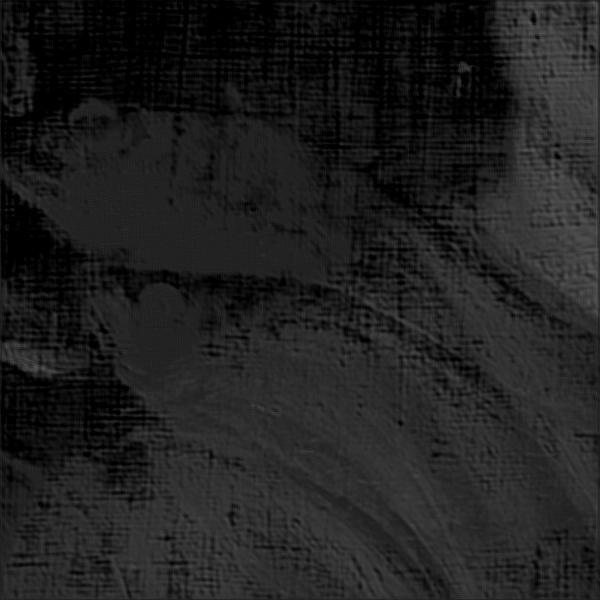}}
    \subfigure{\includegraphics[width=0.0925\textwidth]{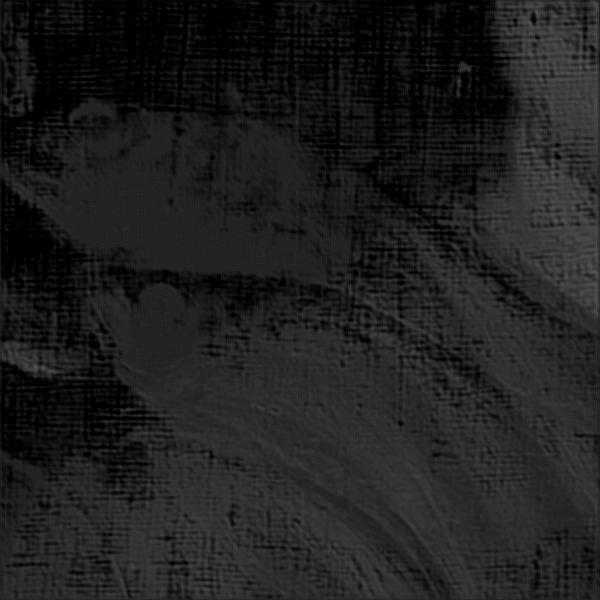}}

    \subfigure{\includegraphics[width=0.0925\textwidth]{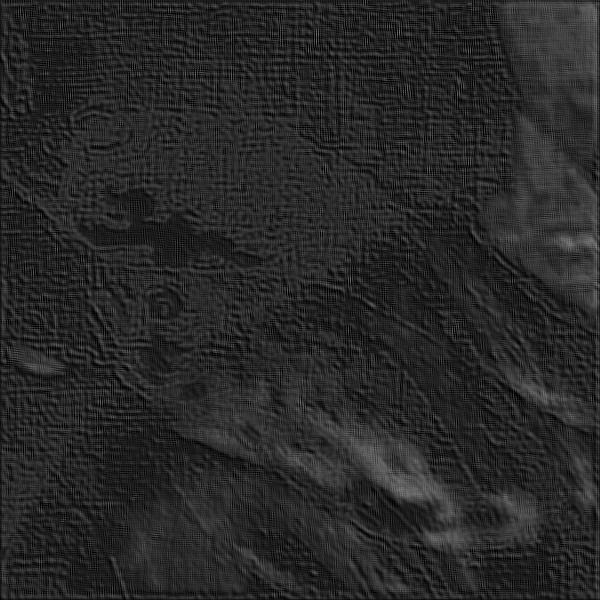}}
    \subfigure{\includegraphics[width=0.0925\textwidth]{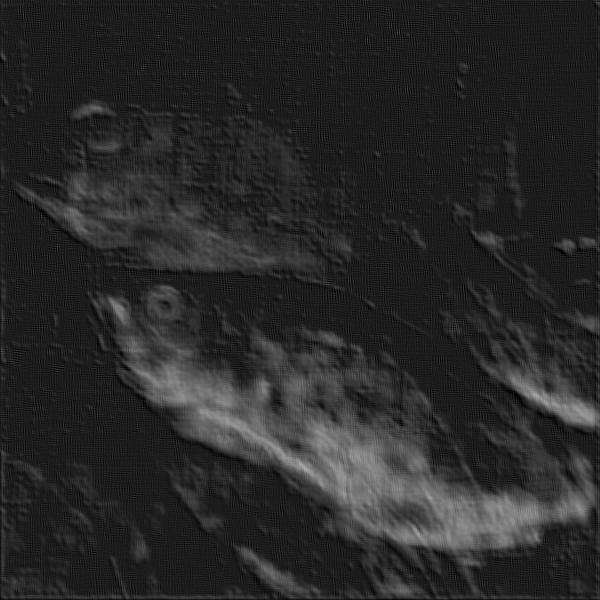}}
    \subfigure{\includegraphics[width=0.0925\textwidth]{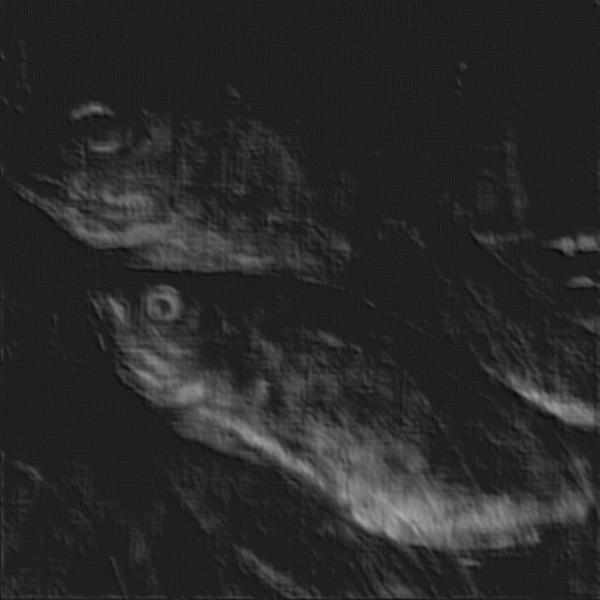}}
    \subfigure{\includegraphics[width=0.0925\textwidth]{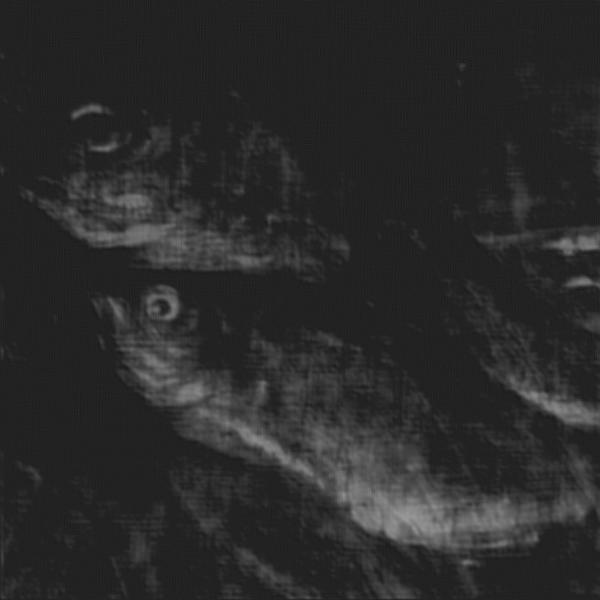}}
    \subfigure{\includegraphics[width=0.0925\textwidth]{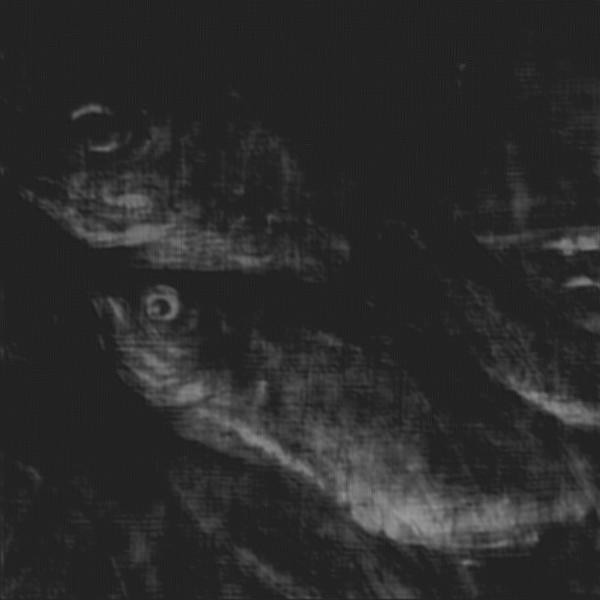}}

    \subfigure{\includegraphics[width=0.0925\textwidth]{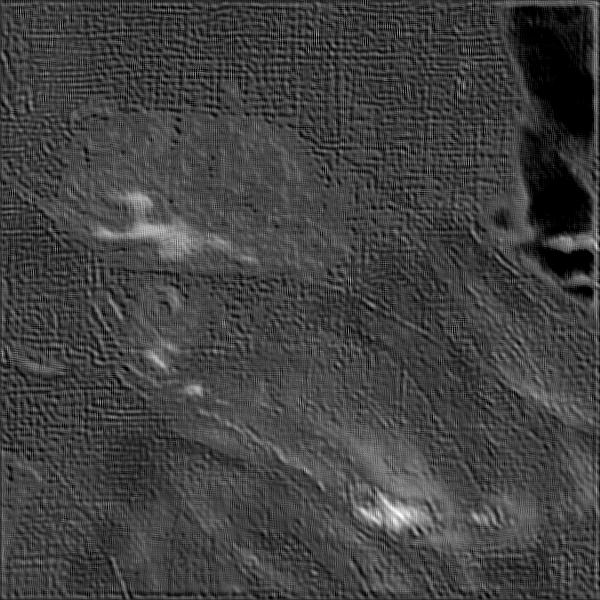}}
    \subfigure{\includegraphics[width=0.0925\textwidth]{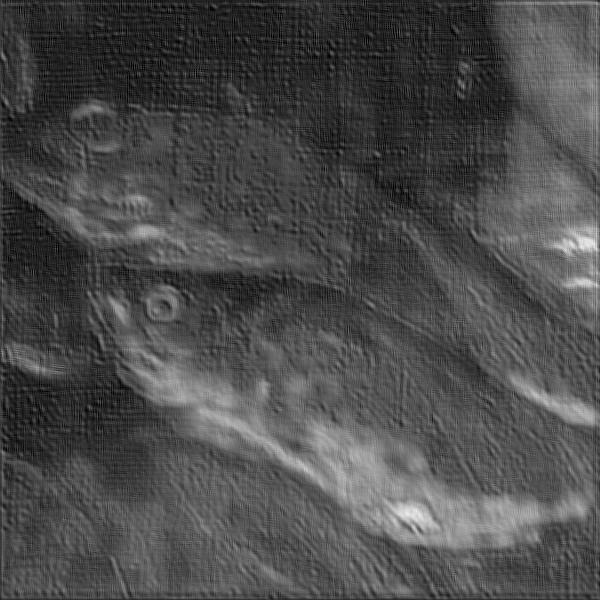}}
    \subfigure{\includegraphics[width=0.0925\textwidth]{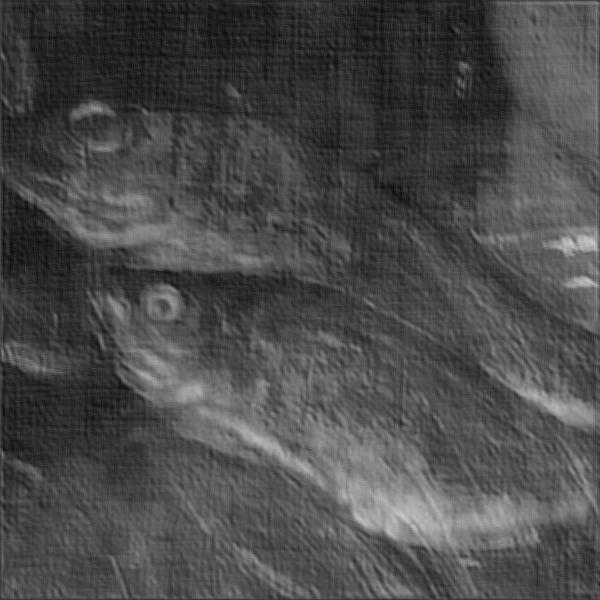}}
    \subfigure{\includegraphics[width=0.0925\textwidth]{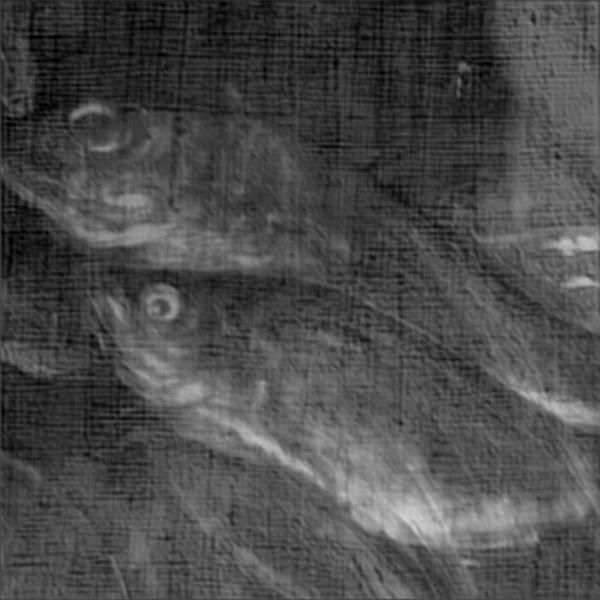}}
    \subfigure{\includegraphics[width=0.0925\textwidth]{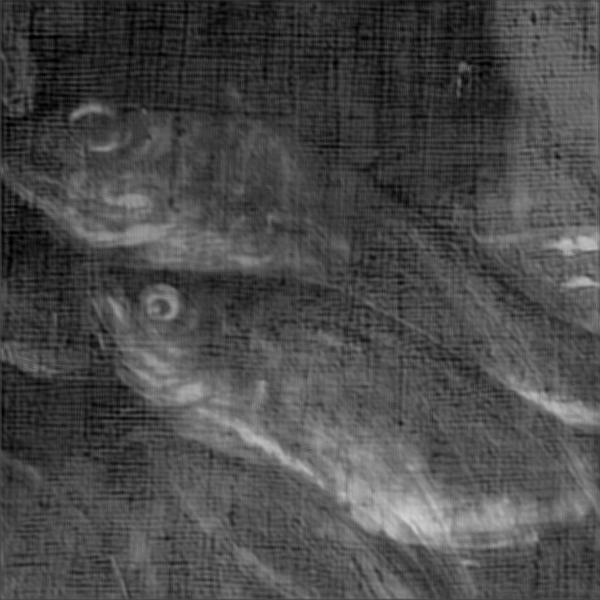}}
\caption{Reconstructed images vs. number of epochs for synthetic data experiments. Columns 1 to 5 correspond to the reconstructed result after 1st, 4th, 10th, 40th, and 120th epoch, respectively. Row 1 represents the reconstructed RGB image of the `surface painting'. Rows 2 and 3 correspond to the reconstructed X-ray images of the `surface painting' and `concealed design', respectively. Row 4 represents the synthetically mixed X-ray image.}\label{F-4-2-2-2}
\end{figure}

The plots shown in Fig. \ref{F-4-2-2-1} depicts the evolution of the overall loss function along with the individual ones as a function of the number of epochs. Fig. \ref{F-4-2-2-2} illustrates results of the proposed network for different training epochs. Various trends can be observed:
\begin{itemize}
\item The overall loss function gradually decreases as the number of epochs increases. This suggests our method will eventually reconstruct the RGB image of the `surface painting', the synthetically mixed X-ray image, and – as a by-product – the individual X-ray images (see Fig. \ref{F-4-2-2-2}).

\item The reconstruction losses of the mixed X-ray image and RGB image of the `surface painting' decrease very rapidly during the initial 5 epochs but decrease less dramatically then onwards. This implies that the method can reconstruct very well the mixed X-ray image and RGB image of the `surface painting' during this initial phase (see Fig. \ref{F-4-2-2-2} row 1).

\item The exclusion loss decreases during the initial 40 epochs and does not change much in the last 80 epochs. From Fig. \ref{F-4-2-2-2}, we can observe that as the exclusion loss decreases in the initial 40 epochs, the information relating to the surface painting, including the shoulder, neck and chin, disappears from the separated X-ray image for the `concealed painting'.
\end{itemize}

\begin{figure}[t]
\centering
    \subfigure{\includegraphics[width=0.0925\textwidth]{S_x1.jpg}}
    \subfigure{\includegraphics[width=0.0925\textwidth]{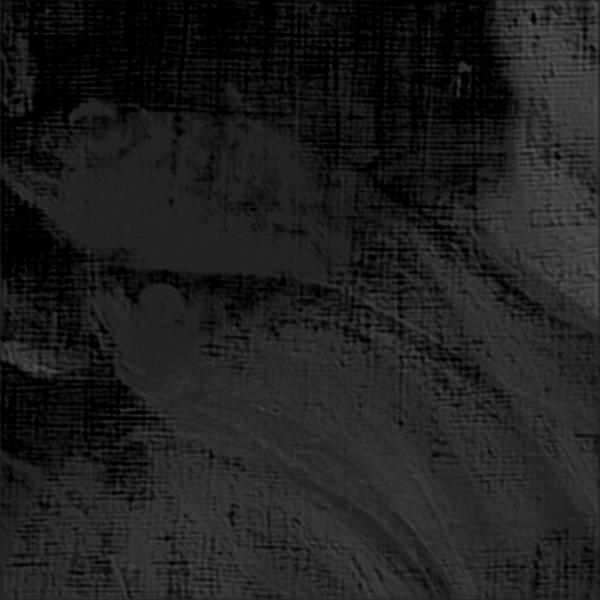}}
    \subfigure{\includegraphics[width=0.0925\textwidth]{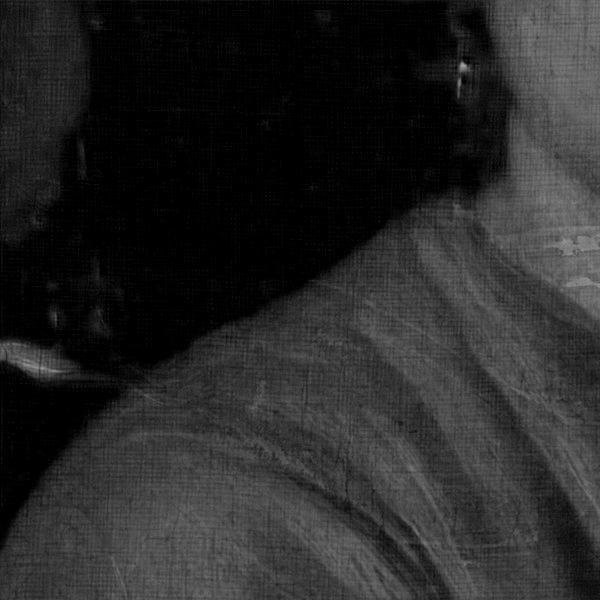}}
    \subfigure{\includegraphics[width=0.0925\textwidth]{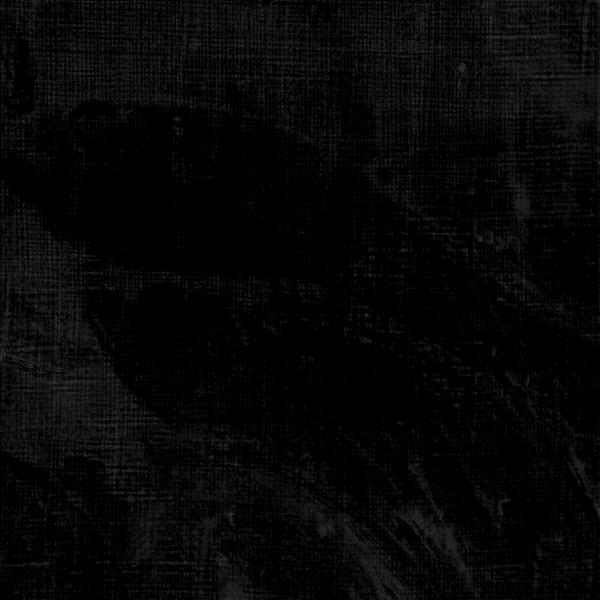}}
    \subfigure{\includegraphics[width=0.0925\textwidth]{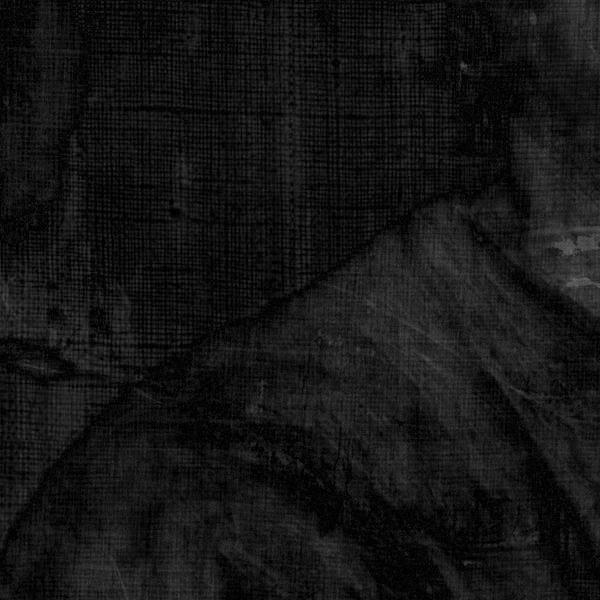}}
    
    \subfigure{\includegraphics[width=0.0925\textwidth]{S_x2.jpg}}
    \subfigure{\includegraphics[width=0.0925\textwidth]{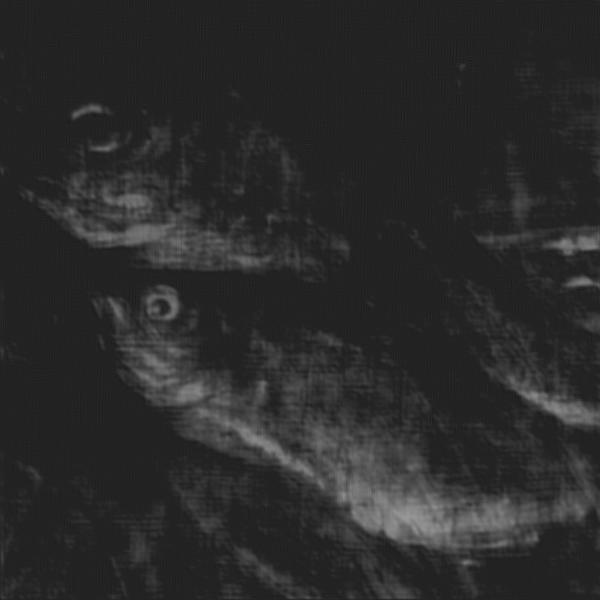}}
    \subfigure{\includegraphics[width=0.0925\textwidth]{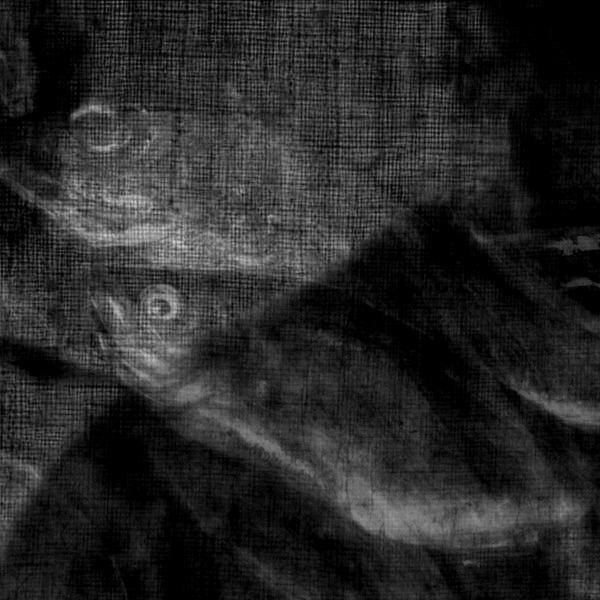}}
    \subfigure{\includegraphics[width=0.0925\textwidth]{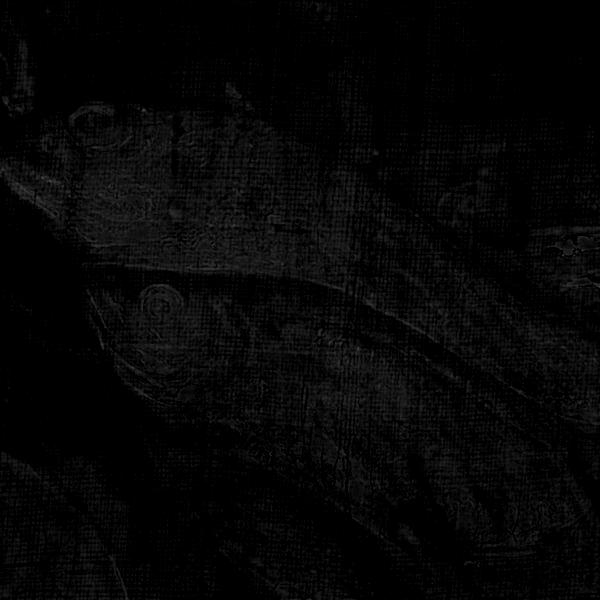}}
    \subfigure{\includegraphics[width=0.0925\textwidth]{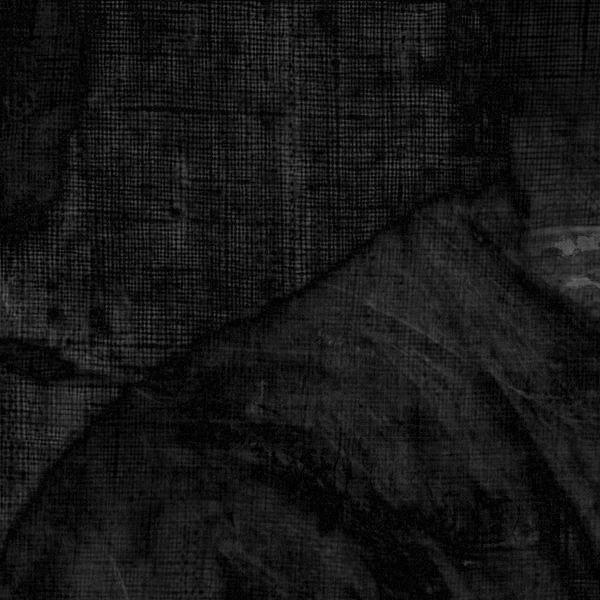}}
    
    \subfigure{\includegraphics[width=0.0925\textwidth]{S_x.jpg}}
    \subfigure{\includegraphics[width=0.0925\textwidth]{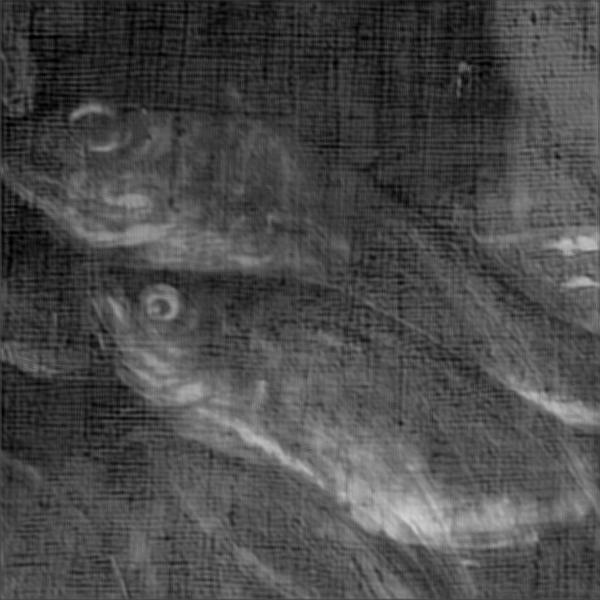}}
    \subfigure{\includegraphics[width=0.0925\textwidth]{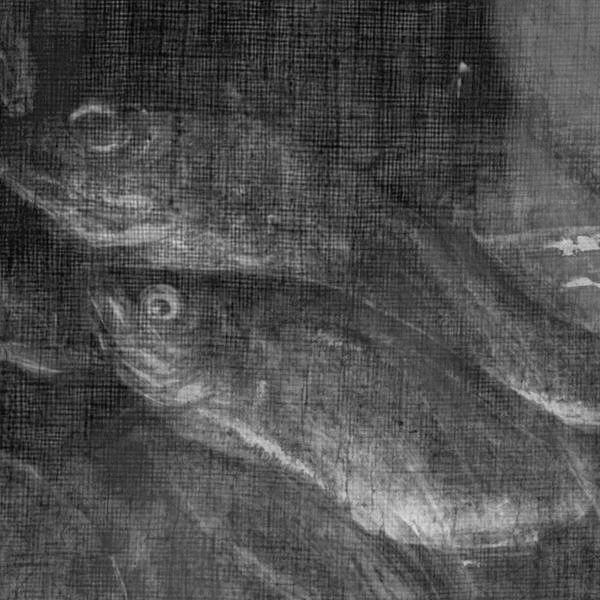}}
    \subfigure{\includegraphics[width=0.0925\textwidth]{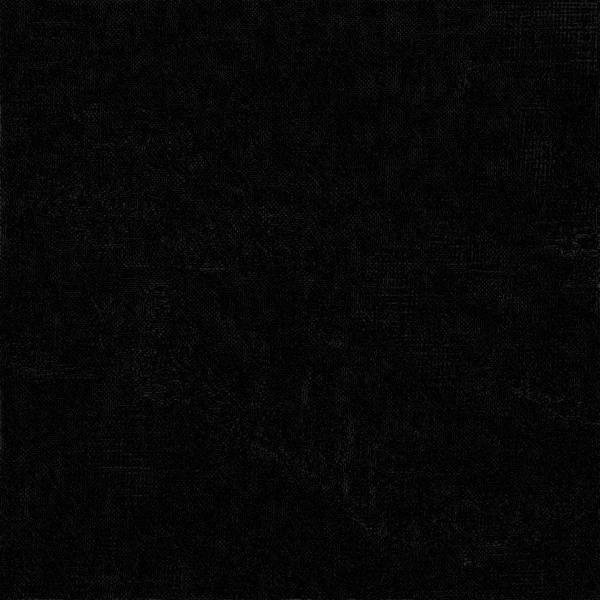}}
    \subfigure{\includegraphics[width=0.0925\textwidth]{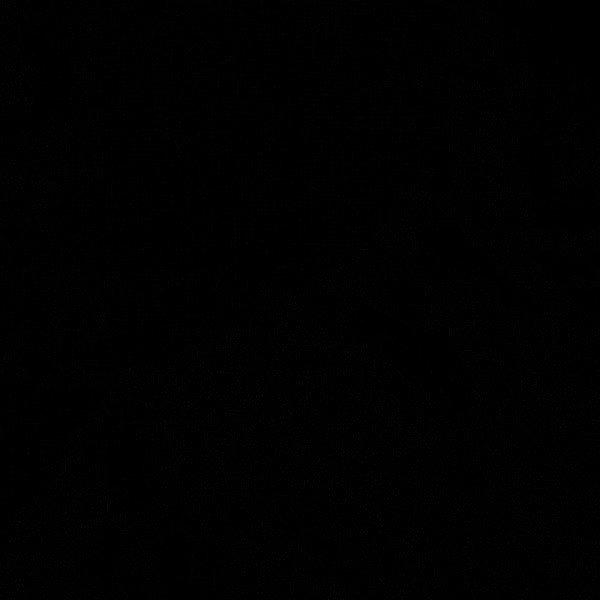}}
\caption{Comparison of separation results using synthetic data. Rows 1-3 correspond to the reconstructed X-ray images of the `surface painting' and `concealed design', and mixed X-ray image, respectively. Columns 1-3 correspond to  the ground truth, results using the proposed algorithm and using the algorithm in \cite{IS6}, respectively. Columns 4 and 5 correspond to the error maps of the results using the proposed approach and method in \cite{IS6}, respectively.}\label{F-4-2-4-1}
\end{figure}

{\color{black}
In Fig. \ref{F-4-2-4-1}, the results obtained using our new approach and those using the method in \cite{IS6} are compared. Using the method in \cite{IS6}, the separated X-ray image for the ‘surface painting’ shows too much similarity to the RGB image in regions (\textit{e.g.} very definite outline to the woman’s shoulder and neck and intensity from the face of the old woman in the upper left corner) that do not appear in the mixed X-ray image (nor are in the ground truth X-ray image). The increased intensity in areas of the X-ray image for the ‘surface painting’ results in negative features in the X-ray image of the ‘concealed design’ where, for example, the shoulder of the young woman appears strongly as a shadow. It should also be noted that much of detail of the canvas weave seen in the mixed X-ray image does not appear in the separated X-ray image for the ‘surface painting’. The results using the current approach appear to avoid this issue of overreliance on the RGB image (although now some features from the ‘concealed design’ are still apparent), and to give particularly impressive results for the separated X-ray image of the ‘concealed design’. 

Overall, the resulting using the current approach also have a lower MSE. In particular, the MSE associated with the reconstruction of the X-ray image of the ‘surface  painting’ is 0.0032 with our method and 0.0039 with the method in \cite{IS6}. The MSE associated with the reconstruction of the X-ray image of the ‘concealed design’ is 0.0052 with our method and 0.0058 with
the method in \cite{IS6}.}

\subsection{Experiments with Real Mixed X-ray Data based on a double-sided panel from the \textsl{Ghent Altarpiece}}
\subsubsection{Set-up}
\begin{figure}[t]
    \centering
    \subfigure[]{\includegraphics[width=0.0925\textwidth]{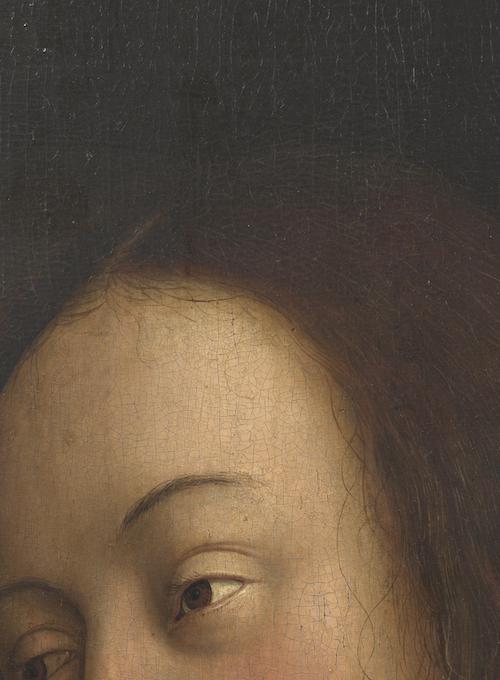}}
    \subfigure[]{\includegraphics[width=0.0925\textwidth]{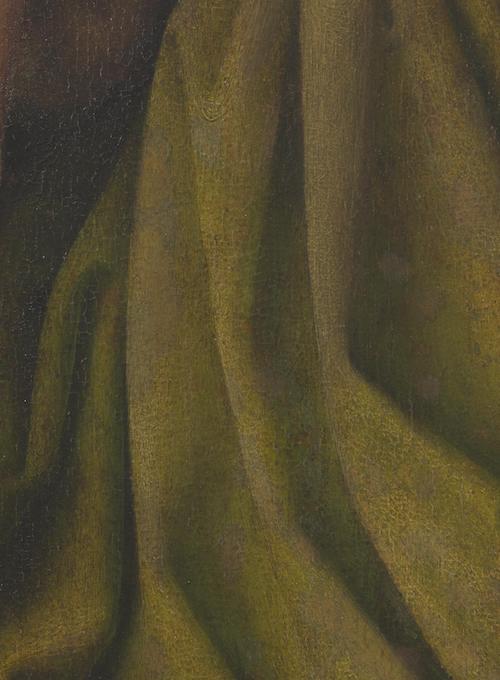}}
    \subfigure[]{\includegraphics[width=0.0925\textwidth]{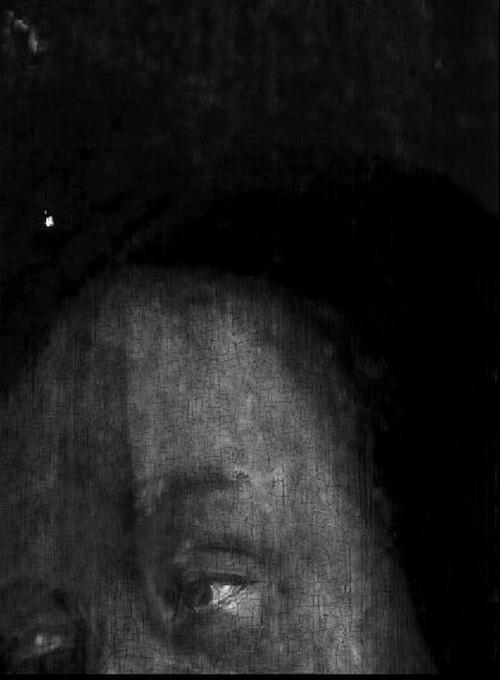}}
    \subfigure[]{\includegraphics[width=0.0925\textwidth]{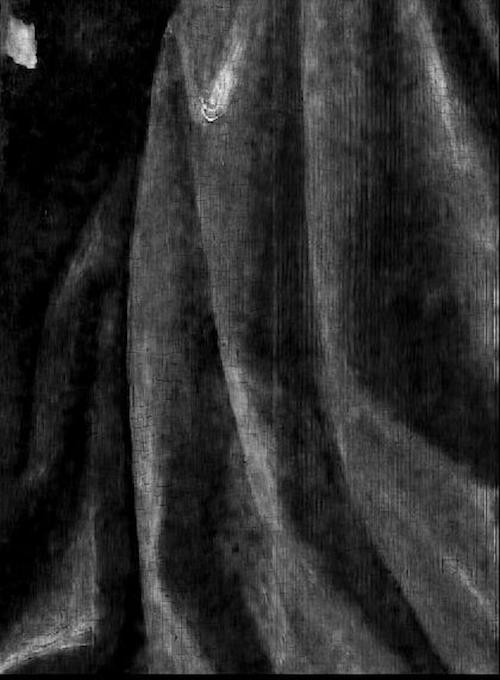}}
    \subfigure[]{\includegraphics[width=0.0925\textwidth]{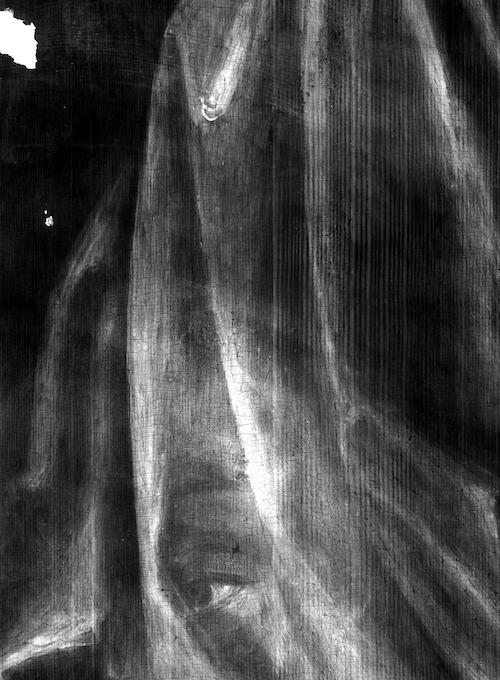}}
    \caption{Images from the \textsl{Ghent Altarpiece} (See Fig. \ref{Ghent}) used for real mixed X-ray image separation. (a). RGB image of the front side. (b). RGB image of the rear side. (c). Separated X-ray image of the front side by \cite{IS5}. (d). Separated X-ray image of the rear side by \cite{IS5}. (e). Mixed X-ray image.}\label{F-4-Gh-1-1}
\end{figure}

In this experiment, we use real mixed X-ray data to showcase the performance of the proposed X-ray image separation algorithm.
As shown in Fig. \ref{F-4-Gh-1-1}, we use one small area of size $680 \times 500$ pixels from one of the double-sided panels of the \textsl{Ghent Altarpiece} (see Fig. \ref{Ghent}). Note that \textsl{Ghent Altarpiece} is a double-sided painting. The RGB image of Eve's face in Fig. \ref{F-4-Gh-1-1} is set as the only available side information.
The previous procedure was again followed: the RGB image of Eve's face in Fig. \ref{F-4-Gh-1-1} and the mixed X-ray image were divided into 2,944 patches with size 50 $\times$ 50 pixels. Adjacent patches have 40 pixels overlap both in the horizontal and vertical directions. We adopted the hyper-parameter values $\eta_1 = 0.5 $ and $\eta_2 =0.1$.

\subsubsection{Results}

In the supplementary material, we show the evolution of the overall loss function along with the individual ones as a function of the number of epochs and the results of the proposed network for different training epochs on real data from the \textsl{Ghent Altarpiece}.

\begin{figure}[t]
\centering
    \subfigure{\includegraphics[width=0.0925\textwidth]{Ghent_x1.jpg}}
    \subfigure{\includegraphics[width=0.0925\textwidth]{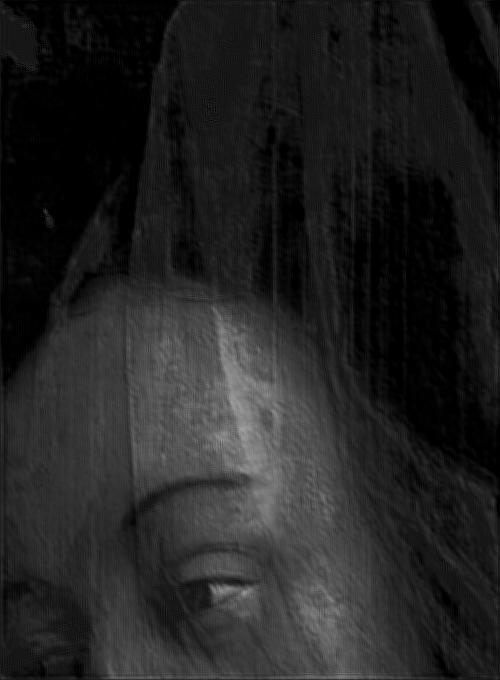}}
    \subfigure{\includegraphics[width=0.0925\textwidth]{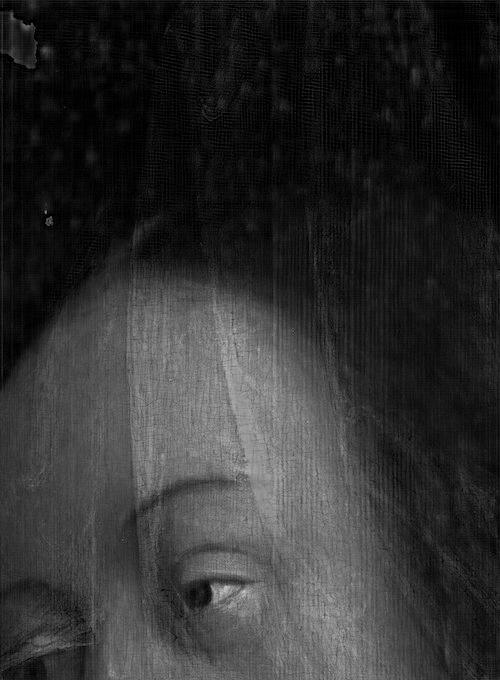}}
    \subfigure{\includegraphics[width=0.0925\textwidth]{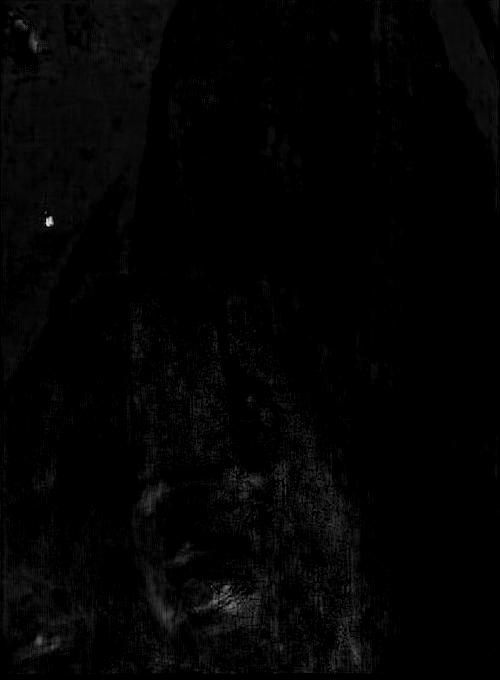}}
    \subfigure{\includegraphics[width=0.0925\textwidth]{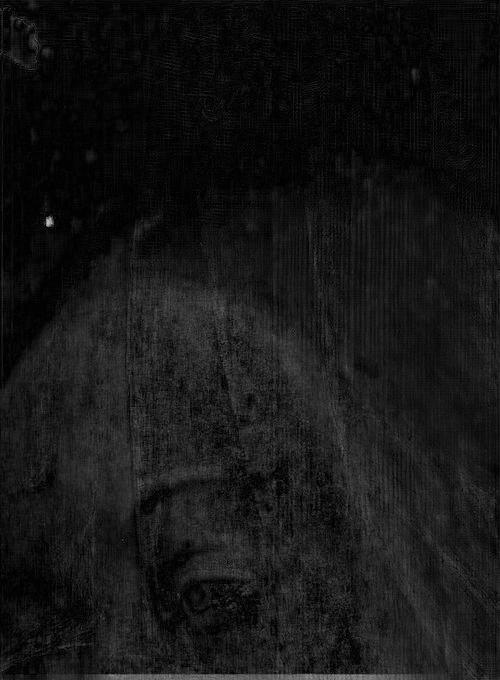}}
    
    \subfigure{\includegraphics[width=0.0925\textwidth]{Ghent_x2.jpg}}
    \subfigure{\includegraphics[width=0.0925\textwidth]{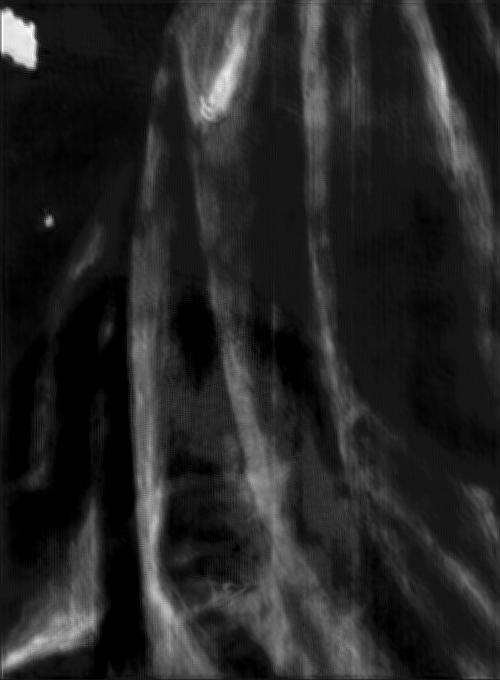}}
    \subfigure{\includegraphics[width=0.0925\textwidth]{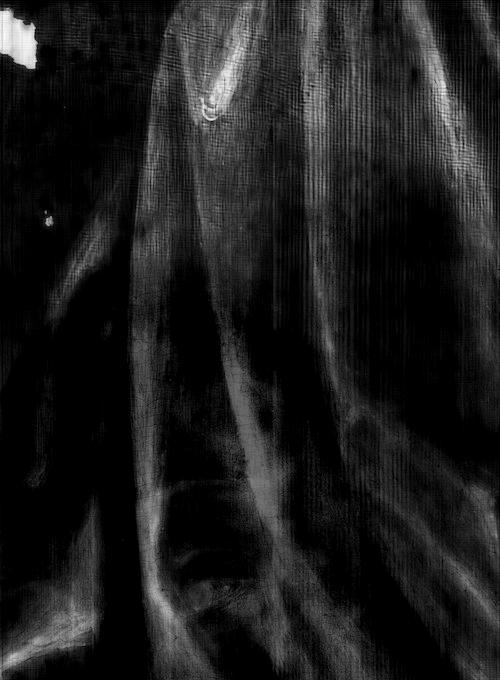}}
    \subfigure{\includegraphics[width=0.0925\textwidth]{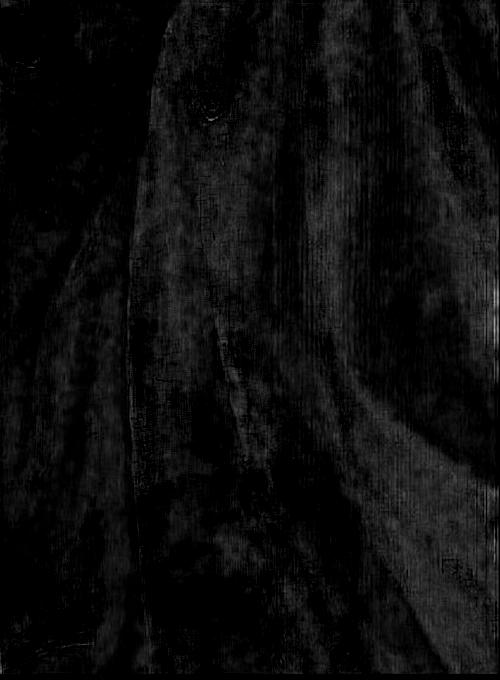}}
    \subfigure{\includegraphics[width=0.0925\textwidth]{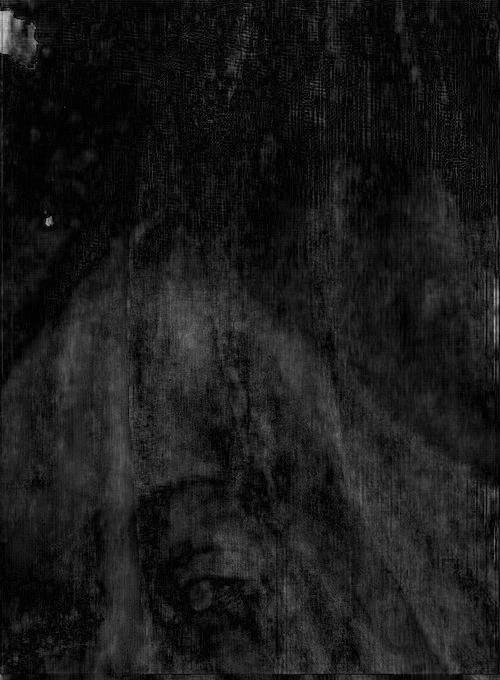}}
    
    \subfigure{\includegraphics[width=0.0925\textwidth]{Ghent_x.jpg}}
    \subfigure{\includegraphics[width=0.0925\textwidth]{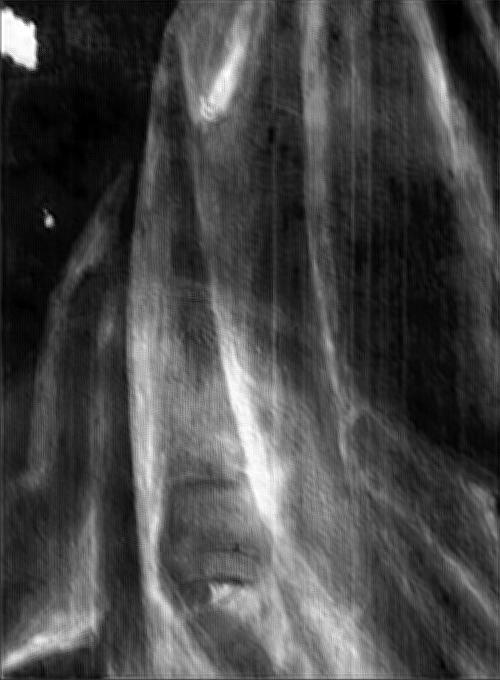}}
    \subfigure{\includegraphics[width=0.0925\textwidth]{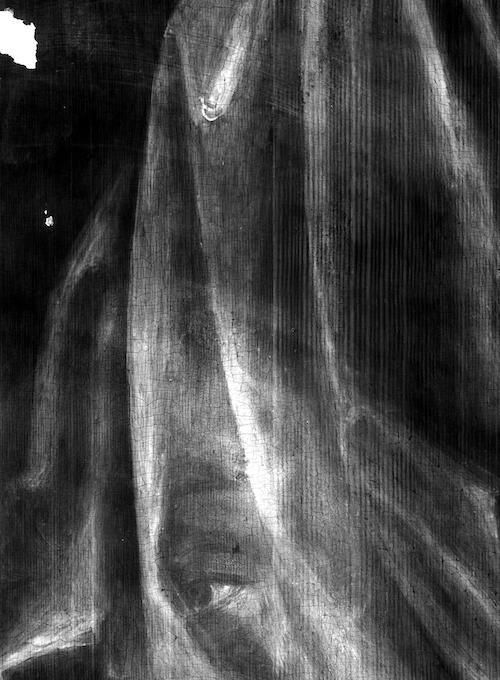}}
    \subfigure{\includegraphics[width=0.0925\textwidth]{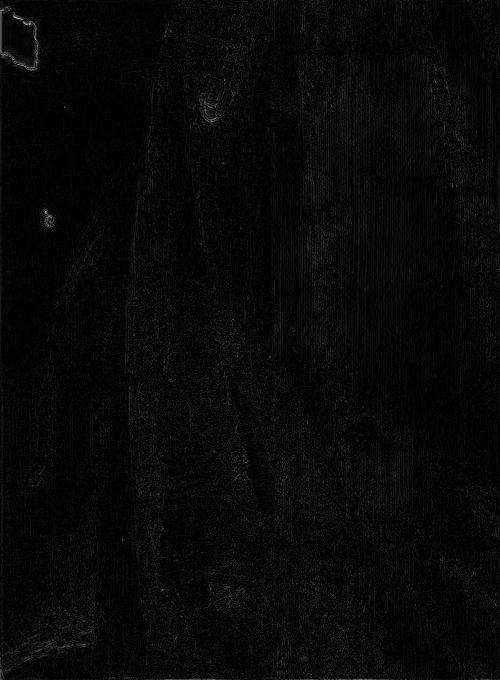}}
    \subfigure{\includegraphics[width=0.0925\textwidth]{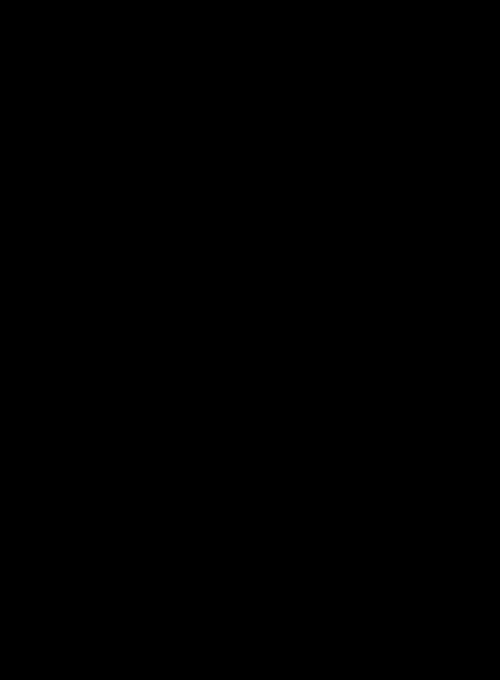}}
\caption{Comparison of results of separation on real mixed X-ray data using double-sided painting the \textsl{Ghent Altarpiece}. Rows 1-3 correspond to the reconstructed X-ray image of the surface painting, the painting on the reverse of the panel, and mixed X-ray image, respectively. Columns 1-3 correspond to the result from \cite{IS5}, result using the proposed algorithm, and result from \cite{IS6}, respectively. Column 4 corresponds to the difference between the result by \cite{IS5} and the proposed algorithm and column 5 corresponds to the difference the between the result by \cite{IS5} and \cite{IS6}.}\label{F-4-Gh-4-1}
\end{figure}

Here, we compare the separation results of the proposed algorithm with the algorithm in \cite{IS6} as shown in Fig. \ref{F-4-Gh-4-1}. 
Note that for the real mixed X-ray image, we do not have the ground truth of the individual separated X-ray images. Instead we used the separation results in \cite{IS5} as reference, wherein RGB images of the paintings on the front and reverse of the panel were used for separation.
The results show that the proposed algorithm produces much better separation than the algorithm in \cite{IS6}. Specifically, as before, using the method in \cite{IS6} results in the separated X-ray image for the surface painting showing too much similarity to the RGB image. The overall intensity of the face, but also details like the parting in Eve’s hair and her second eye are overemphasized when compared to the results using the current approach or those in \cite{IS5}. The separated X-ray image for the painting on the reverse using the method in \cite{IS6} again has more regions of reduced intensity that correspond to overemphasized regions in Eve’s face compared with the result from the proposed algorithm. For the separated X-ray image of the surface painting, the MSE of the difference between the result by \cite{IS5} and the proposed algorithm is 0.0057 and MSE of the difference between the result by \cite{IS5} and \cite{IS6} is 0.0078. For the separated X-ray image for the painting on the reverse, the MSE of the difference between the result by \cite{IS5} and the proposed algorithm is 0.0063 and the MSE of the difference between the result by \cite{IS5} and \cite{IS6} is 0.0098.

\subsection{Experiments with Real Mixed X-ray Data using a single-sided painting with a concealed design, \textsl{Doña Isabel de Porcel}}
\subsubsection{Set-up}

\begin{figure}[t]
    \centering
    \subfigure[]{\includegraphics[height=0.233\textwidth]{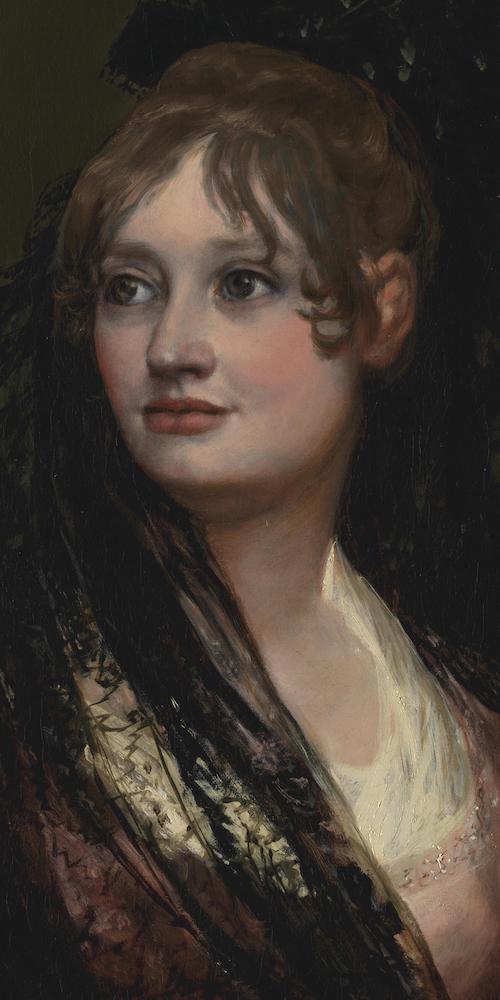}}
    \subfigure[]{\includegraphics[height=0.233\textwidth]{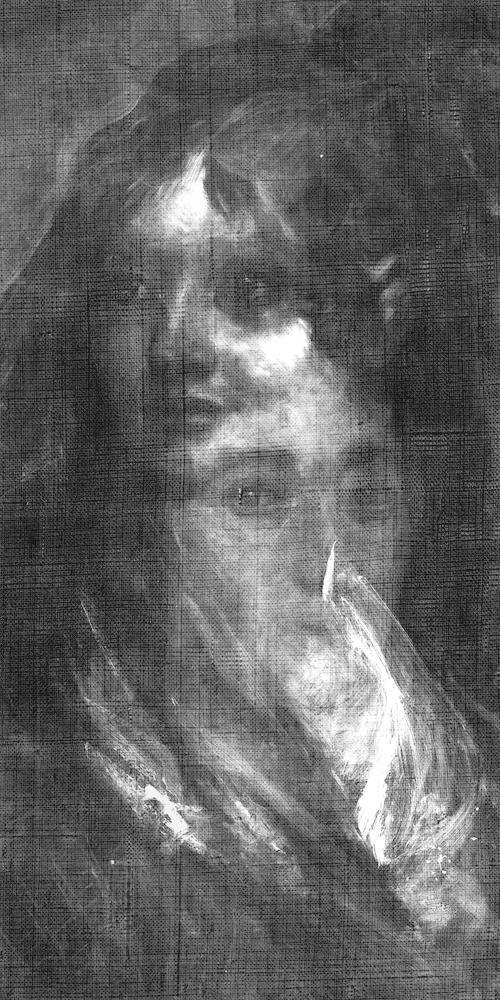}}
    \subfigure[]{\includegraphics[height=0.233\textwidth]{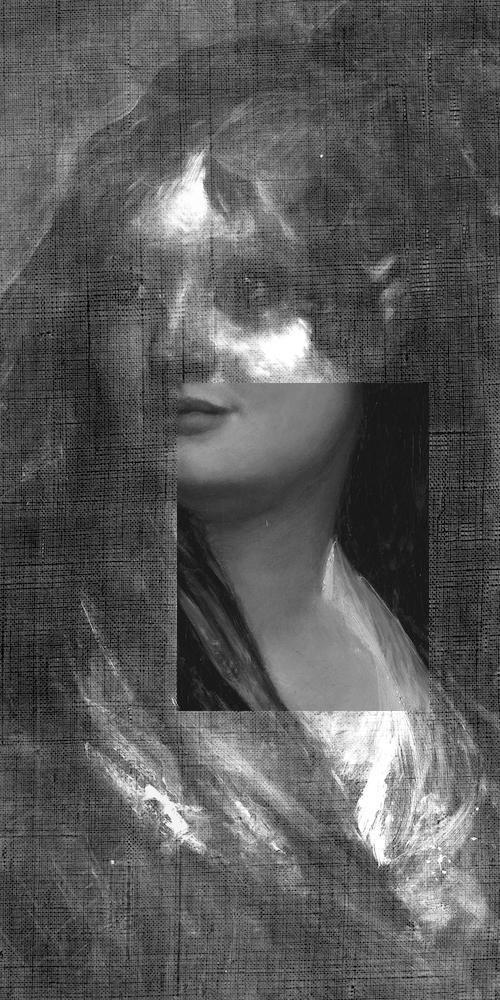}}
    \caption{Images of \textsl{Doña Isabel de Porcel} used in the experiment. (a). RGB image of the surface painting. (b). Mixed X-ray image. (c). Manually modified image of the surface painting. All images are details from the painting shown in Fig. \ref{Goya}.}\label{F-4-4-1-1}
\end{figure}

In this experiment, we use real mixed X-ray data from a painting with a concealed design to demonstrate that our proposed approach can lead to a plausible decomposition in this more challenging scenario.
As shown in Fig. \ref{F-4-4-1-1}, a small area with $500 \times 1000$ pixels from the painting \textsl{Doña Isabel de Porcel} in Fig. \ref{Goya}, wherein both the content of the surface painting and concealed design are obvious in the X-ray image, is utilized to test the proposed method.
The RGB image of the surface painting and the corresponding mixed X-ray image were divided into patches of size 50 $\times$ 50 pixels with 45 pixels overlap (both in the horizontal and vertical direction), resulting in 13,924 patches.

Note that in the previous experiments, the grayscale of the RGB image associated with the surface painting was chosen as ${\boldsymbol g}_1$ in (\ref{2.13}). Here, however, the manually modified image of the surface painting shown in Fig. \ref{F-4-4-1-1} (c) is used as ${\boldsymbol g}_1$ in (\ref{2.13}). Fig. \ref{F-4-4-1-1} (c) is mostly derived from the mixed X-ray image, although a grayscale of the RGB image associated with the surface painting replaces the central portion of the mixed X-ray image because of the observation in Fig. \ref{F-4-4-1-1} (b) that the content of the concealed layer is mainly located in this area.
We again adopted  the  hyper-parameter values $\eta_1 = 0.5 $ and $\eta_2 =0.1$.

\subsubsection{Results}

In the supplementary material, we show the evolution of the overall loss function along with the individual ones as a function of the number of epochs and the results of the proposed network for different training epochs on the painting \textsl{Doña Isabel de Porcel}.

\begin{figure}[t]
\centering
    \subfigure{\includegraphics[height=0.233\textwidth]{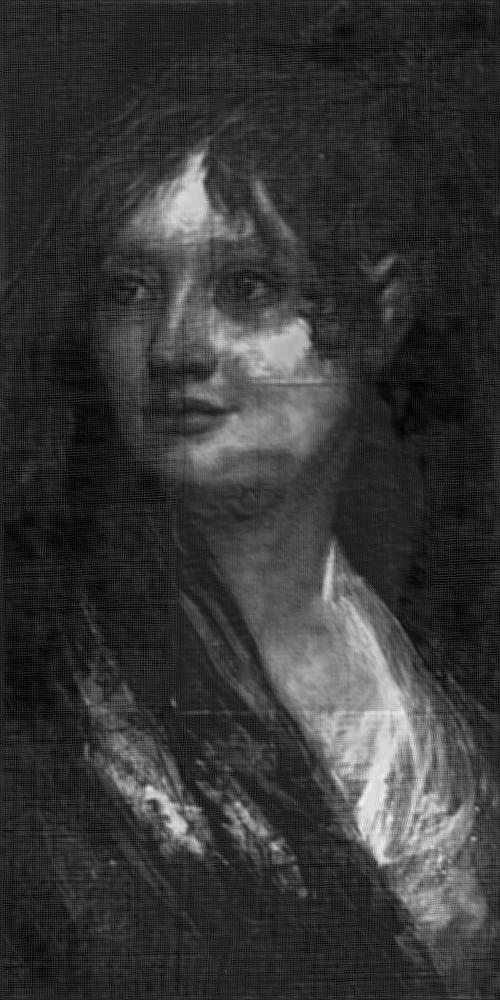}}
    \subfigure{\includegraphics[height=0.233\textwidth]{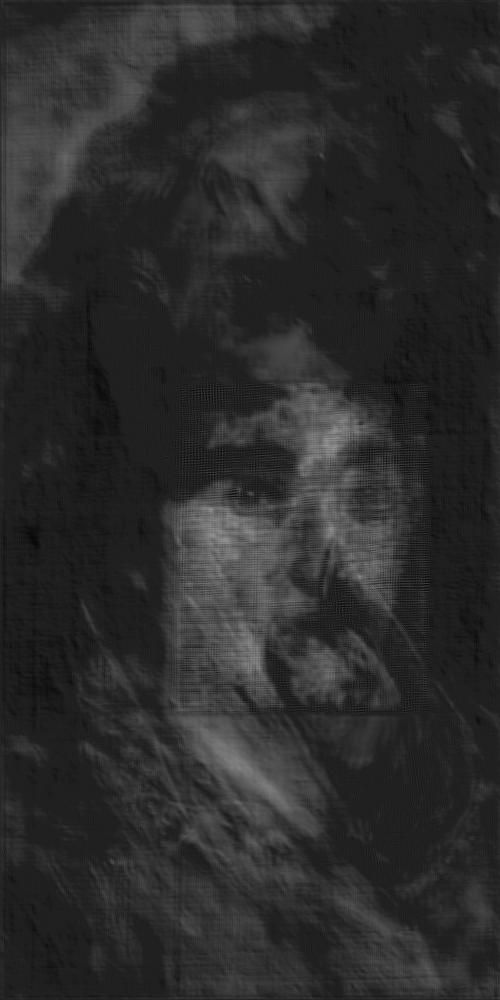}}
    \subfigure{\includegraphics[height=0.233\textwidth]{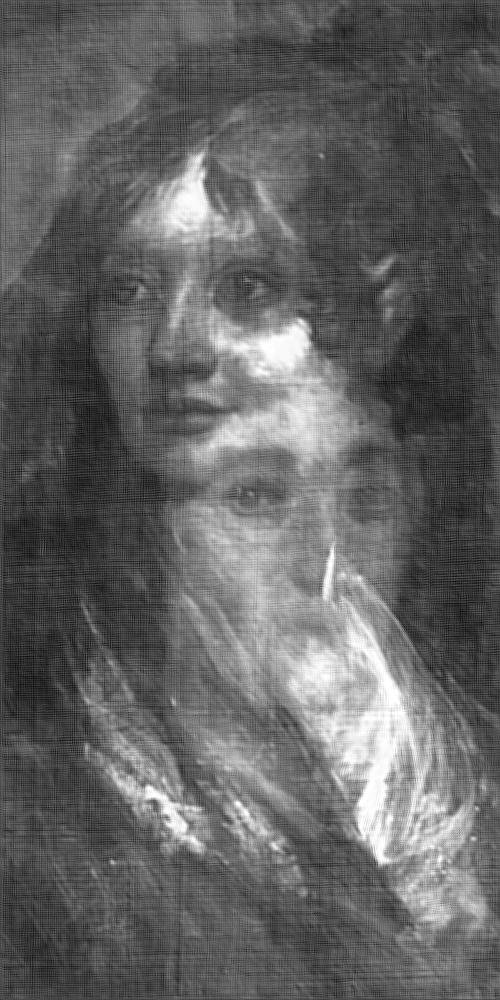}}
    \subfigure{\includegraphics[height=0.233\textwidth]{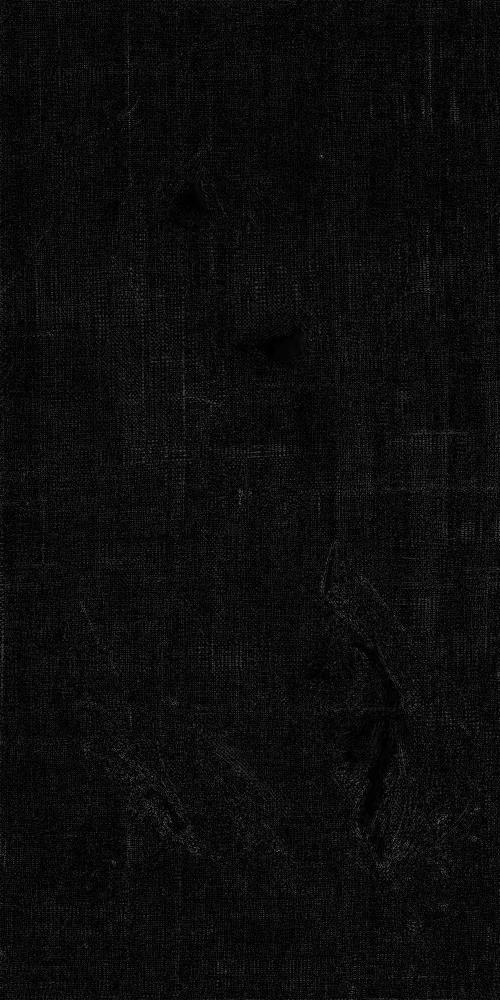}}

    \subfigure{\includegraphics[height=0.233\textwidth]{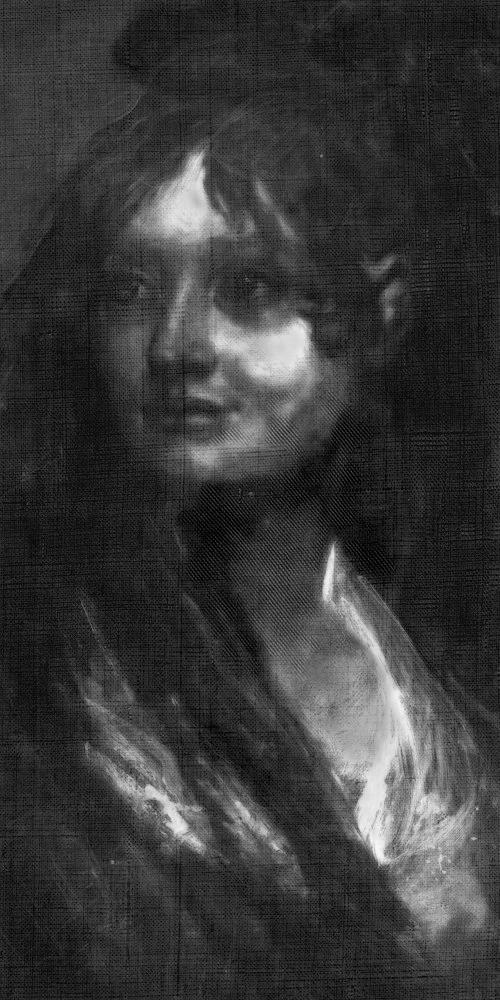}}
    \subfigure{\includegraphics[height=0.233\textwidth]{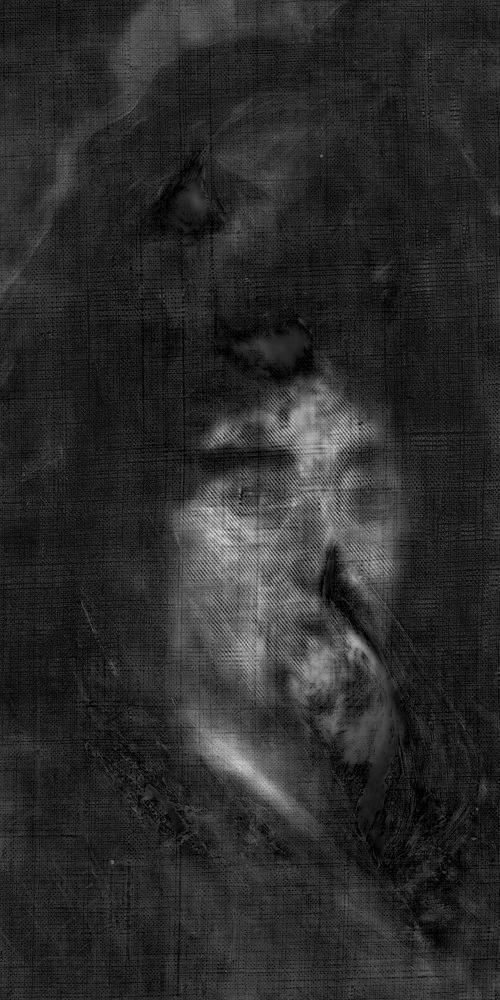}}
    \subfigure{\includegraphics[height=0.233\textwidth]{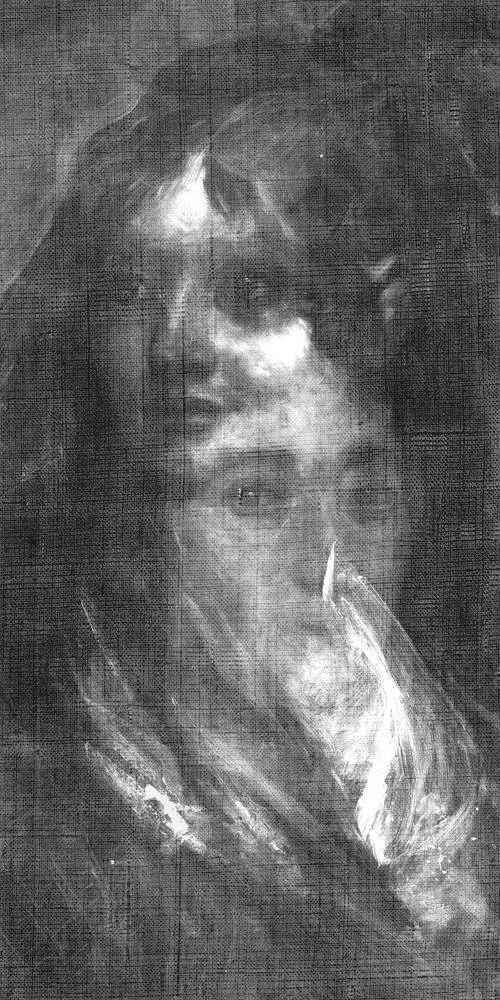}}
    \subfigure{\includegraphics[height=0.233\textwidth]{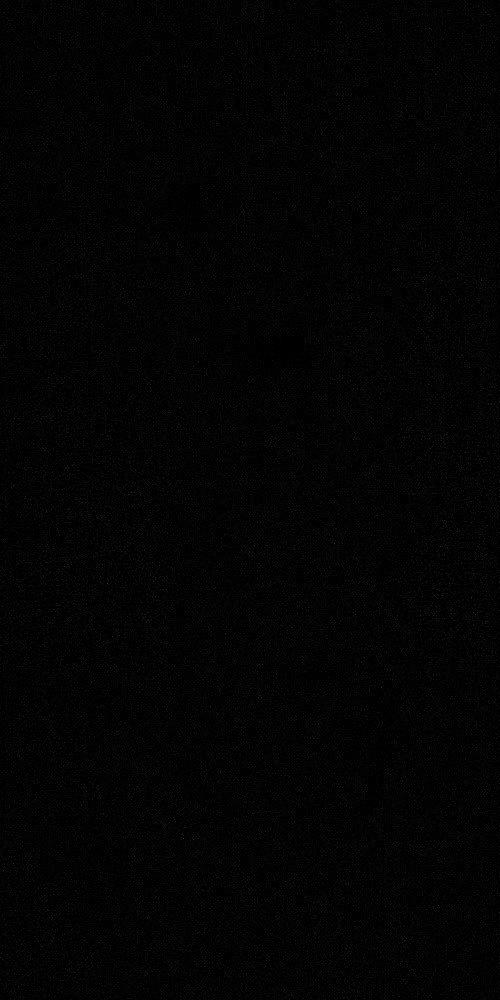}}
\caption{Separation results of \textsl{Doña Isabel de Porcel} data. Columns 1 to 4 correspond to reconstructed X-ray image of the surface painting, reconstructed X-ray image of the concealed design, synthetically mixed X-ray from the separated results, and the error map of the mixed X-ray image, respectively. Rows 1 and 2 correspond to the results by the proposed algorithm and the algorithm in \cite{IS6}, respectively.}\label{F-4-Goya-4-3}
\end{figure}

We compare the separation results of the proposed algorithm with the algorithm in \cite{IS6} as shown in Fig. \ref{F-4-Goya-4-3}. Without access to the ground truth, and with the variable need of different end-users having an influence on how to assess the separation result, a qualitative description of the results is provided. 
Our algorithm performs better on the headdress of Dona Isabel - the method in \cite{IS6} makes it slightly darker in the surface painting X-ray image to try to match the black in the RGB, whereas this feature actually contains materials that absorb X-rays and therefore it should be brighter in the resulting X-ray image. The dark patches in the concealed painting caused by too much intensity given to the forehead and cheek highlights of Dona Isabel in the separated X-ray image of the surface painting are reduced using the proposed approach compared to the method in \cite{IS6} as well.

It is clear there are still some remaining issues with the image separation, for example, the area of the gentleman's face is slightly blurred, but these final images have more of the character that would be anticipated for X-ray images and are likely to feel more familiar and therefore be more appealing to end users.  The fact that Goya may have incorporated aspects of the concealed design into the final portrait adds a further complication as it makes a completely `clean' separation of the two images even more challenging.

\begin{figure}[t]
\centering
    \subfigure{\includegraphics[height=0.233\textwidth]{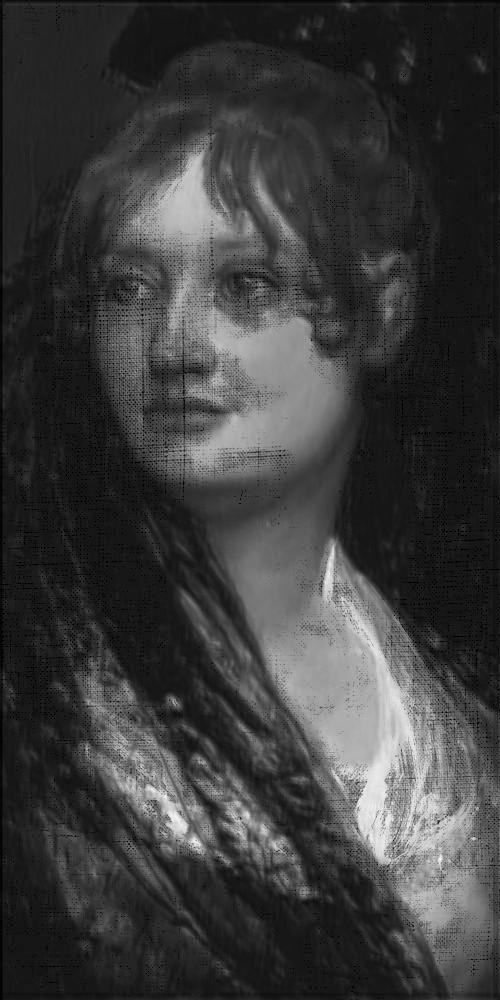}}
    \subfigure{\includegraphics[height=0.233\textwidth]{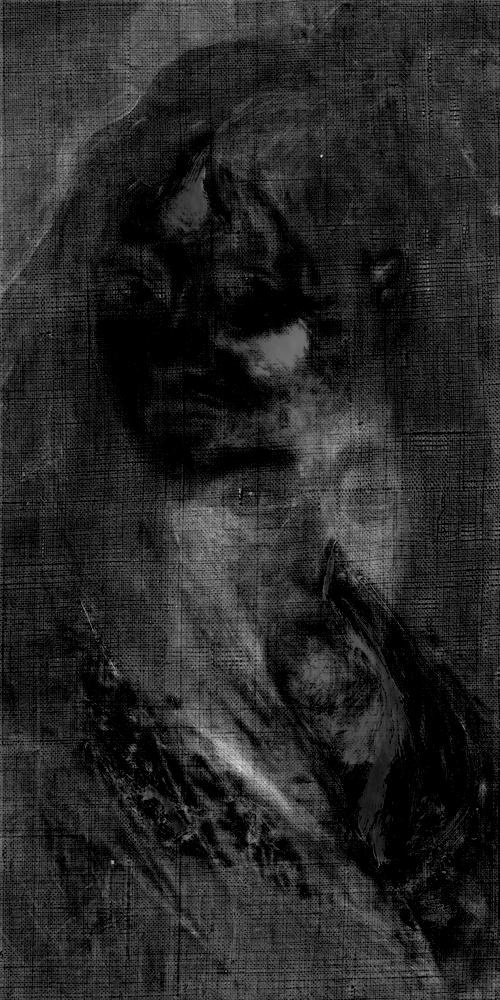}}
    \subfigure{\includegraphics[height=0.233\textwidth]{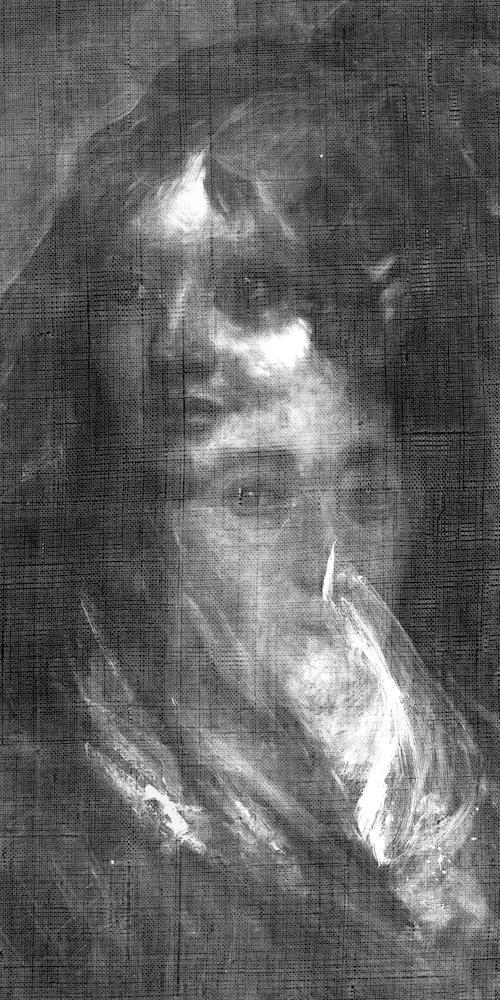}}
    \subfigure{\includegraphics[height=0.233\textwidth]{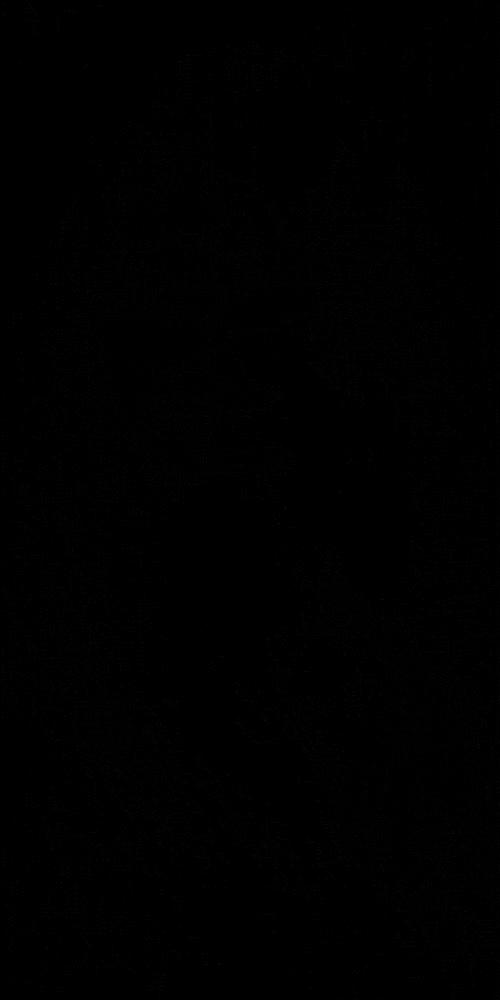}}

    \subfigure{\includegraphics[height=0.233\textwidth]{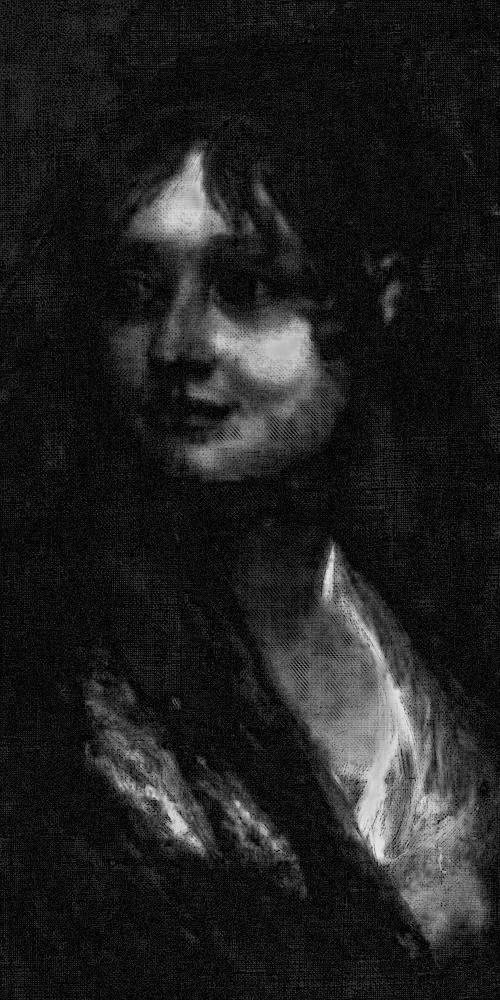}}
    \subfigure{\includegraphics[height=0.233\textwidth]{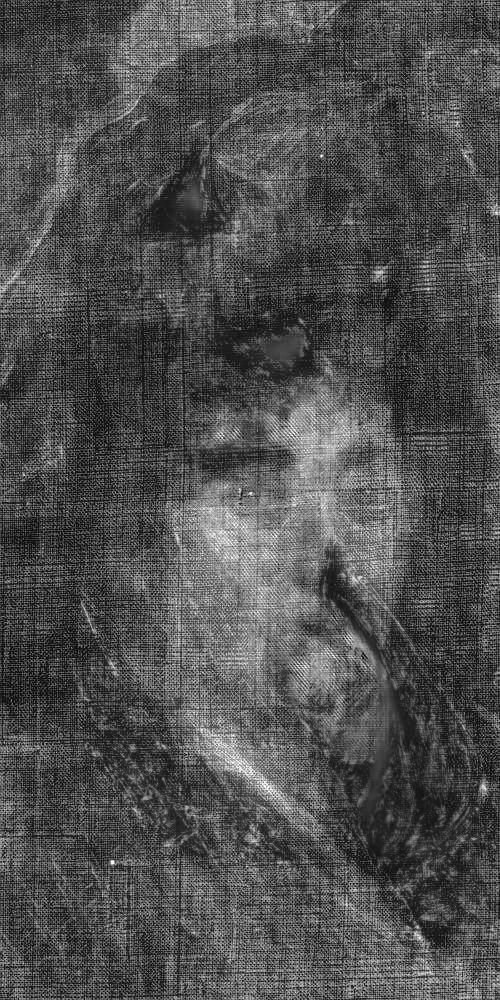}}
    \subfigure{\includegraphics[height=0.233\textwidth]{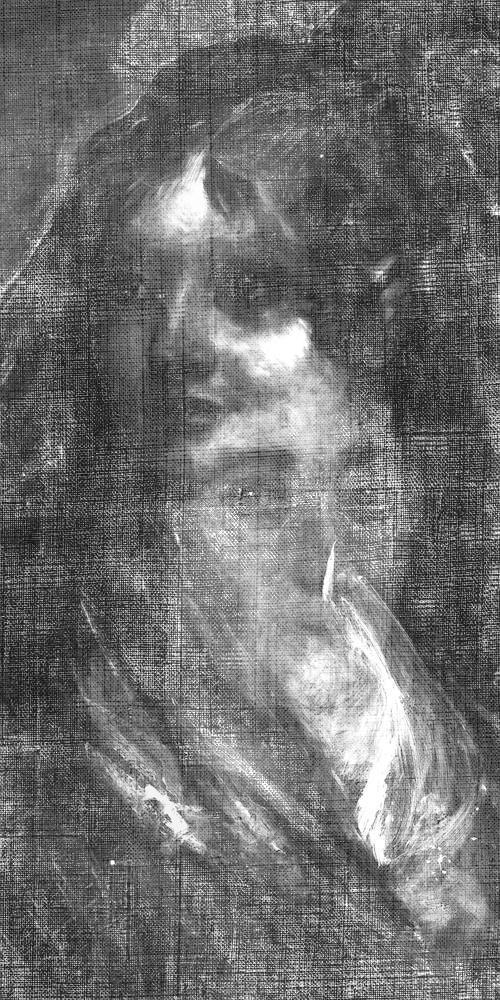}}
    \subfigure{\includegraphics[height=0.233\textwidth]{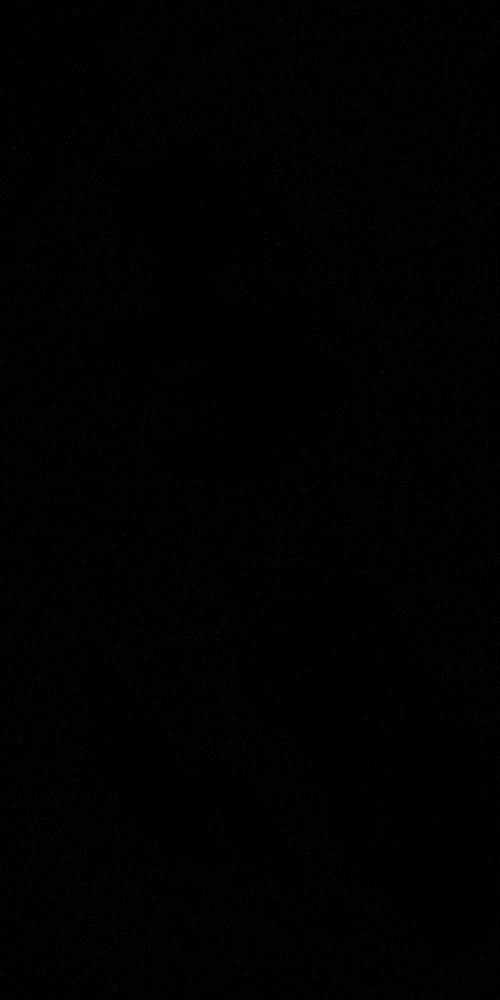}}
\caption{Separation results of \textsl{Doña Isabel de Porcel} data using the grayscale image of the surface painting as ${\boldsymbol g}_1$. Columns 1 to 4 correspond to the reconstructed X-ray image of the surface painting, reconstructed X-ray image of the concealed painting, synthetically mixed X-ray from the separated results, and the error map of the mixed X-ray image, respectively. Rows 1 and 2 correspond to the results by the proposed algorithm and the algorithm in \cite{IS6}, respectively.}\label{F-4-Goya-4-4}
\end{figure}

Note that in Fig. \ref{F-4-Goya-4-3}, we use the manually modified image of the surface painting in Fig. \ref{F-4-4-1-1} (c) as ${\boldsymbol g}_1$ during the initialization. If we still use the grayscale image as ${\boldsymbol g}_1$ during the initialization, the separation results by the proposed approach are shown in the first row of Fig. \ref{F-4-Goya-4-4}. In \cite{IS6}, the manually modified image of the surface painting in Fig. \ref{F-4-4-1-1} (c) is also utilized. If we use the grayscale image of the surface painting instead of the manually modified image, the separation results by the algorithm in \cite{IS6} are shown in the second row of Fig. \ref{F-4-Goya-4-4}. 
{\color{black}
Comparing the results in Fig. \ref{F-4-Goya-4-4}, again the proposed approach still outperforms the algorithm in \cite{IS6} because the separated X-ray image of the surface painting obtained by the proposed method contains much more detailed information, and the content of the surface painting is less obvious in the separated X-ray image of the concealed design obtained by the proposed method. Comparing the results in Fig. \ref{F-4-Goya-4-3} and Fig. \ref{F-4-Goya-4-4}, there is quite obvious improvement if we use the manually modified image as ${\boldsymbol  g}_{1}$ with the algorithm in \cite{IS6}, and the synthetically mixed X-ray  from the separated results clearly differs from the original mixed X-ray image for example. For the proposed approach the comparison is more subtle. The separated X-ray for the surface painting appears to have become more like a grayscale version of the surface painting but some areas of the X-ray image for the concealed design are arguably slightly clearer.}

\section{Conclusions}
X-radiography is a useful tool in the technical study of artworks as, amongst its other benefits, it is capable of providing insights into concealed compositions and \textsl{pentimenti} as well as information about the condition and construction. However, when concealed designs exist under the visible surface the resulting X-ray images contain mixed features associated with both visible and concealed designs as well as features associated with areas of damage and the structure of the painting support for example. As a result, it becomes more difficult for experts to interpret these images. To improve the utility of these X-ray images, it is desirable to separate the content into two (hypothetical) images, each pertaining to a single composition. 
This paper proposes a new approach to X-ray image separation as a valuable addition to methods published previously and as a tool for further work on this challenging problem. Although the precise measure of the success of the separation of such X-ray images is dependent on the exact needs of the different end users, this new approach addresses a number of the issues noted with the state-of-the-art separation algorithms in side-by-side experiments on images from multiple paintings, and outperforms these algorithms based on MSE values (in cases where these can be calculated).
In particular, by leveraging sparsity-driven data models, sparsity-driven data processing, and algorithm unrolling techniques, we have derived a self-supervised learning approach using separation network structure. This approach allows image separation without the need for labelled data. A composite loss function is introduced to guide the training of the separation network.
This proposed method is demonstrated with experiments on images from synthetic data and images from the \textsl{Ghent Altarpiece} by Hubert and Jan van Eyck and \textsl{Doña Isabel de Porcel} by Francisco de Goya with promising results.

\end{document}